\documentclass[12pt,a4paper]{article}

\usepackage{amsfonts}
\usepackage{amsmath}
\usepackage{amsbsy}
\usepackage{theorem}
\usepackage{graphicx}
\usepackage{amssymb}
\usepackage{epstopdf}

\begin{document}

\sf
\title{\bfseries \normalsize  COVID--19 MORTALITY ANALYSIS FROM SOFT--DATA  MULTIVARIATE CURVE REGRESSION AND   MACHINE LEARNING
}

\date{}
\author{\normalsize    A. Torres--Signes, M. P. Fr\'{\i}as  and M. D. Ruiz--Medina}
\maketitle
\begin{abstract}
 A multiple objective
space--time forecasting  approach is presented involving cyclical
curve log--regression, and multivariate time series spatial residual
correlation analysis. Specifically, the mean quadratic loss function is
minimized in the framework of trigonometric regression. While, in
our subsequent  spatial residual correlation analysis, maximization
of the likelihood allows us to compute the posterior mode in a
Bayesian multivariate time series soft--data framework. The
presented approach is applied to the analysis
 of COVID--19 mortality in the first wave affecting   the Spanish Communities, since March, 8, 2020
  until  May, 13, 2020. An empirical comparative study with Machine Learning (ML) regression,
  based on random $k$--fold cross-validation,
 and bootstrapping confidence interval and probability density estimation,  is carried out. This empirical analysis
also investigates  the performance of  ML regression models in a
hard-- and  soft-- data frameworks. The results could be extrapolated to other counts,
countries, and posterior COVID--19 waves.

\medskip



\end{abstract}

\section{Introduction}
Coronavirus disease 2019 (COVID--19)  rapidly spreads around  many
other countries,  since  December 2019 when  arises in China (see  \cite{Sivakumar20};
\cite{Wang20}; \cite{Zhou20}).  The effective allocation of medical resources requires
the derivation of predictive techniques, describing the  spatiotemporal  dynamics of  COVID-19 (see, e.g., \cite{Du20}; \cite{Khan20};  \cite{Nishiura20}; \cite{Remuzzi20},
  just to mention a few).  Epidemiological models can contribute to the  analysis of  the causes,
 dynamics, and spread of this pandemic
(see, e.g,  \cite{Huppert}; \cite{Keeling}; \cite{Laaroussi18},   and the references therein).
Short-term forecasts can be  obtained adopting the framework of compartmental  SIR
  (susceptible-infectious-recovered) models, based on ordinary differential equations  (see, e.g. \cite{Angulo13} \cite{Elhia14}; \cite{Ji12}; \cite{Kermack}; \cite{Kuznetsov20}; \cite{Milner08}; \cite{Pathak10}; \cite{Tornatore}; \cite{Yu09}; \cite{Zhang08}). An extensive literature is available, including different versions of  compartmental  models, like  SIR-susceptible (SIRS, \cite{Dushoff04}), and  delay
differential equations  (see \cite{Beretta}; \cite{McCluskey10}; \cite{Sekiguchi10}). Spatial extensions, based on
reaction-diffusion models,
reflecting the
infectious disease spread over a spatial region can be found, for instance, in
\cite{Guin14} and  \cite{Webb81}.  SEIRD (susceptible, exposed, infected, recovered,
deceased) models,  incorporating the spatial spread of the
disease with inhomogeneous diffusion terms are also analyzed    (see
\cite{Roques16}  and \cite{Roques11}).
 The stochastic version of SIR--type models intends to cover several limitations detected regarding uncertainly in the observations, and the hidden dynamical epidemic process. Markov chain  SIR based modelling (see \cite{Anderson}; \cite{Xu07}), and some
 recent stochastic formulations involving complex networks (see \cite{Volz08}; \cite{Zhou06}) or drug--resistant influenza (see
\cite{Chao12}) constitute some alternatives.  A  Bayesian hierarchical statistical SIRS model
framework  is adopted in \cite{Aalen08};   \cite{Abboud19};   \cite{Anderson}; \cite{Fleming91} taking into account the observation
error in the counts, and uncertainty in  the parameter space.
  Beyond SIR modeling, the multivariate  survival analysis approach offers a suitable  modelling framework, regarding   infection, incubation and  recovering random
periods, affecting the containment of COVID-19
  (see, e.g., \cite{Bolker96} ; \cite{Keeling2}; \cite{Pak2020}; \cite{Wasiur19}).

  In a first stage, most of the above referred  models have been adapted and applied to approximate the space/time evolution of COVID--19
  incidence and mortality. That is the case, for instance, of the three models  presented in \cite{Roosa2020},
which were validated with outbreaks of other diseases different from
COVID--19.  Alternative SEIR type models,  involving stochastic components, are
formulated in
\cite{Kucharski20}. A revised SEIR model  has also been proposed in \cite{Zhang20} (see also \cite{He20}).  A $\theta $--SEIHRD
model,  able to
estimate the number of cases, deaths, and needs of beds in
hospitals, is introduced in \cite{Ivorra20a}, adapted to COVID--19, based on the
Be-CoDiS model (see \cite{Ivorra20b}).  Due to the low quality of the records available, and the hidden sample information, the most remarkable feature in this research area is the balance between complexity and
indentifiability of model parameters. Recently, an attempt to simplify modelling strategies, applied to COVID-19  data analysis,
is presented in \cite{Ramos2020}, in terms of  $\theta $--SEIHQRD model.  Mitigation of undersampling is proposed in \cite{Langousis20}, based on re-scaling of summary statistics characterizing sample properties of the pandemic process, useful between countries with similar levels of health care.

Nowadays ML models have established themselves  as serious
contenders to classical statistical models in the area of
forecasting. Research started in the eighties with the development
of the neural network model. Subsequently, research extended this
concept to alternative models, such as support vector machines,
decision trees, and others (see, e.g., \cite{Alpaydin04};   \cite{Blanqueroetal20};
\cite{Hastie01}; \cite{Mohammady21}).
In  general, curve regression techniques based on  a function basis, usually in the space of square integrable
 functions with respect to a suitable probability measure, allow short-- and long-- term forecast. Thus,   depending on our choice of the  function basis, and the probability measure selected,
 \emph{particle} and \emph{field} views could be combined. Note that the classical stochastic diffusion models offer a \emph{particle}
 rather than a \emph{field} view  (see, e.g., \cite{Malesios16}).

Linear regression, multilayer perceptron and vector
autoregression methods have been applied in
\cite{Sujath20a}-\cite{Sujath20b} to  predicting COVID-19 spread,   anticipating  the potential patterns of COVID-19 effects   (see also Section 2 of \cite{Sujath20a}, on related work). Early stage
location of COVID-19 is addressed in  \cite{Barstugan20}, applying   machine learning strategies
actualized on stomach Computed Tomography pictures.
\cite{Chien20} evaluates association between meteorological factors and COVID--19 spread.  They concluded that
average temperature, minimum relative humidity, and precipitation were better predictors,  displaying possible non--linear correlations with COVID--19 variables.
These conclusions are crucial in the subsequent machine learning regression based analysis.

 This paper presents a multiple objective
space--time forecasting approach, where curve trigonometric  log--regression is combined with multivariate time series spatial residual analysis.    In our curve regression model fitting, we are interested on reflecting the cyclical behavior of
COVID--19 mortality induced   by the hardening or relaxation of the
containment measures, adopted to mitigate the increase of infections
and mortality. The trigonometric basis (sines and cosines) is then
selected in our spatial heterogeneous curve log--regression model
fitting. The ratio of the expected minimized empirical risk,  and
the corresponding expected value of the quadratic loss function at
such a  minimizer is considered for model selection (see, e.g.,
\cite{Chapelle}). Note that this selection procedure provides an
agreement between the expected minimum empirical risk, and the
corresponding expected theoretical loss function value.

The penalized factor proposed in \cite{Chapelle}, applied to our
choice of the truncation parameter, leads to the dimension of the
subspace where our curve regression estimator is approximated at
any spatial location. This model selection procedure  is
asymptotically equivalent to Akaike correction factor. A robust
modification of the Akaike information criterion can be found, for
example, in \cite{Agostinelli01}. As an alternative, one can
consider cross-validation criterion for selecting the best subset of
explanatory variables  (see \cite{Takano20}, where a mixed-integer
optimization approach is proposed in this context).

Beyond asymptotic analysis,
  model selection  from finite
sample sizes constitutes a challenging topic in our approach. To address this problem,  a bootstrap estimator of  the ratio between the  expected quadratic loss function and  the expected training quadratic error, from different sets of explanatory variables, is implemented. Bootstrap confidence
intervals are  also provided for the spatial mean of the  curve regression predictor, and
for the expected training error of the curve regression, and of the  multivariate time--series residual
 predictor.
The bootstrap approximation of the probability distribution of these  statistics is
also computed.

In our multivariate time series
analysis of the regression residuals,  a classical and Bayesian componentwise estimation of the
spatial linear correlation is achieved.
 The presented  multiple objective forecasting approach  is applied
 to the spatiotemporal analysis
 of COVID--19 mortality in the first wave affecting   the Spanish Communities, since March, 8, 2020
  until  May, 13, 2020. Our results show a
  remarkable qualitative agreement with the reported epidemiological
data.

  The spatiotemporal approach presented in this paper makes the fusion of generalized random field theory, and our multiple--objective
space--time forecasting, based on nonlinear parametric regression, and bayesian analysis of the spatiotemporal correlation structure.  Regarding the \emph{site--specific} or \emph{specificatory} knowledge bases (see \cite{Christakos}),  in our approach,  several  information  sources  can be incorporated in the description of the hidden epidemic process. Particularly, we distinguish here between hard--data
or hard measurements providing a satisfactory level of accuracy for practical purposes, and soft--data  displaying a non--negligible amount of uncertainty. That is, in this second data category, we include missing observations or imperfect observations, categorical data and fuzzy inputs
(see  also  \cite{Christakos00}; \cite{Christakos02}; \cite{ChristakosHris}, and the references therein). In this paper, we consider hard--data sets given by  numerical values
of our count process at the Spanish Communities analyzed. Our soft--data sample  complements hard measures, in terms of  interpolated, smoothed,  and spatial projected data. Particularly,  spatial correlations between  regions are incorporated  in terms of soft--data.  Additional information about the continuous functional nature of the underlying
space--time COVID--19 mortality process is also reflected in our soft--data set. This information helps  the implementation of the proposed estimation methodology in the framework of Functional Data Analysis (FDA) techniques.

 As commented before, last advances in spatiotemporal mapping of epidemiological data incorporate ML regression models to improve and  help the understanding of \emph{general or core  knowledge bases}. Thus,  model fitting is achieved according to epidemiological systems laws, population dynamics, and theoretical space--time dependence models (see \cite{Christakos08}, and the references therein).  See also \cite{Barstugan20}, \cite{Chien20} and \cite{Sujath20a} in the hard--data context.
It is well--known that the limited availability of hard--data affects
space-time analysis. Hence, the incorporation of soft--data into
ML regression models  can  help this analysis, providing a global
view of the available sample information (see, e.g.,
\cite{Christakos}).   Particularly, in our  empirical comparative analysis,  involving   ML regression models
and our approach, input hard-- and
soft--data information is incorporated.  Cross--validation,  bootstrapping confidence intervals and probability density estimation support our comparative study.
Specifically, random $k$--fold ($k=5,10$) cross--validation first evaluates the performance of the compared regression models from hard-- and soft--data, in terms of Symmetric Mean Absolute Percentage Errors (SMAPEs). Bootstrap confidence intervals and  probability density estimation of the spatially averaged  SMAPEs  approximate the distributional characteristics of the random  $k$--fold cross--validation  errors. Thus, a complete  picture of SMAPEs supports our evaluation of the predictive ability of the regression models tested, from  the analyzed hard-- and soft--data  sets.

From the empirical comparative analysis carried out, we can  conclude that almost the best performance in both, hard-- and soft--data  categories,   is displayed by Radial Basis Function Neural Network (RBF),
and Gaussian Processes (GP). Both approaches are improved, when soft--data are incorporated into the regression analysis. Slightly   differences are observed in the performance of  Support
Vector Regression (SVR) and  Bayesian Neural Networks (BNN).      Multilayer Perceptron (MLP) gets over GRNN, presenting  better estimation results  when hard--data are analyzed.   The sample values and distributional characteristics  of cross--validation  SMAPEs, in  Generalized Regression Neural Network (GRNN),  are similar to the ones obtained in trigonometric curve regression,  when spatial residual analysis is achieved in terms of empirical second--order moments.
Note that, GRNN is also favored by the soft--data category.
In this category, BNN and our approach show very similar performance, when  trigonometric regression is combined with  Bayesian multivariate time series residual prediction. Indeed,
some slightly better bootstrapping distributional characteristics of our approach
respect to BNN are observed in the soft--data category.

The outline of the paper is the following. The  modeling approach is
introduced in Section \ref{s2}. Section \ref{sectptp} describes the
multiple objective forecasting  methodology. This methodology is
applied to the spatiotemporal statistical analysis of COVID--19
mortality in Spain in Section \ref{sac19}. The empirical comparative
study with ML regression  models is given in Section \ref{subtsb}. Conclusions about our
data--driven model ranking can be found  in Section \ref{CR}. In the
Supplementary Material,  a brief introduction to our implementation of ML models
from hard-- and  soft--data  is provided. Additional numerical estimation results, based on the complete sample,  are also displayed.
Particularly, the observed and predicted mortality cumulative cases, and log--risk  curves are displayed.

\section{Data model}
\label{s2}
 Let $(\Omega ,\mathcal{A},\mathcal{P})$   be the basic probability space.
 Consider  $H=L^{2}(\mathbb{R}^{d}),$ $d\geq 2,$ the space of square--integrable functions on $\mathbb{R}^{d},$ to be  the  underlying  real separable  Hilbert space. In the following, we denote by $\mathcal{B}^{d}$  the Borel $\sigma$--algebra  in $\mathbb{R}^{d},$ $d\geq 1.$

  Let  $X=\{ X_{t}(\mathbf{z}),\  \mathbf{z}\in  \mathbb{R}^{d},
 \ t\in \mathbb{R}_{+}\}$ be our spatiotemporal input hard--data process on $(\Omega ,\mathcal{A},\mathcal{P}),$  satisfying  $E\left[\|X_{t}(\cdot)\|_{H}^{2}\right]<\infty,$  for any time $t\in \mathbb{R}_{+}.$  The  input soft--data process over any spatial bounded set  $D\in \mathcal{B}^{d}$ is then defined as
 \begin{equation}
 \left\{X_{t}(h)= \int_{D}X_{t}(\mathbf{z})h(\mathbf{z})d\mathbf{z},\ h\in \mathcal{C}_{0}^{\infty}(D),\ t\in \mathbb{R}_{+}\right\},
 \label{lr}
 \end{equation}
 \noindent where $\mathcal{C}_{0}^{\infty}(D)$ denotes the space of infinite differentiable functions, with compact support contained in $D.$
  For each bounded set $D\in \mathcal{B}^{d},$  define $$\Lambda =\{\Lambda_{t}(h)=\exp\left(X_{t}(h)\right),\  h\in \mathcal{C}_{0}^{\infty}(D),\ t\in \mathbb{R}_{+}\}.$$
   Assume that, for any finite positive interval $\mathcal{T}\in \mathcal{B},$ and  bounded set  $D\in \mathcal{B}^{d},$
 \begin{eqnarray}
   \mathcal{I}_{\mathcal{T}}(h)&\underset{\mathcal{L}^{2}(\Omega ,\mathcal{A},P)}{=}&\int_{\mathcal{T}}\exp\left(X_{t}(h)\right)dt<\infty,\quad \forall h\in \mathcal{C}_{0}^{\infty}(D),
   \label{eg1}
   \end{eqnarray}
 \noindent where $\underset{\mathcal{L}^{2}(\Omega ,\mathcal{A},P)}{=}$ denotes the identity in the second--order moment sense. Let  $\{N_{h}:(\Omega ,\mathcal{A},\mathcal{P})\times \mathcal{B}\longrightarrow \mathbb{N}, \ h\in H\}$ be a family
  of random counting measures. Given the observation $\left\{x_{t}(h),\ t\in \mathcal{T}\right\}$ at the finite  temporal interval $\mathcal{T}\in \mathcal{B}$  of the input soft--data process over the spatial $h$--window in $D,$
   the  conditional probability distribution of the number of random  events   $N_{h}(\mathcal{T})$  that occur in $\mathcal{T} \in \mathcal{B}$  is a Poisson probability distribution  with  mean $\int_{\mathcal{T}}\exp\left(x_{t}(h)\right)dt,$ for every   $h\in \mathcal{C}_{0}^{\infty}(D)$  and  $D\in \mathcal{B}^{d}.$
   We refer to  $\mathcal{I}_{\mathcal{T}}(h)$    as the generalized  cumulative  mortality risk random process over the interval $\mathcal{T}.$ Hence, the input hard--data process   $X=\{ X_{t}(\mathbf{z}),\  \mathbf{z}\in  \mathbb{R}^{d},
 \ t\in \mathbb{R}_{+}\}$ defines the spatiotemporal  mortality  log--risk process.

 From the sample values of our  input soft--data process, the following observation model is considered in the curve regression model fitting
   \begin{eqnarray}
  && \hspace*{-0.5cm}\ln\left(\Lambda_{t} \right)(\psi_{p,\varpi_{p}})= g_{t}(\psi_{p,\varpi_{p}}, \boldsymbol{\theta }(p) )+\varepsilon_{t}(\psi_{p,\varpi_{p}})\nonumber\\
   &&= \left\langle g_{t}(\cdot,\boldsymbol{\theta }(p) ), \psi_{p,\varpi_{p}}(\cdot)\right\rangle_{H}+ \left\langle \varepsilon_{t}(\cdot ),\psi_{p,\varpi_{p}}(\cdot)\right\rangle_{H},\ t\in \mathbb{R}_{+},\ p=1,\dots,P,\label{om}
   \end{eqnarray}
\noindent   where
\begin{eqnarray} g_{t}(\psi_{p,\varpi_{p}}, \boldsymbol{\theta }(p) )&=&\int_{\mathcal{D}_{p}}g_{t}(\mathbf{z}, \boldsymbol{\theta }(p))
\psi_{p,\varpi_{p}}(\mathbf{z}) d\mathbf{z}\nonumber\\
\left\langle f,g\right\rangle_{H}&=&\int_{\mathbb{R}^{d}}f(\mathbf{z})g(\mathbf{z})d\mathbf{z}, \label{eqgreg}
 \end{eqnarray}
\noindent with $\{\psi_{p,\varpi_{p}},\ p=1,\dots,P\}\subset H$
denoting a function family in $H,$  whose elements    have respective compact
supports $\mathcal{D}_{p},$   $p=1,\dots,P,$ defining the $p$
small--areas where the counts are aggregated, satisfying suitable regularity conditions.  For each
$p=1,\dots,P,$  the vector $\varpi_{p}$  contains the center and
bandwidth parameters, defining the window  selected in the analysis
of the small--area $p.$  For each $p\in \{1,\dots,P\},$
$\boldsymbol{\theta }(p) =(\theta ^{1}(p),\ldots ,\theta ^{q}(p))\in
\Theta $ represents   the unknown parameter vector to be estimated
at the $p$ region, and   $\Theta $ is the open set defining  the
parameter space, whose closure  $\Theta ^{c}$ is a  compact set  in
$\mathbb{R}^{q}.$
 We assume that  $g_{t}$ is of the form  (see, e.g., \cite{Ivanov15})
\begin{equation}
g_{t}(\boldsymbol{\theta }(p))=\sum_{k=1}^{N} \left(
A_{k}(p)\cos (\varphi _{k}(p) t)+B_{k}(p)\sin (\varphi
_{k}(p)t)\right), \ p=1,\dots,P,\quad t\in \mathbb{R}_{+},\label{nonregr2}
\end{equation}
\noindent whose  spatial--dependent parameters are given by the
temporal scalings \linebreak $\left(\varphi _{1}(\cdot),\dots,
\varphi _{N}(\cdot)\right),$ and the Fourier coefficients $\left(A_{1}(\cdot), B_{1}(\cdot),\ldots, A_{N}(\cdot),
B_{N}(\cdot)\right).$ For simplifications purposes, we will consider that the scaling
parameters $\varphi _{k},$ $k=1,\dots,N,$ are known, and fixed over the $P$ spatial regions.
Also,  $C_{k}^{2}(\cdot)=
A_{k}^{2}(\cdot)+B_{k}^{2}(\cdot)>0,$ for $k=1,\ldots ,N,$ where $N$
denotes the truncation parameter, that will be selected according to
the penalized factor proposed in \cite{Chapelle}, as we explain in
more detail in Section \ref{sectptp}. Thus,
   \begin{eqnarray}\boldsymbol{\theta }(p)&=&(A_{1}(p),B_{1}(p),\ldots
,A_{N}(p),B_{N}(p)),\quad p=1,\dots, P.
 \nonumber\label{pvfs}\end{eqnarray}

To analyze the spatial correlation between regions, a multivariate
autoregressive model is considered  for prediction of the regression residual
 term at each region $p\in \{1,\dots, P\}.$ Particularly,
for any $T\geq 2,$   $\varepsilon_{t}$ in
 equation (\ref{om})    is assumed to satisfy   the  state equation, for $p=1,\dots, P,$
\begin{equation}
\varepsilon_{t}(\psi_{p,\varpi_{p}})=\sum_{q=1}^{P}\rho(\psi_{q,\varpi_{q}})(\psi_{p,\varpi_{p}})\varepsilon_{t-1}(\psi_{q,\varpi_{q}})
+\nu_{t}(\psi_{p,\varpi_{p}}),\label{arh1}
\end{equation}
\noindent where, for any $t\in \mathbb{R}_{+},$ and $p,q=1,\dots, P,$
\begin{eqnarray}
\varepsilon_{t}(\psi_{p,\varpi_{p}})&=&\int_{\mathcal{D}_{p}}\varepsilon_{t}(\mathbf{z})\psi_{p,\varpi_{p}}(\mathbf{z})d\mathbf{z}\nonumber\\
\nu_{t}(\psi_{p,\varpi_{p}})&=&\int_{\mathcal{D}_{p}}\nu_{t}(\mathbf{z})\psi_{p,\varpi_{p}}(\mathbf{z})d\mathbf{z}\nonumber\\
\rho(\psi_{q,\varpi_{q}})(\psi_{p,\varpi_{p}})&=&\int_{\mathcal{D}_{p}\times
\mathcal{D}_{p}}\rho(\mathbf{z},\mathbf{y})\psi_{p,\varpi_{p}}(\mathbf{z})
\psi_{q,\varpi_{q}}(\mathbf{y})d\mathbf{y}d\mathbf{z}.\nonumber \label{eqvarp}
\end{eqnarray}
\noindent   Here,
$\left(\nu_{t}(\psi_{p,\varpi_{p}}),\ p=1,\dots, P\right),$  $t\in \mathbb{R}_{+},$  are
assumed to be independent zero--mean Gaussian  $P$--dimensional vectors.  For $p,q\in \{1,\dots P\},$
the  projection
$\rho(\psi_{p,\varpi_{p}})(\psi_{q,\varpi_{q}})$ then keeps the
temporal linear   autocorrelation at each spatial region  for $p=q,$
  and the  temporal linear   cross-correlation between
regions   for $p\neq q$ of the regression error $\{\varepsilon_{t}(\cdot),\ t\in \mathbb{R}_{+}\}$ (see,
\cite{Bosq2000}).

\section{Implementation of the curve regression model and spatial residual analysis}

\label{sectptp}

   Let $\mathcal{D}_{1},\dots,\mathcal{D}_{P}$ be the  small-areas, where the counts are aggregated, and $\{\psi_{p,\varpi_{p}},\ \varpi_{p}=(c_{p},\rho_{p}),\   p=1,\dots ,P\}\subset H$ be the functions with respective compact supports  $\mathcal{D}_{1},\dots,\mathcal{D}_{P}.$
     Particularly, we denote by $c_{p},$ $p=1,\dots,P,$  the centers  respectively allocated at the
      regions $\mathcal{D}_{1},\dots,\mathcal{D}_{P},$ and by $\rho_{1},\dots,\rho_{P},$
     the    bandwidth parameters  providing the associated  window sizes.

In practice, from the observation model (\ref{om}),   to find $g_ {t}$
in (\ref{nonregr2}) minimizing the expected quadratic loss function,
or expected risk, we look for the minimizer
$\widehat{\boldsymbol{\theta }}_{T}(p)$ of the empirical regression risk
\begin{equation}
L_{T}(\widehat{\boldsymbol{\theta
}}_{T}(p))=\inf_{\boldsymbol{\theta }(p) \in \Theta^{c}}
L_{T}(\boldsymbol{\theta }(p))=\inf_{\boldsymbol{\theta }(p) \in
\Theta^{c}
}\frac{1}{T}\sum_{t=1}^{T}\left|\ln\left(\Lambda_{t}\right)(\psi_{p,\varpi_{p}})-g_{t}(\boldsymbol{\theta
}(p))\right|^{2}. \label{lse}
\end{equation}
\noindent  Truncation
parameter $N$ is then selected to controlling  the ratio between the
expected quadratic loss function at $\widehat{\boldsymbol{\theta }}_{T}(p),$ and the
expected value of the minimized  empirical risk from the identity
\begin{eqnarray}&&
E\left[\ln\left(\Lambda_{t} \right)(\psi_{p,\varpi_{p}})-
g_{t}(\psi_{p,\varpi_{p}}, \widehat{\boldsymbol{\theta
}}_{T}(p)\right]^{2}\nonumber\\
&&=E\left[L_{T}(\widehat{\boldsymbol{\theta
}}_{T}(p))\right]\left(1-\frac{N}{T}\right)^{-1}\left(1+\frac{\sum_{i=1}^{N}1/\lambda
_{i}}{T}\right), \label{eqctp}
\end{eqnarray}
\noindent  where, for $i=1,\dots,N,$  $1/\lambda _{i}$ denotes the
inverse of the $i$th eigenvalue of the matrix $\Phi^{T}\Phi,$
with $\Phi $ being a $T\times N$ matrix, whose elements are the
values of the $N$ trigonometric basis functions selected at the time
points $t=1,\dots,T.$ Parameter $N$ should be  such that $N<<T.$ Note
that, asymptotically, when $N\to \infty,$ $\Phi^{T}\Phi$ goes
to the identity matrix, and for $i=1,\dots,N,$ $1/\lambda
_{i}\sim 1.$ In equation (\ref{eqctp}), we have considered the
minimized empirical risk
\begin{equation}L_{T}(\widehat{\boldsymbol{\theta }}_{T}(p))=\frac{1}{T}\widetilde{\mathcal{R}}^{T}(p)\left(I_{T\times
T}-\Phi\left(\Phi^{T}\Phi\right)^{-1}\Phi^{T}\right)\widetilde{\mathcal{R}}(p),
\label{merv}
\end{equation}
\noindent  for each spatial region  $p=1,\dots, P,$ where
$$\widetilde{\mathcal{R}}(p)=\left(\sum_{k=N+1}^{\infty} \left(
A_{k}(p)\cos (\varphi _{k} t)+B_{k}(p)\sin (\varphi
_{k}t)\right)+ \varepsilon_{t}(\psi_{p,\varpi_{p}}),\
t=1,\dots,T\right).$$

Our regression predictor is then computed, for any $t\in \mathbb{R}_{+},$  from the identity
\begin{equation}\widehat{\ln\left(\Lambda_{t} \right)}(\psi_{p,\varpi_{p}})=g_{t}(\widehat{\boldsymbol{\theta
}}_{T}(p)),\quad p=1,\dots,P \label{prednpr}
\end{equation}
\noindent (see Theorem 1 in \cite{Ivanov15} about conditions for the
weak--consistency of (\ref{prednpr})).

 The  regression residuals
 $$\mathbf{Y}=\left\{Y_{t}(\psi_{p,\varpi_{p}})= \ln\left(\Lambda_{t}\right)(\psi_{p,\varpi_{p}})- g_{t}(\widehat{\boldsymbol{\theta }}_{T}(p)),\
  t=1,\dots,T,\ p=1,\dots, P\right\},$$\noindent and the  empirical nuclear   autocovariance and cross--covariance operators
  \begin{eqnarray}
\widehat{R}_{0,T}^{\mathbf{Y}}(\psi_{p,\varpi_{p}})(\psi_{q,\varpi_{q}})=\frac{1}{T}\sum_{t=1}^{T}Y_{t}(\psi_{p,\varpi_{p}})Y_{t}(\psi_{q,\varpi_{q}})
\nonumber\\
\widehat{R}_{1,T}^{\mathbf{Y}}(\psi_{p,\varpi_{p}})(\psi_{q,\varpi_{q}})=\frac{1}{T-1}\sum_{t=1}^{T-1}Y_{t}(\psi_{q,\varpi_{q}})Y_{t+1}(\psi_{p,\varpi_{p}}),\
p,q=1,\dots,P,\nonumber\\
\label{accop}
\end{eqnarray}
\noindent    will be considered in the
estimation of the spatial linear residual correlation  (see
\cite{Bosq2000}).  A truncation parameter
$k(T)$ is also considered here to remove the ill--posed nature of this estimation problem. Particularly, $k(T)$ must satisfy   $k(T)\to \infty,$ $k(T)/T\to 0,$ $T\to
\infty .$ A suitable choice of $k(T)$ also ensures strong--consistency of the estimator
\begin{eqnarray}
  &&\widehat{\rho}_{k(T)}(\psi_{p,\varpi_{p}})(\psi_{q,\varpi_{q}})=
  \sum_{k,l=1}^{k(T)}\frac{\left\langle \psi_{p,\varpi_{p}},\phi_{k,T} \right\rangle_{H}\left\langle   \psi_{q,\varpi_{q}},
   \phi_{l,T} \right\rangle_{H}}{\lambda_{k,T}(\widehat{R}_{0,T}^{\mathbf{Y}})}
    \widehat{R}_{1,T}^{\mathbf{Y}}(\phi_{k,T})(\phi_{l,T}), \nonumber\\ \label{cerho}
\end{eqnarray}
\noindent for
$p,q=1,\dots,P$ (see \cite{Bosq2000}).
 Here,
\begin{eqnarray}
\widehat{R}_{0,T}^{\mathbf{Y}} &=&
\sum_{k=1}^{T}\lambda_{k,T}(\widehat{R}_{0,T}^{\mathbf{Y}})[\phi_{k,T}\otimes
\phi_{k,T}], \label{covop}
\end{eqnarray}
\noindent where $\{\lambda_{k,T}(\widehat{R}_{0,T}^{\mathbf{Y}}),\
k=1,\dots,T\}$ and  $\left\{\phi_{k,T}, \ k\geq 1\right\}$ denote
the  empirical eigenvalues and eigenvectors of
$\widehat{R}_{0,T}^{\mathbf{Y}},$ respectively. Particularly,  we consider  $k(T)=\ln(T)$  (see \cite{Bosq2000}).
 The  classical plug--in predictor is then computed, for each $p=1,\dots,P,$ as   \begin{equation}
 \widehat{Y}_{t}^{k(T)}(\psi_{p,\varpi_{p}})=\sum_{q=1}^{P}\widehat{\rho}_{k(T)}(\psi_{q,\varpi_{q}})(\psi_{p,\varpi_{p}})
 Y_{t-1}(\psi_{q,\varpi_{q}}),\quad t\geq 1.
 \label{ppARH1}
 \end{equation}

  Under the  Gaussian distribution of $\nu_{t},$  in the  Bayesian  estimation of $\rho,$
    from (\ref{arh1}),  the likelihood function, defining the objective function,   is given by, for each  $p=1,\dots,P,$ \begin{eqnarray}&&
\widetilde{L}_p(\varepsilon_{1p},\dots,\varepsilon_{Tp},\varepsilon_{0q},\dots,\varepsilon_{(T-1)q}\rho (\psi_{q,\varpi_{q}})(\psi_{p,\varpi_{p}}),q=1,\dots,P)\nonumber\\
&&= \frac{\exp \left( -\frac{1}{2\sigma^2_{p}}\sum_{t=1}^T\left(\varepsilon_{t}(\psi_{p,\varpi_{p}})-\sum_{q=1}^{P}\varepsilon_{t-1}(\psi_{q,\varpi_{q}})\rho (\psi_{q,\varpi_{q}})(\psi_{p,\varpi_{p}}) \right)^2 \right)}{\left(\sigma_{p}\sqrt{2\pi}\right)^T}\nonumber\\
& &\times\prod_{q=1}^{P} \left[\rho (\psi_{q,\varpi_{q}})(\psi_{p,\varpi_{p}})\right]^{a_{pq}-1}\left(1-\rho (\psi_{q,\varpi_{q}})(\psi_{p,\varpi_{p}})\right)^{b_{pq}-1}\nonumber\\ &&\hspace*{2.5cm}\times \frac{\mathbb{I}_{\{0<\rho (\psi_{q,\varpi_{q}})(\psi_{p,\varpi_{p}})<1\}}}{\mathbb{B}(a_{pq}, b_{pq})}\nonumber\\
&&= \frac{1}{\left(\sigma_{p}\sqrt{2\pi}\right)^T} \exp \left( -\frac{1}{2\sigma^2_{p}}\sum_{t=1}^T\left[\nu_{t}(\psi_{p,\varpi_{p}})\right]^2 \right)\nonumber\\
& &\hspace*{2cm}\times\prod_{q=1}^{P} \left[\rho (\psi_{q,\varpi_{q}})(\psi_{p,\varpi_{p}})\right]^{a_{pq}-1}\left(1-\rho (\psi_{q,\varpi_{q}})(\psi_{p,\varpi_{p}})\right)^{b_{pq}-1}\nonumber\\ &&\hspace*{2.5cm}\times \frac{\mathbb{I}_{\{0<\rho (\psi_{q,\varpi_{q}})(\psi_{p,\varpi_{p}})<1\}}}{\mathbb{B}(a_{pq}, b_{pq})},\nonumber\\
\label{eqbayest}
\end{eqnarray}
 \noindent where, for each $p=1,\dots,P,$ the  beta probability distributions with  shape
    parameters $a_{pq}$ and  $ b_{pq},$ $q=1,\dots,P,$ respectively define the prior probability distributions of the   independent random variables $\{\rho (\psi_{q,\varpi_{q}})(\psi_{p,\varpi_{p}}), \ q=1,\dots,P\}.$
    Here, for  each $p=1,\dots,P,$
   $\varepsilon_{tp}=\varepsilon_{t}(\psi_{p,\varpi_{p}})=\left\langle \varepsilon_{t},\psi_{p,\varpi_{p}}\right\rangle_{H},$ and
    $\sigma_{p} = \sqrt{E[\varepsilon_t(\psi_{p,\varpi_{p}})]^2},$  for $t=0,\dots,T.$ As before, $\psi_{p,\varpi_{p}}$  weights
    the spatial sample  information about the $p$ small--area, for $p=1,\dots,P.$ As usual, $\mathbb{I}_{0<\cdot<1}$
    denotes the indicator  function on the interval $(0,1),$ and   $\mathbb{B}(a_{pq}, b_{pq})$ is the beta
    function,
    $$ \mathbb{B}(a_{pq}, b_{pq})=\frac{\Gamma(a_{pq})\Gamma(b_{pq})}{\Gamma(a_{pq}+b_{pq})}.$$
    From (\ref{eqbayest}),  the Bayesian predictor is obtained, for
 $p= 1,\dots,P,$ as
\begin{equation}
\widetilde{\varepsilon}_{t}(\psi_{p,\varpi_{p}})=\sum_{q=1}^{P}\widetilde{\rho}
(\psi_{q,\varpi_{q}})(\psi_{p,\varpi_{p}})\varepsilon_{t-1}(\psi_{q,\varpi_{q}}),\quad t\geq 1,\label{brhoest}
\end{equation}
\noindent with  $\left(\widetilde{\rho}
(\psi_{1,\varpi_{1}})(\psi_{p,\varpi_{p}}),\dots,\widetilde{\rho}
(\psi_{P,\varpi_{P}})(\psi_{p,\varpi_{p}}) \right)$ being computed by
maximizing  (\ref{eqbayest}), to find the posterior mode (see
\cite{Bosq14}, where Bayesian estimation is introduced in an
infinite--dimensional framework). We refer to (\ref{brhoest}) as the
Bayesian plug--in predictor of the residual mortality   log--risk
process at the $p$ small area, for $p= 1,\dots,P.$ In practice,
equation (\ref{eqbayest}) is approximated from the computed values
of the regression  residual   process.

\section{Statistical   analysis of COVID--19 mortality  }
\label{sac19}

 Our analysis is based on  daily records of
 COVID--19 mortality  reported by the Spanish
Statistical National Institute, since
 March, 8 to May, 13, 2020, at the 17 Spanish Communities.
 We first describe the main steps of the proposed estimation algorithm, referring to
  the inputs and outputs at  different stages.
 \begin{itemize}
 \item[Step 1]  Daily records of
 COVID--19 mortality   are accumulated over the entire period  at every Spanish Community.  The resulting step cumulative curves are
  interpolated  at $265$ temporal nodes, and cubic B--spline smoothed. Their derivatives and logarithmic transforms are then computed.
 \item[Step 2] Our soft--input--data process is obtained from the spatial projection of the  outputs in Step 1 onto the compactly supported basis  $\{\psi_{p,\varpi_{p}},\ p=1,\dots,P=17\}.$ We choose  the tensorial product of Daubechies wavelet bases. Here,  for $p=1,\dots,17,$    $\varpi_{p}=(N(p), j(p),\mathbf{k}(p)),$ whose components respectively provide the order of the Daubechies wavelet functions,  the resolution level, and the vector of spatial displacements, according to the area occupied by  each Spanish community (see, e.g., \cite{Daubechies88}).
     \item[Step 3] The choice   $N=6$ in   (\ref{eqctp}) corresponds to  $1.1304$  value of the  ratio between the mean quadratic loss function and  expected minimized  empirical risk.  Hence $12$ coefficients should be estimated.  Note that the eigenvalues  in (\ref{eqctp}) are computed from the trigonometric basis.
         \item[Step 4] Under   $N=6$ in Step 3, the least--squares estimates of the $12$ Fourier coefficients are computed from  (\ref{lse}),
         in terms of  the
         soft--input--data process  obtained as output in Step 2.
             \item[Step 5] The regression residuals are then calculated  from Step 4.
             \item[Step 6] The auto-- and cross-- covariance operators in  (\ref{accop}) are computed from the outputs of Step 5. The residual spatial linear correlation matrix is then obtained from (\ref{cerho}). The truncation scheme  $k(T)=\ln(T)$ has been adopted,  with $T=265.$
             \item[Step 7]  The residual predictor (\ref{ppARH1}) is computed from Step 6.
                 \item[Step 8]  100 bootstrap samples are generated  from the empirical  autocorrelation projections. The bootstrap   prior fitted suggests us to consider  a scaled beta probability density with shape parameters $14$ and $13.$
                 \item[Step 9] Assuming a Gaussian scenario for our log--regression residuals, our constrained nonlinear multivariate  objective function (\ref{eqbayest}) is computed from the prior proposed in Step 8.
                     \item[Step 10] The maximize the objective function  computed in Step 9,  we implement an hybrid genetic algorithm, constructed from 'gaoptimset' MaLab function, implemented with the 'HybridFcn' option that handles to a function to continuing  optimization after the genetic algorithm terminates. This last function applies quasi-Newton methodology  in the optimization procedure,  involving an inverse Hessian matrix  estimate.
                         \item[Step 11] The soft--data based  bayesian predictor (\ref{brhoest}) of the residual COVID--19 mortality log--risk is finally computed from the outputs in Step 10.
                             \item[Step 12] Our multiple objective space--time predictor is obtained from  Steps 4 and 11, by addition the regression and residual predictors, applying inverse spatial wavelet transform.
 \end{itemize}

 Tables \ref{tb1}--\ref{tb2} below display the parameter
 estimates  $\widehat{A}_{k}(\cdot)$ and $\widehat{B}_{k}(\cdot),$
$k=1,\dots,6,$ where $\varphi_{k}=\frac{2\pi}{265}$ has been
considered, for $k=1,\dots, N=6.$ In these tables and below, the
following  Spanish Community (SC) codes appear: C1 for
Andaluc\'{\i}a; C2 for Arag\'on; C3 for Asturias; C4 for Islas
Baleares; C5 for Canarias; C6 for Cantabria; C7 for Castilla La
Mancha; C8 for Castilla y Le\'on; C9 for Catalu\~na; C10 for
Comunidad Valenciana; C11 for Extremadura; C12 for Galicia;  C13 for
Comunidad de Madrid; C14 for Murcia; C15 for Navarra; C16 for
Pa\'{\i}s Vasco, and C17 for La Rioja.

\begin{table}[!h]
\caption{Regression parameter estimates $\widehat{A}_k(\cdot),$ $k =
1, \ldots, 6, $ at the 17 Spanish Communities} \label{tb1}
\begin{center}
\begin{tabular}{|ccccccc|}
 \hline
 {\bf SC/PE} &  $\widehat{A}_1(\cdot) $ & $\widehat{A}_2(\cdot) $ & $\widehat{A}_3(\cdot) $ & $\widehat{A}_4(\cdot) $ &
 {\bf $\widehat{A}_5(\cdot) $} & $\widehat{A}_6(\cdot) $ \\
  \hline
C1 & 3.6343 & -0.4814 & -0.0075 & -0.0258 & 0.0189 & 0.0193 \\
C2 & 3.4345 & -0.3923 & 0.0416 & 0.0265 & -0.0709 & -0.0572 \\
C3 & 3.2031 & -0.1364 & -0.0088 & 0.0221 & 0.0430 & 0.0289 \\
C4 & 3.1445 & -0.1118 & 0.0041 & 0.0337 & 0.0062 & 0.0072 \\
C5 & 3.1015 & -0.0693 & -0.0345 & 0.0352 & -0.0112 & 0.0003 \\
C6 & 3.1347 & -0.1397 & 0.0020 & 0.0300 & -0.0061 & -0.0002 \\
C7 & 4.0591 & -0.5487 & -0.0907 & 0.0951 & 0.0992 & 0.0842 \\
C8 & 3.8032 & -0.5500 & -0.1007 & 0.0633 & 0.0139 & 0.0277 \\
C9 & 4.5095 & -0.7435 & -0.1134 & 0.1809 & 0.2231 & 0.2026 \\
C10 & 3.6321 & -0.4685 & -0.0540 & 0.0384 & -0.0152 & 0.0011 \\
C11 & 3.2967 & -0.2274 & -0.0083 & 0.0553 & 0.0250 & 0.0240 \\
C12 & 3.3454 & -0.2122 & -0.0927 & -0.0330 & 0.0724 & 0.0679 \\
C13 & 4.8419 & -0.6790 & -0.2455 & 0.0311 & 0.0554 & 0.0667 \\
C14 & 3.0941 & -0.1037 & 0.0210 & 0.0141 & -0.0016 & 0.0041 \\
C15 & 3.2877 & -0.2598 & -0.0524 & 0.0842 & -0.0423 & -0.0348 \\
C16 & 3.6870 & -0.4302 & -0.0086 & 0.0078 & -0.0027 & -0.0017 \\
C17 & 3.2197 & -0.2071 & 0.0162 & 0.0079 & 0.0206 & 0.0110 \\
  \hline
\end{tabular}
\end{center}
\end{table}

\begin{table}[!h]
\caption{Regression parameter estimates $\widehat{B}_k(\cdot),$ $k =
1, \ldots, 6, $  at the 17 Spanish Communities} \label{tb2}
\begin{center}
\begin{tabular}{|ccccccc|}
 \hline
 {\bf SC/PE} & {\bf $ \widehat{B}_1(\cdot ) $} & {\bf $ \widehat{B}_2(\cdot ) $} & {\bf $ \widehat{B}_3(\cdot ) $} & {\bf $ \widehat{B}_4(\cdot ) $} & {\bf $ \widehat{B}_5(\cdot ) $} & {\bf $ \widehat{B}_6(\cdot ) $} \\
  \hline
C1 & 0 & -0.0052 & -0.1330 & -0.0123 &  0.0064 & -0.0195 \\
C2 & 0 & -0.0367 & -0.0998 & -0.0462 & -0.0343 & -0.0107 \\
C3 & 0 & -0.0531 & -0.0074 & -0.0142 & -0.0003 &  0.0020 \\
C4 & 0 & -0.0074 & -0.0284 & -0.0151 & -0.0092 &  0.0012 \\
C5 & 0 &  0.0433 & -0.0438 & -0.0116 & -0.0118 &  0.0046 \\
C6 & 0 &  0.0018 & -0.0174 & -0.0068 & -0.0089 &  0.0000 \\
C7 & 0 & -0.0365 & -0.2451 & -0.1791 & -0.0820 &  0.0026 \\
C8 & 0 &  0.0953 & -0.2389 & -0.0431 & -0.0313 & -0.0045 \\
C9 & 0 & -0.1587 & -0.4054 & -0.2269 & -0.1010 &  0.0047 \\
C10 & 0 &  0.1118 & -0.1579 & -0.0458 & -0.0418 & -0.0220 \\
C11 & 0 &  0.0754 & -0.1138 & -0.0166 & -0.0048 &  0.0072 \\
C12 & 0 & -0.1104 & -0.1338 &  0.1330 &  0.0761 & -0.0017 \\
C13 & 0 &  0.4654 & -0.1302 & -0.1602 & -0.1061 & -0.0038 \\
C14 & 0 &  0.0355 & -0.0560 &  0.0119 &  0.0025 & -0.0044 \\
C15 & 0 & -0.0187 & -0.0021 & -0.0897 & -0.0562 &  0.0134 \\
C16 & 0 &  0.0025 & -0.0707 & -0.0638 & -0.0439 & -0.0267 \\
C17 & 0 &  0.0389 & -0.0270 & -0.0174 & -0.0006 &  0.0019 \\
  \hline
\end{tabular}
\end{center}
\end{table}

Bootstrap curve confidence intervals at confidence level $1-\alpha
=0.95$, based on $1000$ bootstrap samples,  are computed for the
spatial mean, over the $17$ Spanish Communities, of the curve regression predictors. Their construction is based on
the bias corrected and accelerated percentile
method ($\mathcal{I}_{1}$); Normal approximated interval with
bootstrapped bias and standard error ($\mathcal{I}_{2}$); basic
percentile method ($\mathcal{I}_{3}$), and  bias corrected
percentile method ($\mathcal{I}_{4}$) (see Figure \ref{erpci}
below). The minimized  regression  empirical  risk values
$L_{265}(\widehat{\boldsymbol{\theta }}_{265}(p)),$ $p=1,\dots,17,$
are displayed in Table \ref{tb2b}.

    \begin{center}
 \begin{figure}[!h]
     \includegraphics[width=5cm,height=5cm]{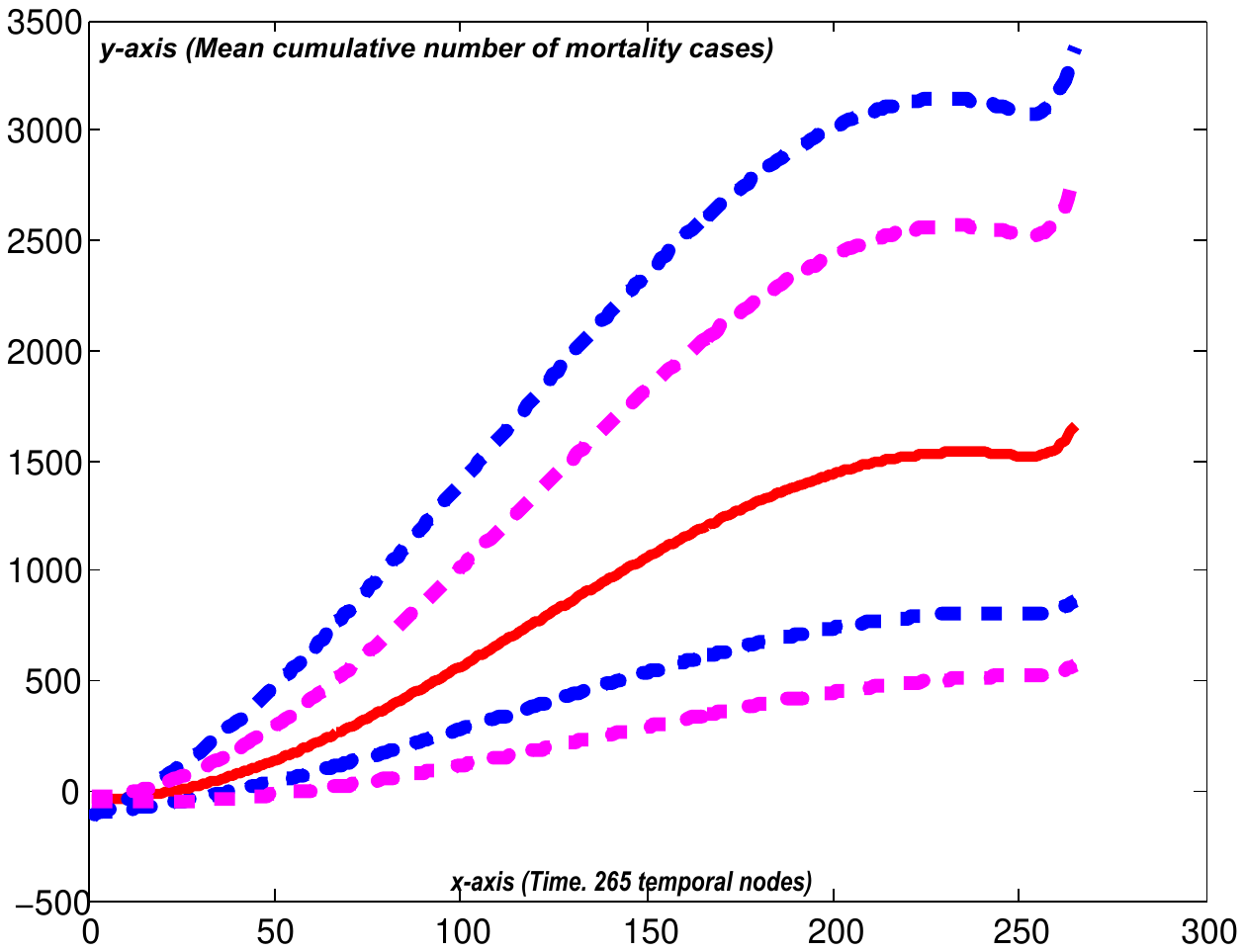}\hspace*{0.5cm}
     \includegraphics[width=5cm,height=5cm]{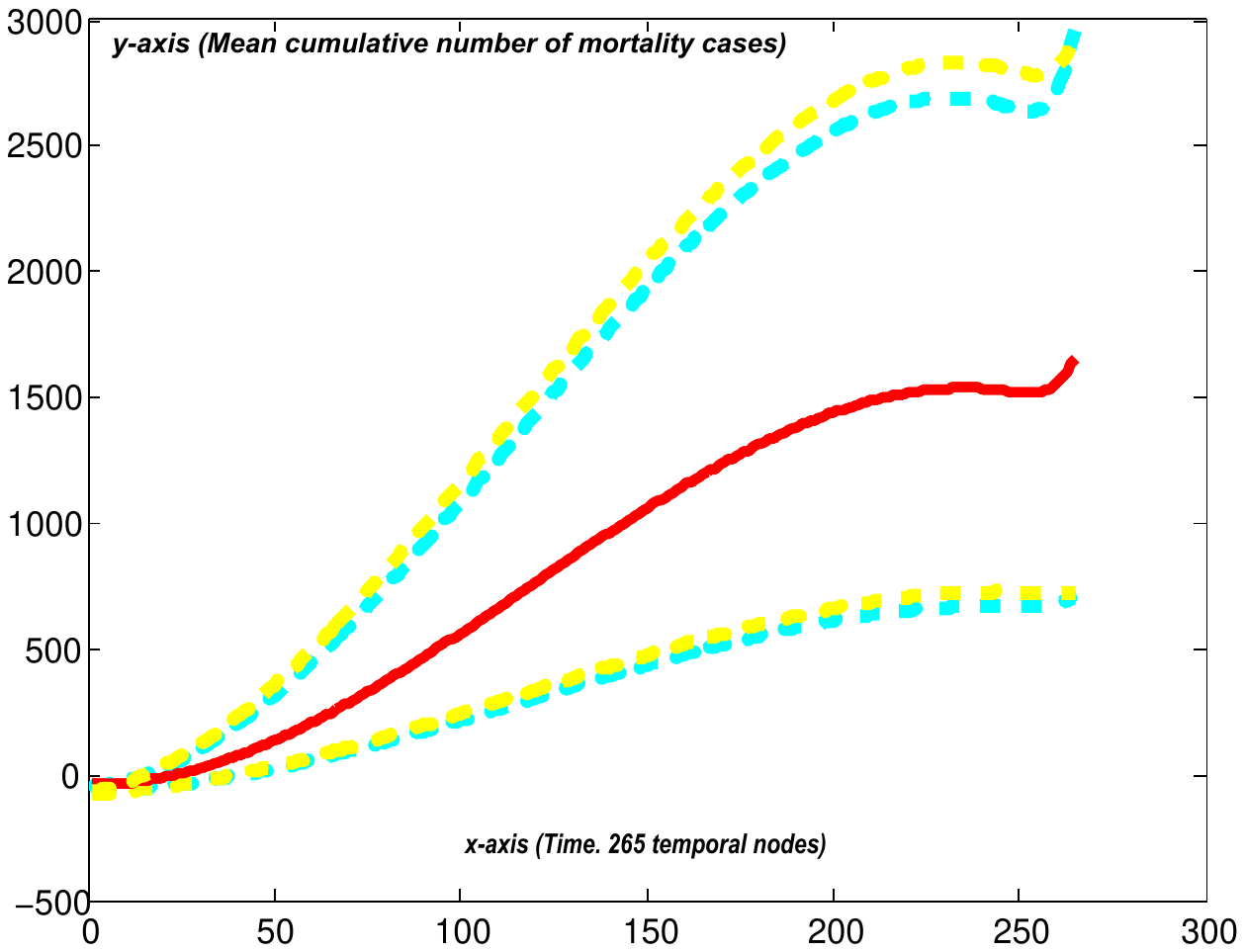}
     \includegraphics[width=5cm,height=5cm]{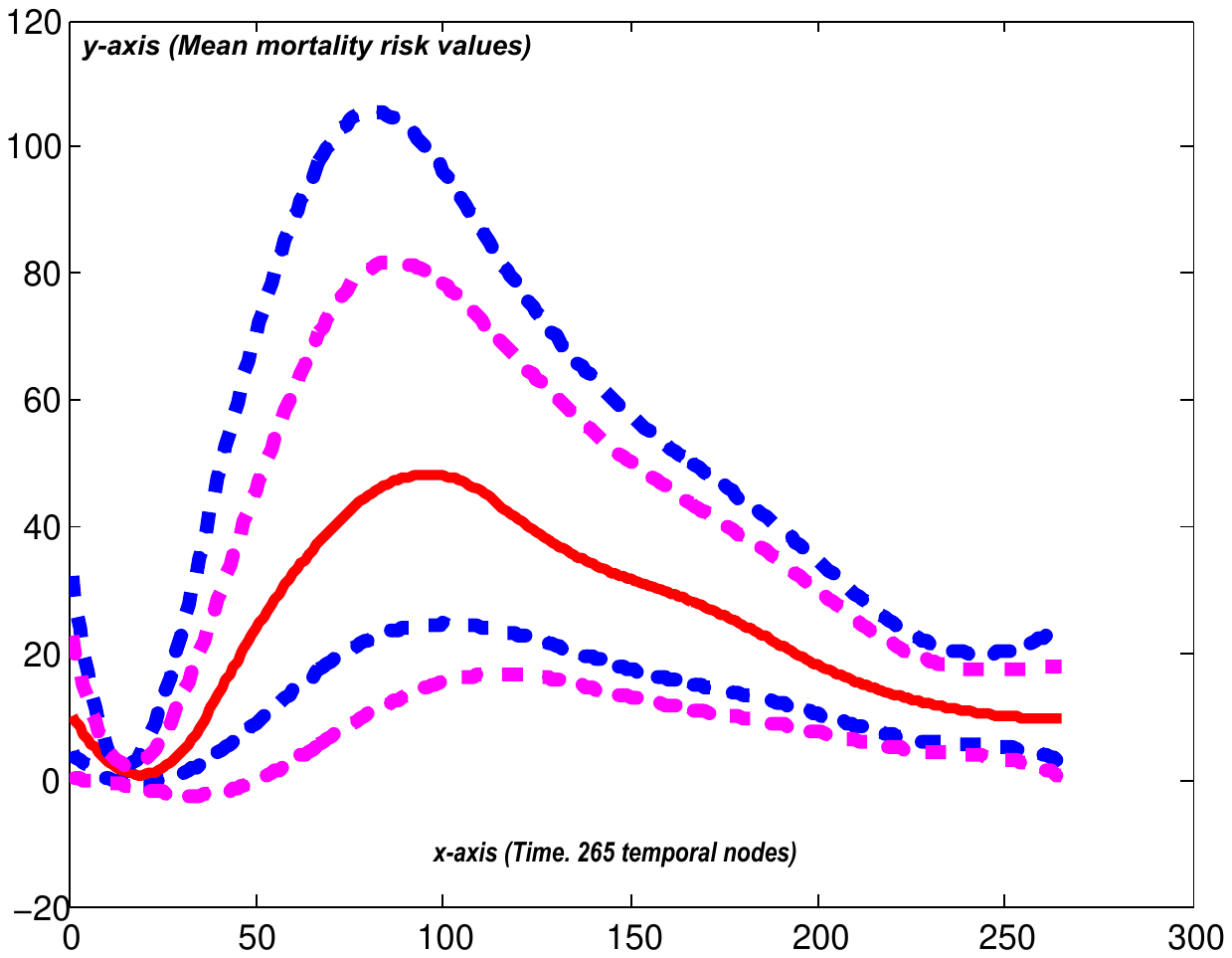}\hspace*{0.5cm}
     \includegraphics[width=5cm,height=5cm]{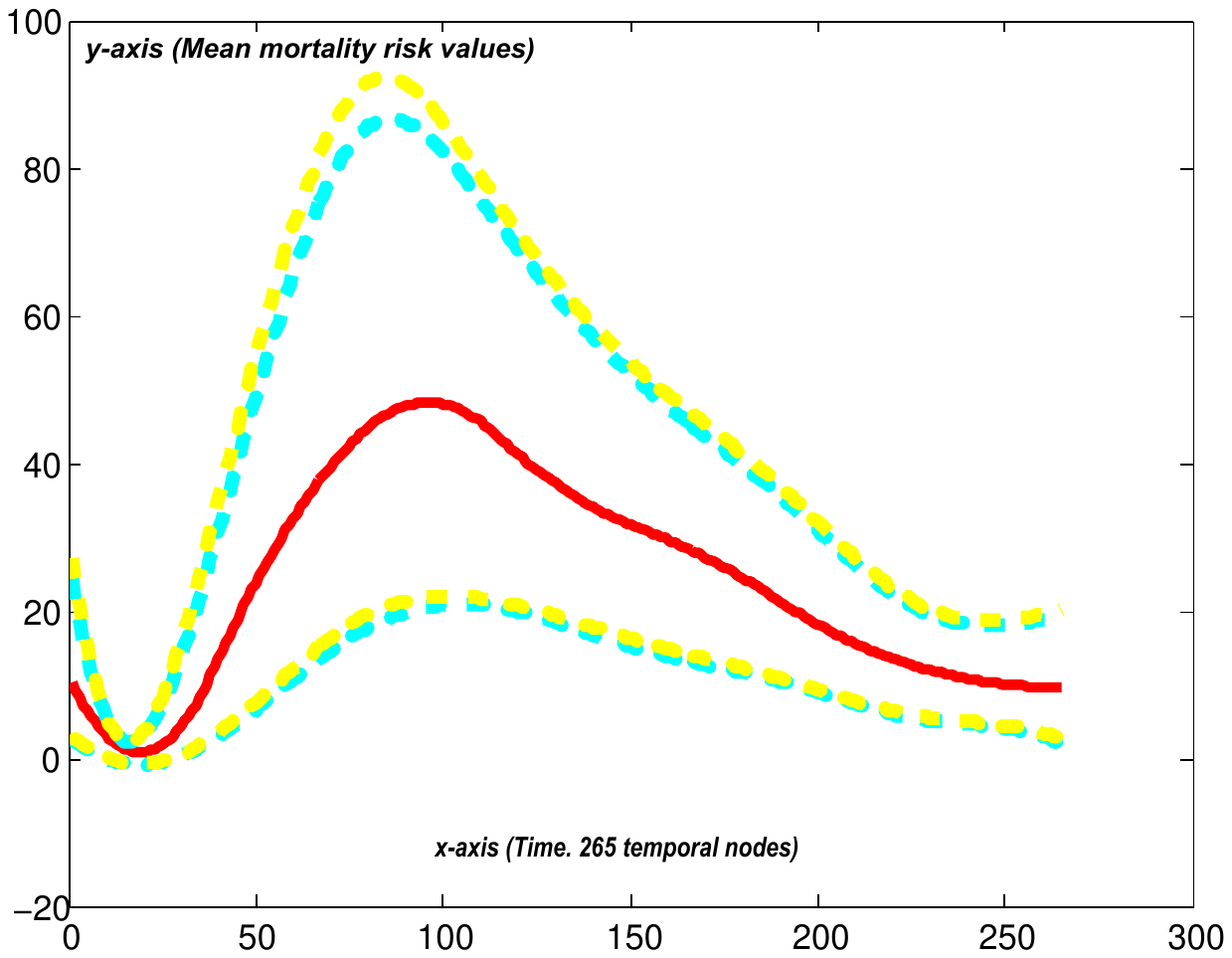}
     \includegraphics[width=5cm,height=5cm]{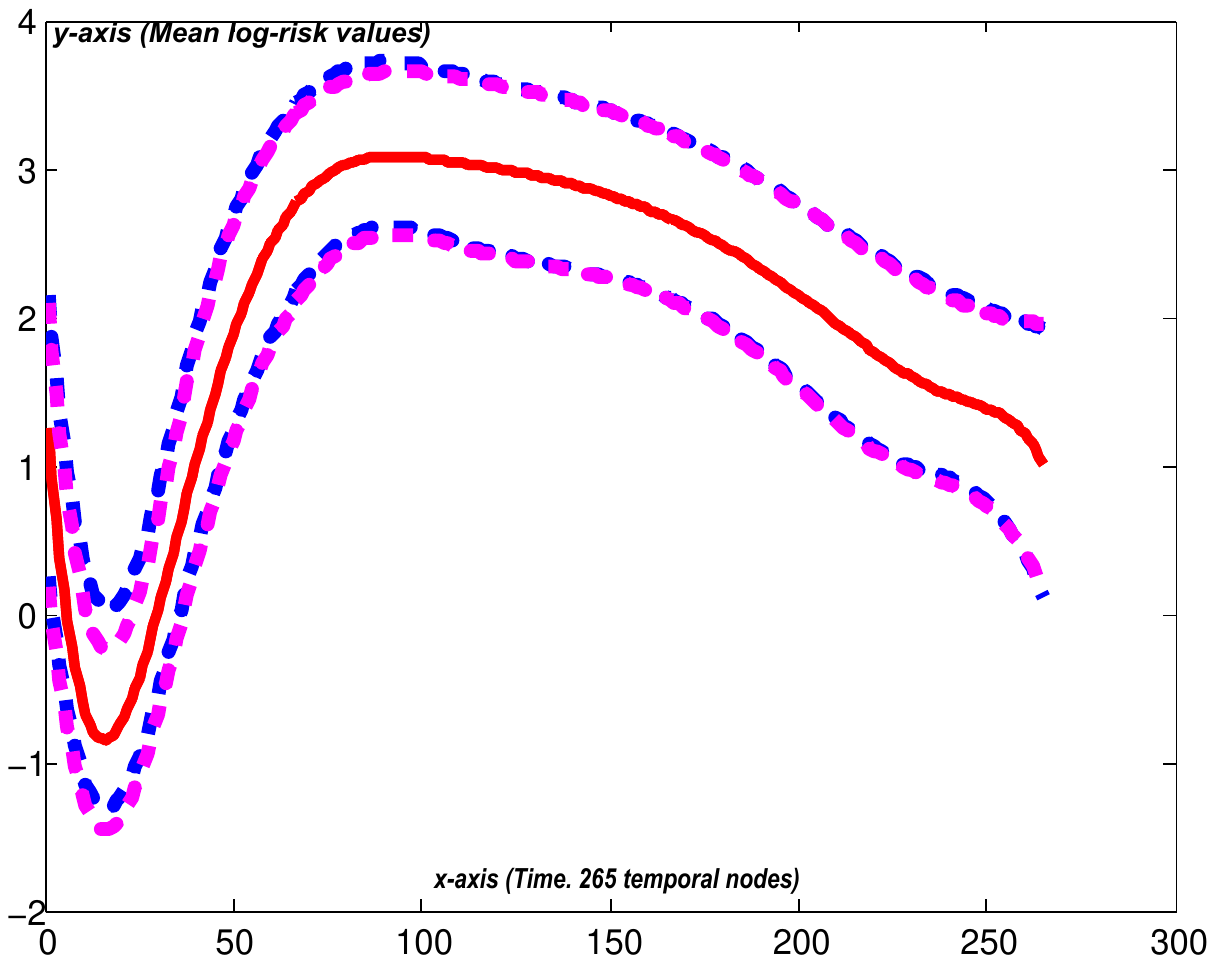}\hspace*{3.5cm}
     \includegraphics[width=5cm,height=5cm]{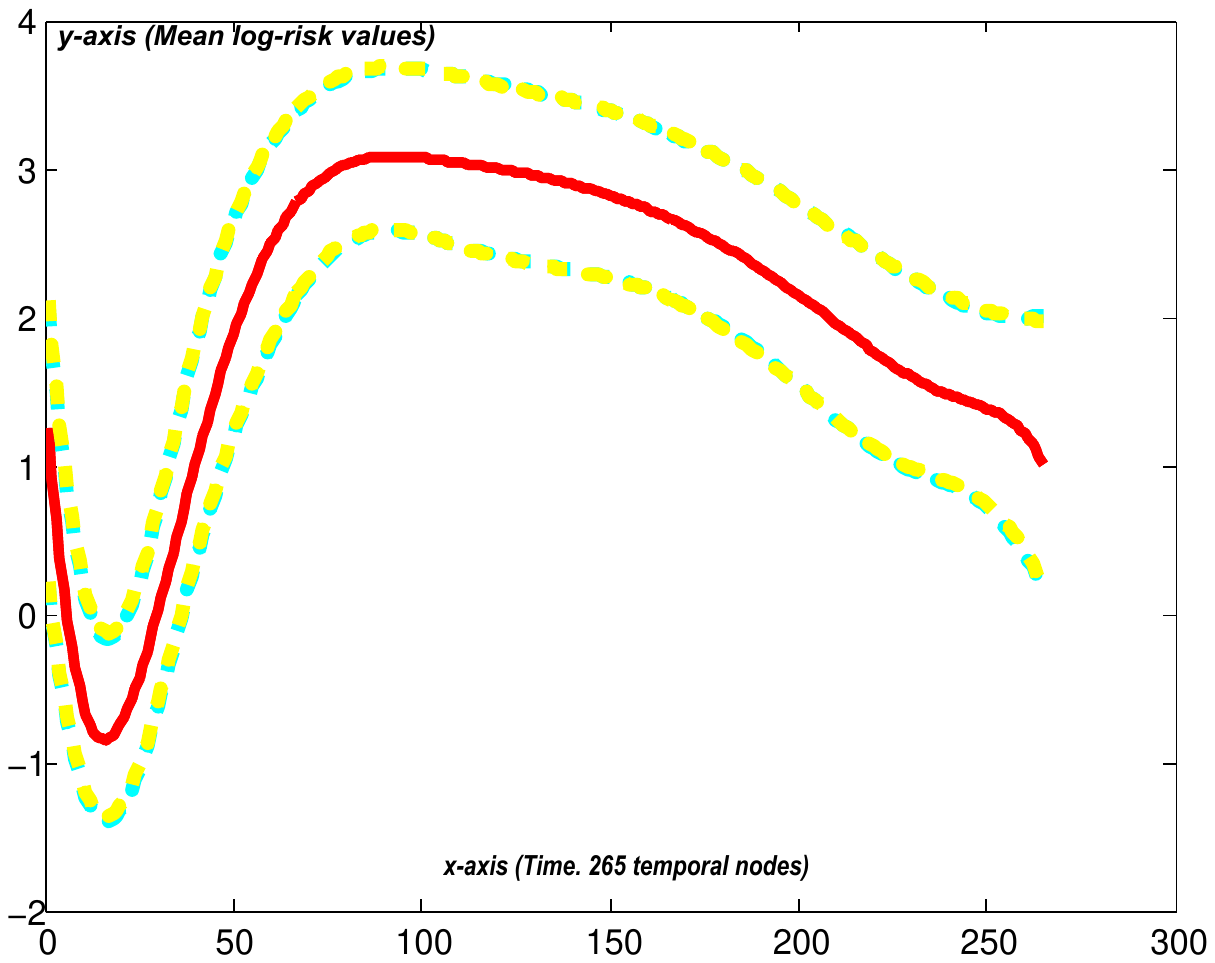}
 \caption{At the top,  COVID--19 mortality  mean cumulative  curve in Spain, since March, 8, 2020 to
 May, 13, 2020  (continuous
 red line, 265 temporal nodes), and  bootstrap curve confidence
 intervals, at the left--hand--side,
  $\mathcal{I}_{1}$
 (dashed blue lines) and  $\mathcal{I}_{2}$
  (dashed  magenta
 lines), and  at the right--hand--side,   $\mathcal{I}_{3}$ (dashed  green lines)
 and $\mathcal{I}_{4}$ (dashed yellow lines). Plots at the center and bottom reflect the same information
respectively  referred to the mean intensity (spatial  averaged COVID--19 mortality risk curve), and log--intensity (spatial averaged COVID--19 mortality log--risk curve) curves in Spain.
 All the confidence bootstrap intervals are computed at confidence level $1-\alpha =0.95,$ from $1000$ bootstrap samples
 }
 \label{erpci}
 \end{figure}
\end{center}
\clearpage

\begin{table}[!h]
\caption{Computed values $L_{265}(\widehat{\boldsymbol{\theta
}}_{265}(p)),$ $p=1,\dots,17$} \label{tb2b}
\begin{center}
\begin{tabular}{ccccc}
 \hline
   $L_{265}(\widehat{\boldsymbol{\theta }}_{265}(p))$ & $p=$ & $1\dots 17$ &
    &
  \\
  \hline
 0.0155  &   0.0259 &   0.0668 &   0.0408  &   0.0927\\
0.0623  & 0.1642  & 0.0883  &   0.2174  &  0.0313\\     0.0559 &
0.1904
& 0.0054   & 0.1602 & 0.1640   \\  0.0003  &    0.1238 & & & \\
\hline
\end{tabular}
\end{center}
\end{table}

\begin{center}
 \begin{figure}[!h]
 \centering
 \includegraphics[width=8.5cm,height=4.5cm]{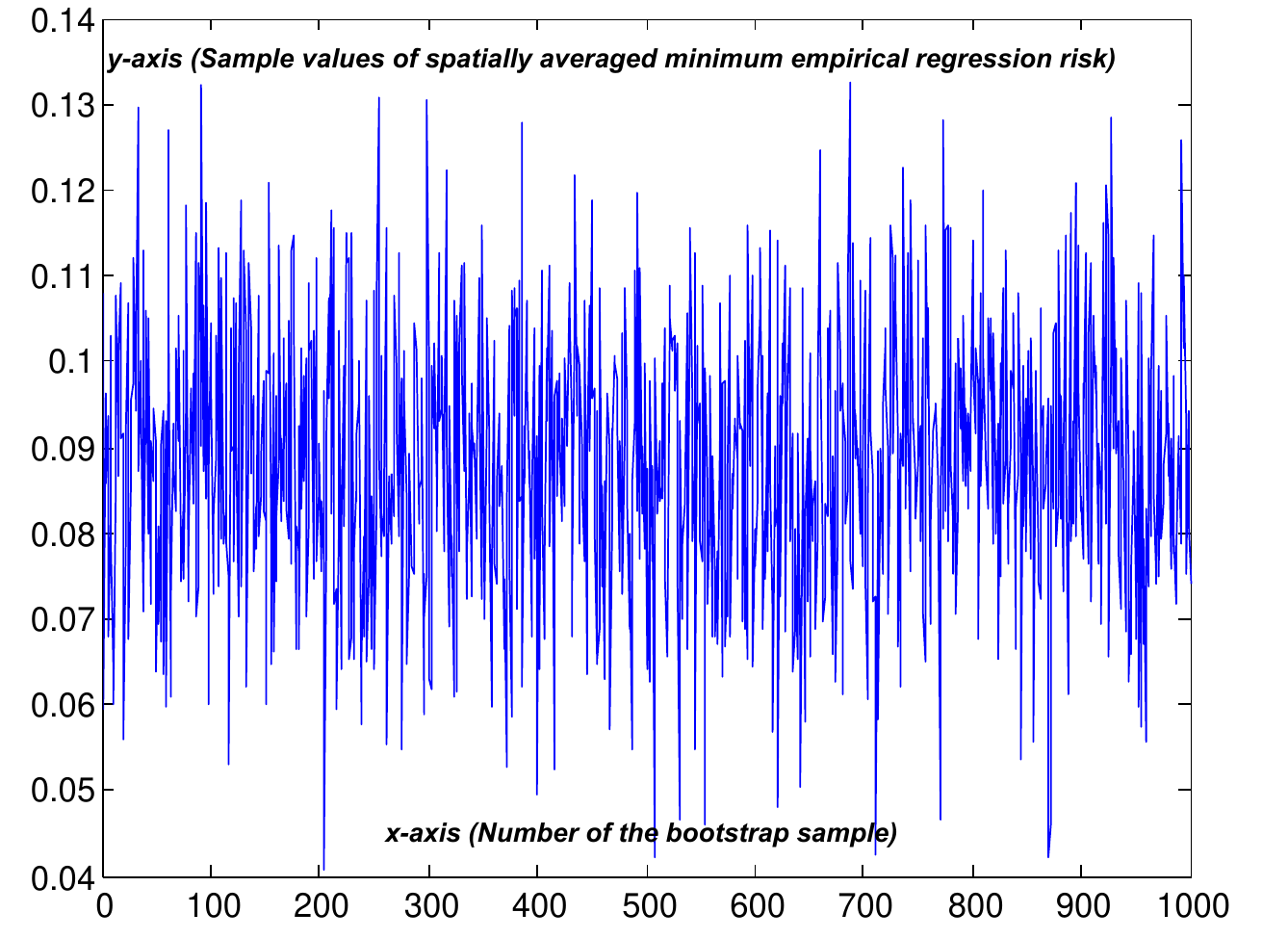}

 \vspace*{1.5cm}

   \includegraphics[width=4.5cm,height=4.5cm]{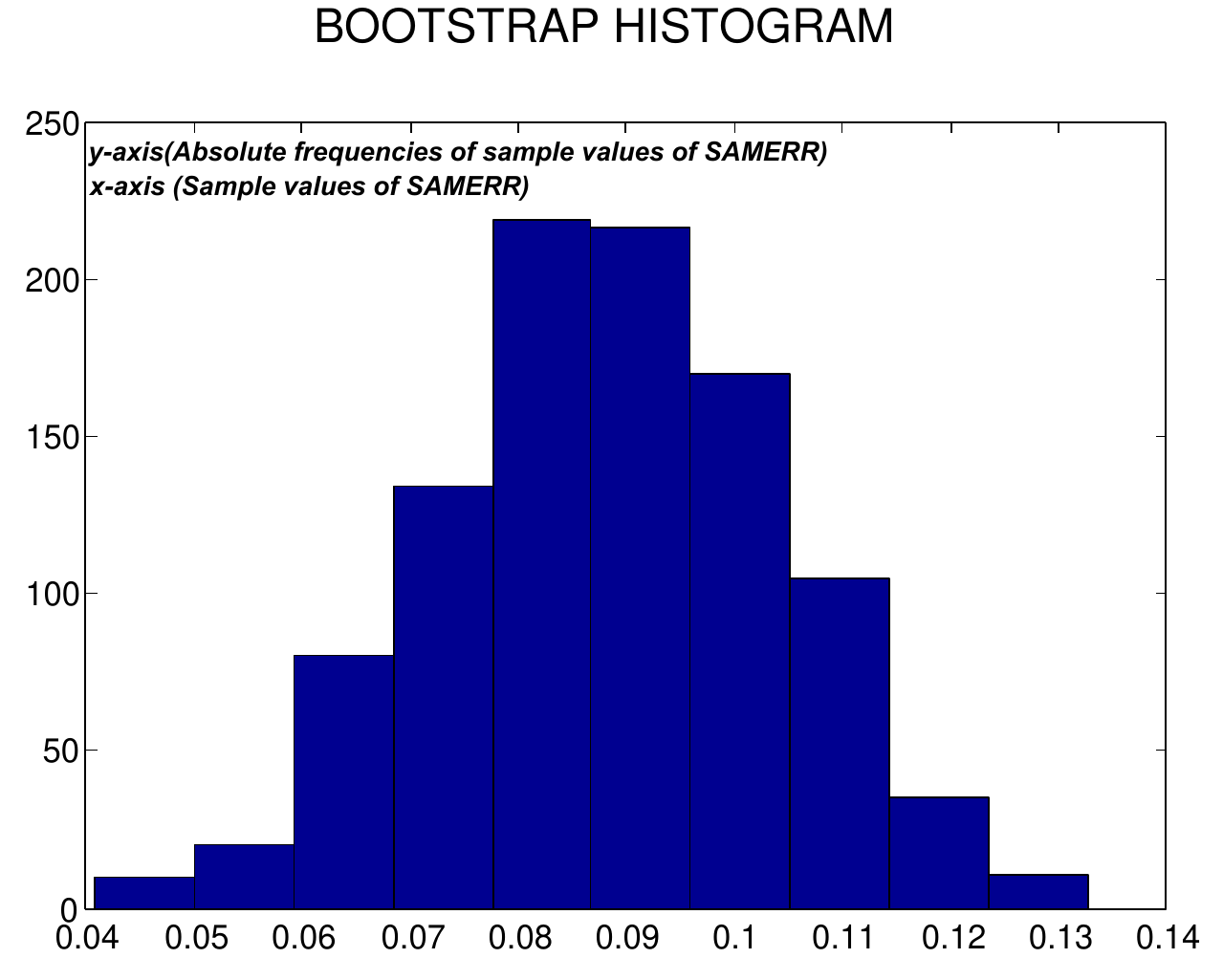}\hspace*{0.75cm}
        \includegraphics[width=4.5cm,height=4.5cm]{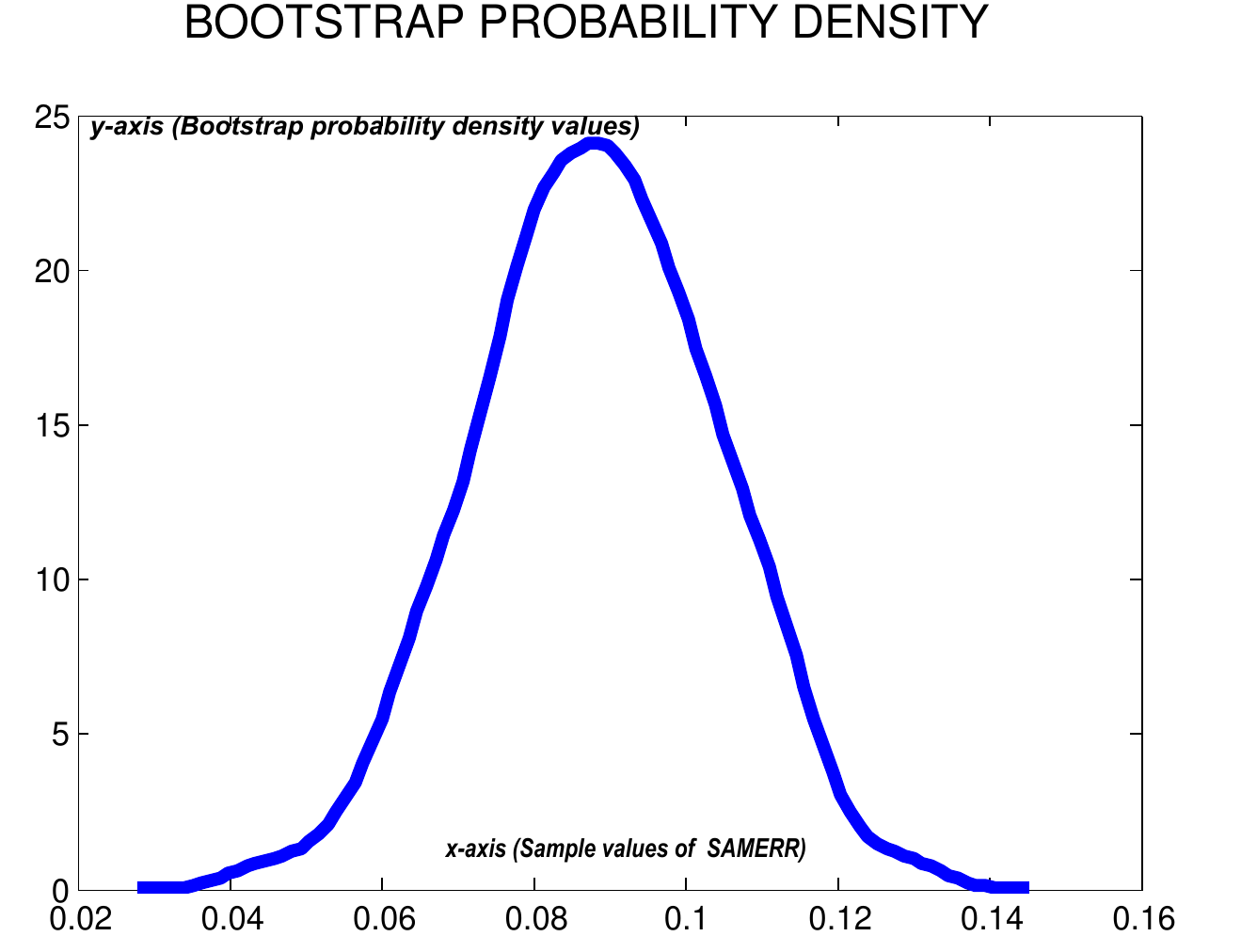}
      \caption{$1000$ bootstrap samples  have been generated of the spatially averaged
 minimum empirical  regression  risk (SAMERR). The corresponding sample values are displayed at the top.
 The bootstrap histogram can be found at the bottom--left--hand side. The bootstrap probability density is plotted at the bottom--right--hand--side}
 \label{erphisto}
 \end{figure}
\end{center}

\noindent  Figure \ref{erphisto}  at the top displays the $1000$ bootstrap  sample
values
 $$\overline{L_{265}}(\omega_{i})=\frac{1}{P}
 \sum_{p=1}^{P}L_{265}(\omega_{i},\widehat{\boldsymbol{\theta }}_{265}(p)),
 \quad \omega_{i}\in \Omega, \ i=1,\dots,1000,$$
\noindent of the
 spatial averaged minimized    empirical quadratic  risk in the trigonometric regression. Note that the sample mean
 of these values is
 $\overline{\overline{L}}= 0.0262,$ showing a good performance of
 the least--squares regression predictor, according to the value
 $T(265,12)= 1.1304$ obtained. The bootstrap histogram and the
 corresponding approximation of the probability density function,
 computed from
 $\overline{L_{265}}(\omega_{i}),$ $i=1,\dots, 1000,$
 are also plotted at the bottom  of Figure \ref{erphisto}.

Bootstrap confidence intervals for $\overline{L_{265}}$ have also
been computed at level $1-\alpha =0.95,$ from 1000 and 10000
bootstrap samples.  Table \ref{tbbi} displays these intervals respectively based on the bias corrected and accelerated
percentile method ($\mathcal{I}_{1}$); Normal approximated interval
with bootstrapped bias and standard error ($\mathcal{I}_{2}$); basic
percentile method ($\mathcal{I}_{3}$); bias corrected percentile
method ($\mathcal{I}_{4}$), and Student--based confidence interval
($\mathcal{I}_{5}$).
\begin{table}[!h]
\caption{Bootstrap confidence intervals for $\overline{L_{265}}$
(confidence level $1-\alpha =0.95$)}
\begin{center}
\begin{tabular}{ccc}
 \hline
CI/S & 1000 & 10000\\
$\mathcal{I}_{1}$ & $[0.0593,
    0.1222]$ & [0.0594,
    0.1236]\\
$\mathcal{I}_{2}$ & [0.0564,
    0.1196] & [0.0567,
    0.1207]
\\
$\mathcal{I}_{3}$
 & [0.0584,
    0.1215] & [0.0579,
    0.1217]\\
 $\mathcal{I}_{4}$ & [0.0592,
    0.1233]& [0.0581,
    0.1208]\\
 $\mathcal{I}_{5}$ & [0.0484,
    0.1281]  & [0.0494,
    0.1215]\\
  \hline
  \label{tbbi}
\end{tabular}
\end{center}
\end{table}

The classical  and Bayesian  plug--in predictors of the residual COVID--19
mortality     log--risk  process at each one of the Spanish
Communities are  respectively  computed from equations
(\ref{ppARH1}) and (\ref{brhoest})  for $P=17$.

\begin{center}
 \begin{figure}[!h]
 \centering
 \hspace*{-2cm}
 \vspace*{-4cm}
  \includegraphics[width=8cm,height=12cm]{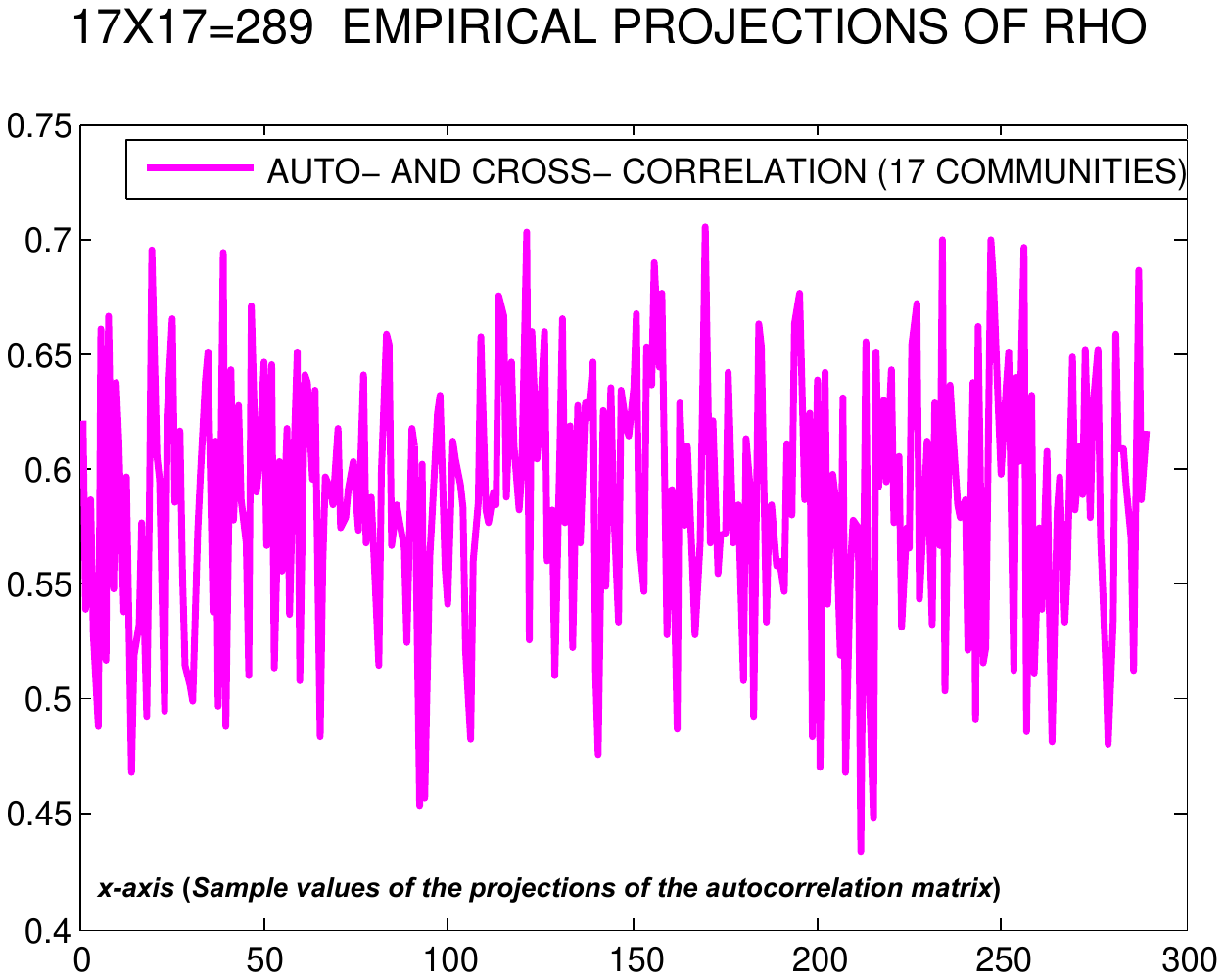}
   \hspace*{-2cm}
  \includegraphics[width=8cm,height=12cm]{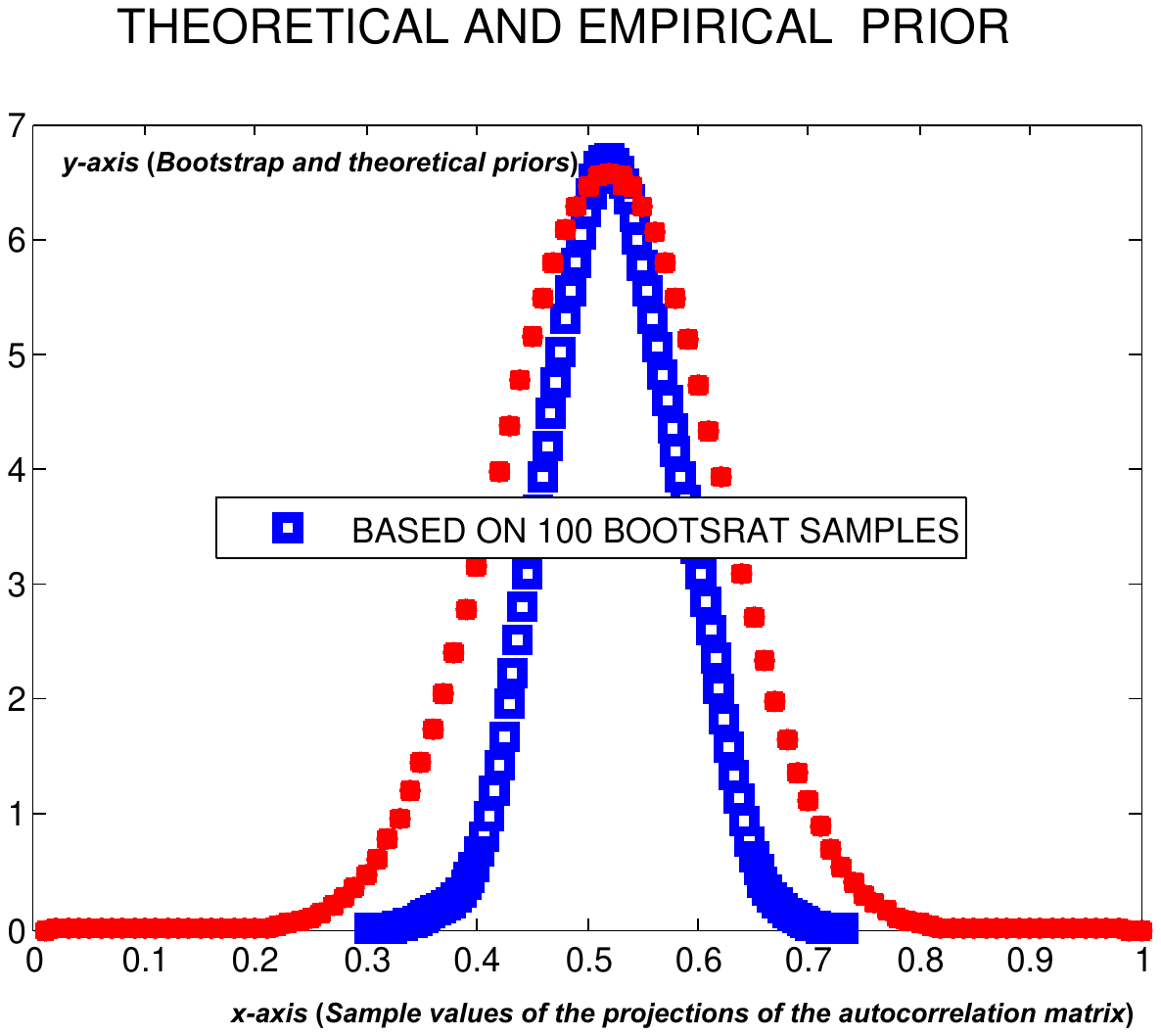}\hspace*{-2cm}
 \caption{At the left--hand side, empirical projections of the
 autocorrelation operator $\rho,$ reflecting temporal
 autocorrelation and cross--correlation between the
 $17$ Spanish Communities analyzed.  At the right--hand side,
  the considered   prior probability density (red squares)
  of a scaled, by factor $1/3,$ Beta distributed random variable
  with shape parameters $14$ and $13$  is compared with the bootstrap  fitting
 of an  empirical prior (blue squares)}
 \label{erp}
 \end{figure}
\end{center}

Given the empirical spectral characteristics observed in the   regularized approximation $\widehat{\rho}_{k(T)}$ of $\rho$  in  (\ref{cerho}),
 from the
singular value decomposition of the empirical operators in (\ref{accop}),  our choice of the prior for the  projections of $\rho $ has
been a scaled, by factor $1/3,$ Beta prior with hyper--parameters
$a_{pq}= 14,$ and  $b_{pq}=13,$  for $p,q=1,\dots, 17.$ The
suitability of this data--driven choice, regarding  localization of the mode, and
the  tails thickness, is illustrated in Figure \ref{erp}.
Specifically,   at the right plot in Figure \ref{erp},   both, the
scaled  Beta probability density, with shape parameters $14$ and
$13$ (red--square line), and the fitted probability density
(blue--square line), from the generated  bootstrap samples,
based on the empirical projections of $\rho,$ are displayed. Note
that the observed range of the empirical projections of $\rho$ is
well fitted, as one can see from the left plot in Figure \ref{erp}.

Bootstrap confidence intervals $\mathcal{I}_{1},\dots,\mathcal{I}_{5}$ at level  $1-\alpha =0.95$, for the expected training standard error of the multivariate  time series classical and
Bayesian  residual COVID--19 mortality log--risk  predictors, based on  $1000$ bootstrap
samples, are  displayed  in Table \ref{ci2cc}:
\begin{table}[!h]
\caption{Bootstrap confidence intervals for the expected training standard error
of the  classical  and  Bayesian  residual COVID--19 mortality log--risk predictors    ($1-\alpha =0.95$)}
\begin{center}
\begin{tabular}{ccc}
 \hline
CI/S & Classical  & Bayesian   \\
$\mathcal{I}_{1}$ &  $[0.0474, 0.0597]$   & $[0.0173, 0.0228]$\\
$\mathcal{I}_{2}$  &
  $[0.0455, 0.0578]$ & $[0.0167, 0.0220]$
\\
$\mathcal{I}_{3}$ &
  $[0.0463, 0.0588]$  & $[0.0169, 0.0225]$\\
 $\mathcal{I}_{4}$ &  $[0.0460, 0.0586]$ & $[0.0172, 0.0226]$\\
 $\mathcal{I}_{5}$ & $[0.0421, 0.0563]$  &  $[0.0158, 0.0215]$ \\
  \hline
     \label{ci2cc}
\end{tabular}
\end{center}
\end{table}

 Maps plotted  in Figure \ref{figrmbe}  show the observed
spatiotemporal evolution of COVID--19 mortality risk, and its prediction, from the fitted curve trigonometric regression model, and the subsequent  classical and  Bayesian  time series analysis.

 \begin{center}
 \begin{figure}[!h]
 \centering
  \includegraphics[width=0.49\textwidth]{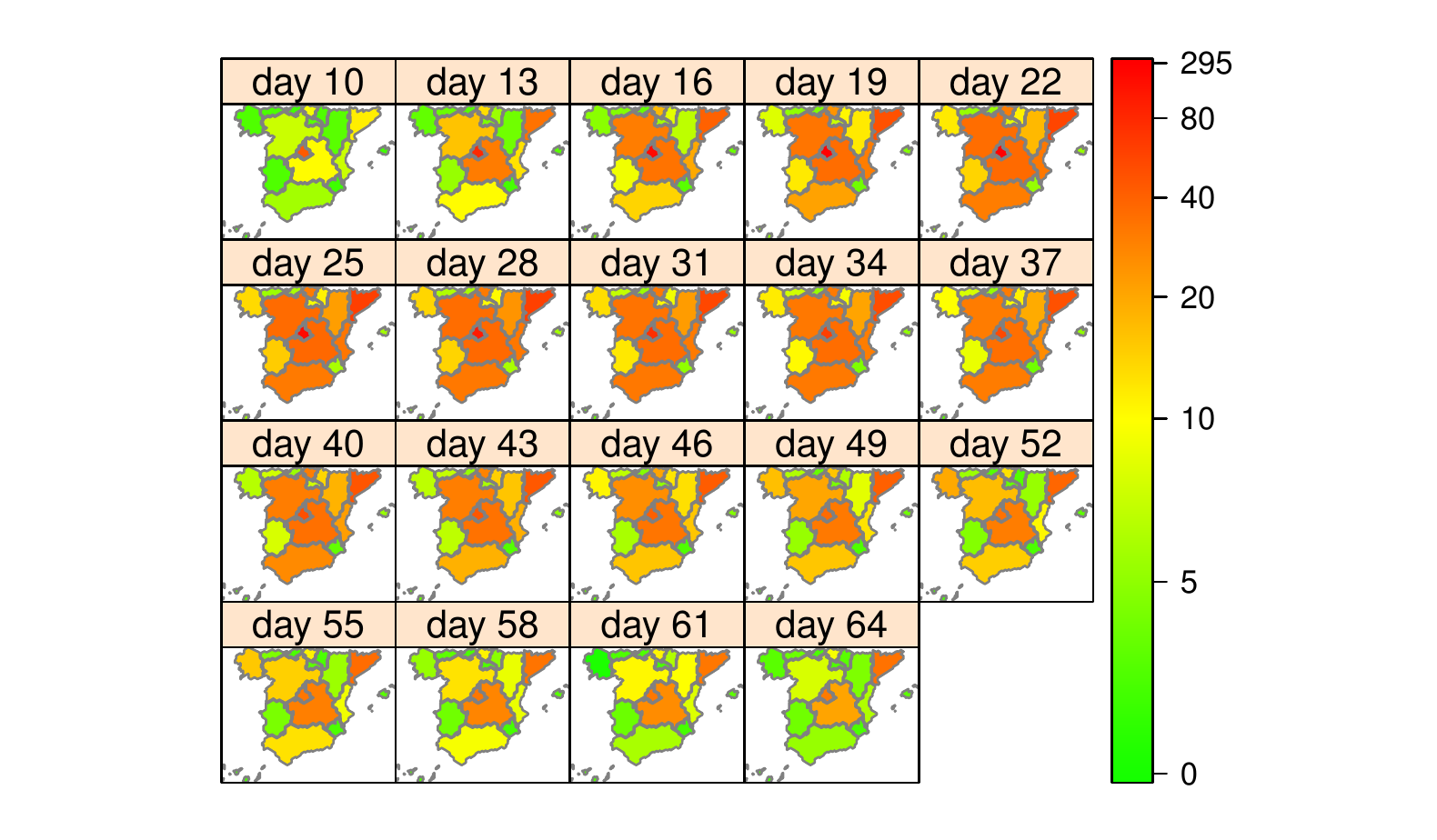}
 \includegraphics[width=0.49\textwidth]{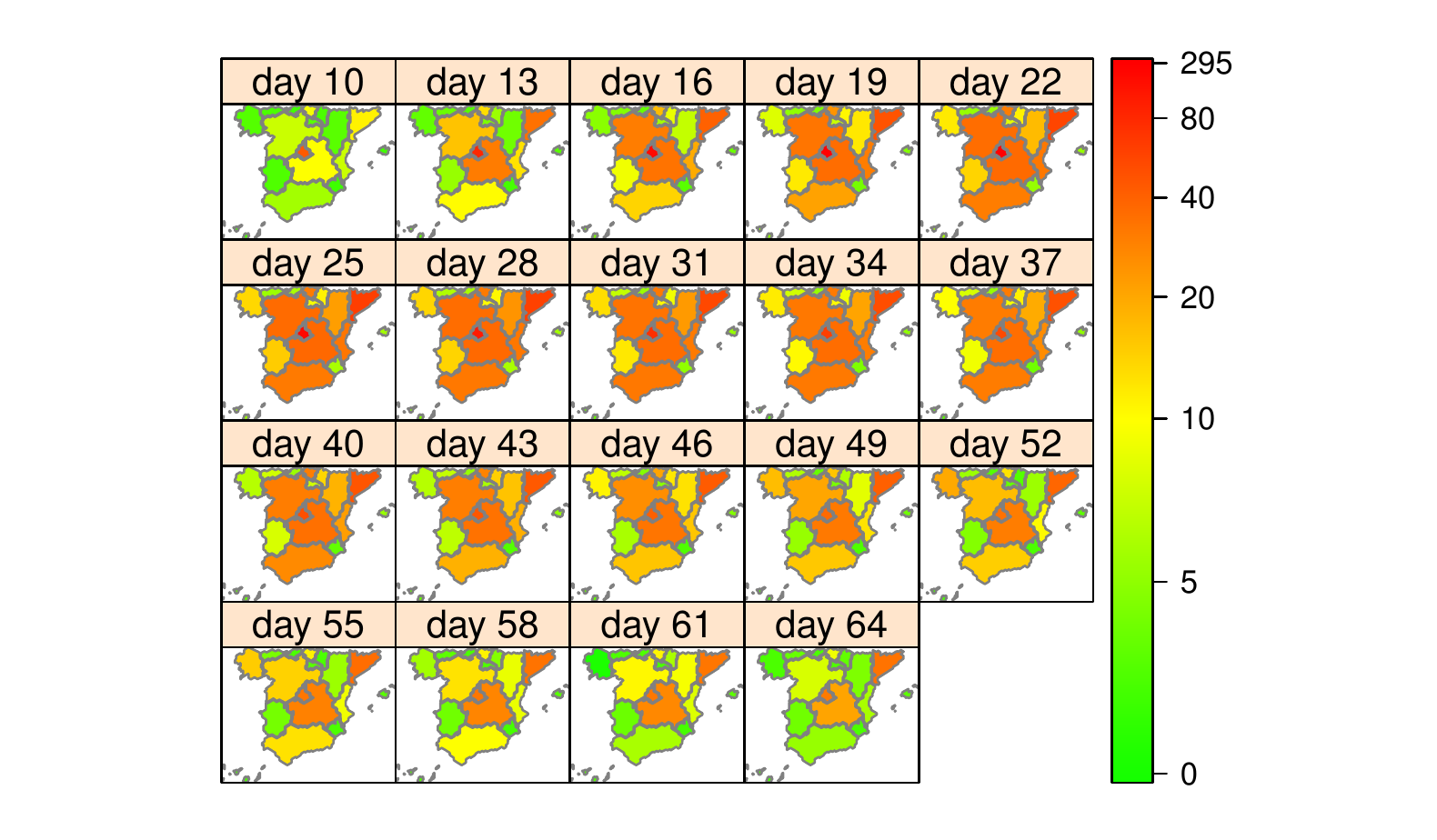}
 \includegraphics[width=0.49\textwidth]{m_I_OBS.pdf}
 \includegraphics[width=0.49\textwidth]{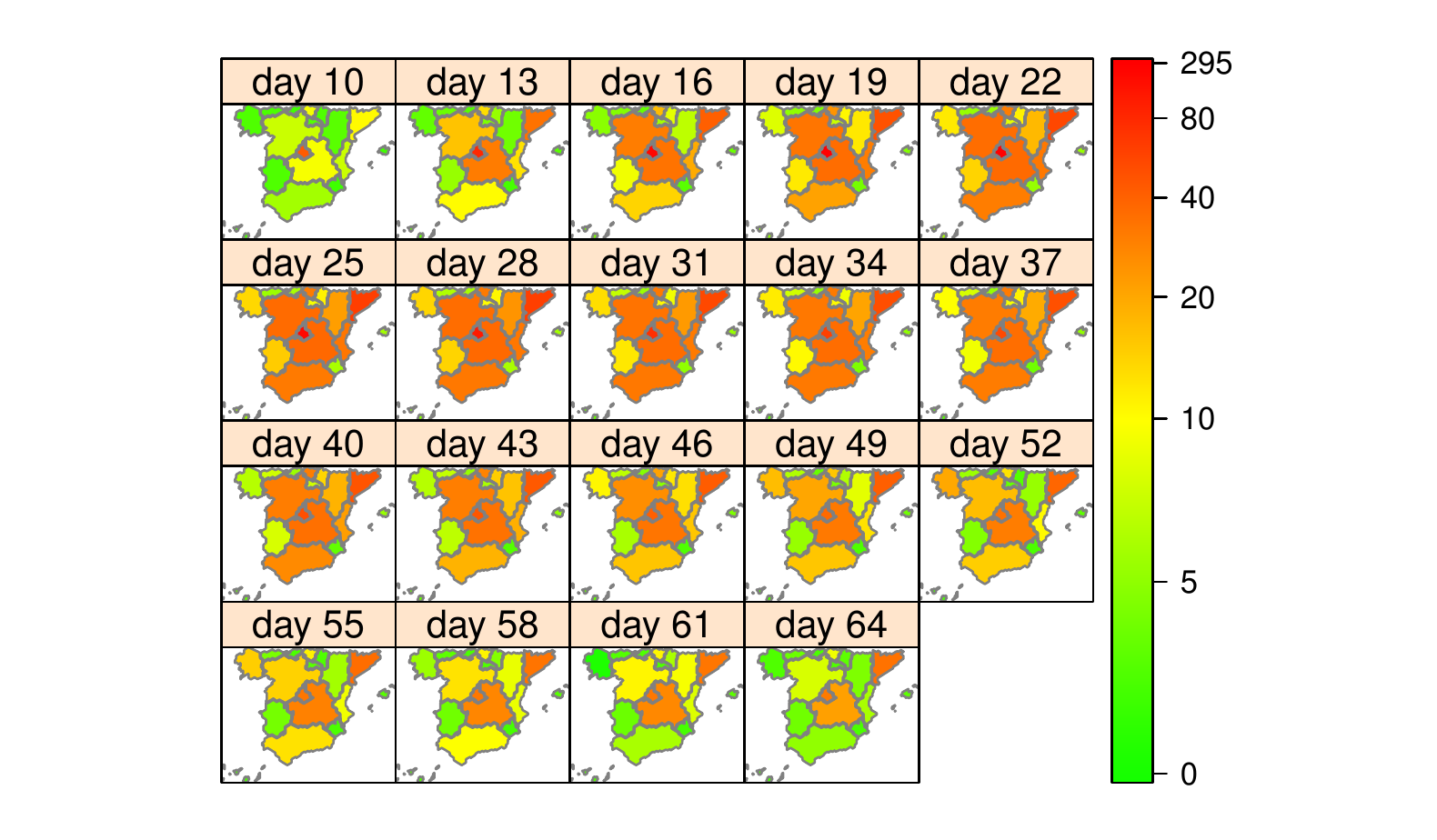}
 \caption{COVID--19 mortality risk maps, since  March, 8 to May, 13, 2020.
 Observed (left--hand--side) and estimated (right--hand side) maps,  computed  from trigonometric regression, combined with classical
  (first line) and Bayesian (second line)
 residual  predictors}
 \label{figrmbe}
 \end{figure}
\end{center}

\section{An empirical  comparative study}
\label{subtsb} The  ML regression models introduced in the Supplementary Material are applied to
COVID--19 mortality analysis, and compared, via random  $k$--fold
cross--validation and bootstrap estimators, with the multiple objective
space--time forecasting approach  presented.   We distinguish two categories respectively
referred to the strong--sense (hard--data) and weak--sense
(soft--data) definition of our data set. Random $k$--fold ($k=5,10$) cross--validation,  in terms of  Symmetric Mean Absolute Percentage Errors (SMAPEs), evaluates the performance of the compared regression models, from hard-- and soft--data. Bootstrap confidence intervals, and  probability density estimates of the spatially averaged  SMAPEs are also computed. Section \ref{CR} provides a  data--driven model  classification, based on SMAPEs, in the two categories analyzed, from random $k$--fold cross--validation, and the bootstrap estimation procedures applied.

\subsection{Results from random $k$--fold cross--validation}
\label{ecs}

After interpolation and  cubic $B$-spline smoothing of our original
data set,  the  logarithmic transform and linear scaling  are applied. We held out the first ten  points and the last three, for
each  COVID--19 mortality log--risk curve,  as an out of sample set. Our approach is implemented in the second--category from soft--data.
In this implementation, we consider $N=6,$ adopting the model selection criterion given in Section  \ref{sac19} (see equation (\ref{eqctp}) and reference \cite{Chapelle}). In the multivariate time series classical and Bayesian  prediction, our  choice of
$k(T)=k(265)=8$ provides a balance between
$k(T)=[\ln(T)]^{-}=[\ln(265)]^{-}=5,$ signing  an agreement with the
separation and velocity decay of the empirical eigenvalues of the
autocovariance operator, and the parameter value $k(T)=9,$
controlling model complexity according to the sample size $T=265.$ The random fluctuations observed at the  $k(T)$ empirical projections of the spatial autocorrelation matrix $\rho$  are also well--fitted by  our   choice of the  shape hyperparameters, characterizing  the prior Beta probability density.

 Model fitting is evaluated in
terms of the Symmetric Mean Absolute Percentage Errors (SMAPEs),
given by, for $P=17,$ and $T=265,$
\begin{equation}
\frac{1}{T}\sum_{t=1}^{T}\frac{\left|\widehat{\ln(\Lambda_{t})}(\psi_{p,\rho_{p}})-\ln(\Lambda_{t})(\psi_{p,\rho_{p}})\right|}{\left(\left|\ln(\Lambda_{t})(\psi_{p,\rho_{p}})\right|+
\left|\widehat{\ln(\Lambda_{t})}(\psi_{p,\rho_{p}})\right|\right)/2},\quad
p=1,\dots,P. \label{SMAPE}
\end{equation}

We have computed  the mean of the SMAPEs obtained at each one of the
$k$ iterations of the random $k$--fold cross--validation procedure.
   This validation technique consists of random splitting the
functional sample into a training and validation samples at each one
of the $k$ iterations. Model fitting is performed from the training
sample, and the target outputs are defined from the validation or
testing sample. By running each model ten times and averaging
SMAPEs, we remove the fluctuations due to the random initial weights
(for MLP and BNN models), and the differences in the parameter
estimation in all methods, due to the random specification of the
sample splitting  in the  random $k$--fold
cross--validation procedure.

The ten--running based random $10$--fold cross--validation  SMAPEs
are displayed in Table \ref{T2}, for the  six ML techniques tested, GRNN, MLP, SVR, BNN,
RBF,  and GP, when hard--data are considered  (see also Table 3 of the Supplementary Material on random $5$--fold cross--validation results).
 Table  \ref{T5}
 provides the ten--running based  random $10$--fold
cross--validation  results,  from
 soft--data  category  (see also Table 4 of the Supplementary Material on random $5$--fold cross validation results). The  corresponding   cross--validation  results of the presented
 approach from  soft--data are displayed
   in Table \ref{T7}.

ML model hyperparameter selection has been achieved  by applying
random $k$--fold cross--validation ($k=5,10$). Our selection has been made
from  a suitable set of candidates. Specifically, the optimal
numbers of hidden (NH) nodes in the implementation of MLP and BNN
have been selected from the candidate sets $[0, 1, 3, 5, 7, 9]$ and
[1, 3, 5, 7, 9], respectively.  The random cross--validation results in
both cases, $k=5,10,$ lead to the same choice of the NH optimal
value.  Namely, NH$=1$ for MLP, and  NH$=5$ for BNN. The last one
displays slight differences with respect to the values NH$=3,7,$ in
the random $10$--fold cross--validation implementation.
 In the same way, we have selected the respective  spread $\beta $ and bandwidth $h$ parameters in the RBF and GRNN procedures. Thus, after applying random
 $k$--fold cross--validation, with $k=5,10,$ the optimal values
 $\beta =2.5,$ and $h=0.05$ are obtained,
 from the candidate sets $[2.5, 5, 7.5, 10, 12.5, 15, 17.5, 20]$
 and $[0.05, 0.1, 0.2, 0.3, 0.5, 0.6, 0.7],$  respectively (see Supplementary Material).
     Better performance from hard--data is observed in linear SVR. In its  implementation, automatic
 hyperparameter optimization
     from \emph{fitrsvm} MatLab function is applied.
  While, from the
     soft--data
     category, the best option corresponds to the  Gaussian kernel
     based  nonlinear SVR model fitting  (applying the same
     option  of automatic hyperparameter optimization, in the argument
     of  \emph{fitrsvm} MatLab function). In the implementation of
     GP, we follow the same  tuning
procedure for model selection. In this case, for both categories, we
have selected   Bayesian  cross-validation optimization (in the
hyperparameter optimization argument of the \emph{fitrgp} MatLab
function).

In all the results displayed, the  SMAPE--MEAN (M.) and SMAPE--TOTAL
(T.) have been computed as  performance measures, for
comparing the ML models tested, and our approach.

\begin{table}[!h]
\caption{\textbf{\emph{Hard--data category}}. Averaged SMAPEs,
  based on $10$ running of random $10$--fold
cross--validation} \label{T2}
\begin{center}
\begin{tabular}{|c|c|c|c|c|c|c|}
\hline
 {\bf SC}($\mathbf{x10^{-2}}$) & {\bf GRNN} & {\bf MLP} & {\bf SVR} &
 {\bf BNN} & {\bf RBF} &  {\bf GP} \\
  \hline
C1 & 0.1957 & 0.0777 & 0.0700 & 0.0594 & 0.0543 & 0.0554 \\
C2 & 0.6132 & 0.1490 & 0.0663 & 0.0738 & 0.0680 & 0.0654 \\
C3 & 0.1556 & 0.0473 & 0.0350 & 0.0303 & 0.0331 & 0.0304 \\
C4 & 0.0971 & 0.0342 & 0.0135 & 0.0200 & 0.0182 & 0.0211 \\
C5 & 0.2049 & 0.0457 & 0.0318 & 0.0370 & 0.0369 & 0.0372 \\
C6 & 0.1572 & 0.0368 & 0.0177 & 0.0234 & 0.0233 & 0.0247 \\
C7 & 0.4898 & 0.0698 & 0.0644 & 0.0590 & 0.0616 & 0.0588 \\
C8 & 0.0804 & 0.0340 & 0.0171 & 0.0191 & 0.0211 & 0.0177 \\
C9 & 0.7258 & 0.1976 & 0.0979 & 0.0812 & 0.0326 & 0.0437 \\
C10 & 0.2191 & 0.0704 & 0.0556 & 0.0482 & 0.0471 & 0.0463 \\
C11 & 0.1262 & 0.0530 & 0.0310 & 0.0395 & 0.0375 & 0.0355 \\
C12 & 0.5228 & 0.1578 & 0.1341 & 0.1282 & 0.0940 & 0.0993 \\
C13 & 0.3594 & 0.0647 & 0.0576 & 0.0579 & 0.0533 & 0.0458 \\
C14 & 0.1345 & 0.0366 & 0.0209 & 0.0204 & 0.0194 & 0.0207 \\
C15 & 0.6080 & 0.1523 & 0.1411 & 0.1141 & 0.0982 & 0.1039 \\
C16 & 0.2464 & 0.0889 & 0.0709 & 0.0622 & 0.0568 & 0.0594 \\
C17 & 0.0660 & 0.0370 & 0.0148 & 0.0222 & 0.0203 & 0.0227 \\
\hline \hline
M. & 0.2942 & 0.0796 & 0.0553 & 0.0527 & 0.0456 & 0.0463 \\
\hline
T. & 5.0022 & 1.3528 & 0.9397 & 0.8959 & 0.7757 & 0.7879 \\
\hline
\end{tabular}
\end{center}
\end{table}

\begin{table}[!h]
\caption{\textbf{\emph{Soft--data category}}. Averaged SMAPEs, based
on $10$ running of random $10$--fold cross--validation} \label{T5}
\begin{center}
\begin{tabular}{|c|c|c|c|c|c|c|}
\hline
 {\bf SC}($\mathbf{x10^{-2}}$) & {\bf GRNN} & {\bf MLP} & {\bf SVR} & {\bf BNN} & {\bf RBF} &  {\bf GP} \\
  \hline
C1 & 0.1545 & 0.0983 & 0.0666 & 0.0573 & 0.0234 & 0.0312 \\
C2 & 0.1844 & 0.1730 & 0.0660 & 0.0749 & 0.0277 & 0.0301 \\
C3 & 0.1029 & 0.1192 & 0.0481 & 0.0452 & 0.0273 & 0.0274 \\
C4 & 0.0432 & 0.0286 & 0.0165 & 0.0158 & 0.0124 & 0.0123 \\
C5 & 0.0610 & 0.0476 & 0.0258 & 0.0248 & 0.0144 & 0.0149 \\
C6 & 0.0260 & 0.0217 & 0.0133 & 0.0140 & 0.0124 & 0.0125 \\
C7 & 0.3750 & 0.2026 & 0.1095 & 0.0924 & 0.0307 & 0.0399 \\
C8 & 0.0764 & 0.0482 & 0.0305 & 0.0300 & 0.0262 & 0.0187 \\
C9 & 0.4894 & 0.3198 & 0.1753 & 0.1212 & 0.0229 & 0.0372 \\
C10 & 0.1680 & 0.0815 & 0.0521 & 0.0462 & 0.0252 & 0.0290 \\
C11 & 0.1537 & 0.0839 & 0.0436 & 0.0397 & 0.0199 & 0.0219 \\
C12 & 0.3689 & 0.2558 & 0.1505 & 0.1249 & 0.0401 & 0.0490 \\
C13 & 0.2848 & 0.1582 & 0.0968 & 0.0792 & 0.0240 & 0.0320 \\
C14 & 0.0367 & 0.0226 & 0.0120 & 0.0143 & 0.0106 & 0.0104 \\
C15 & 0.3618 & 0.2264 & 0.1201 & 0.1227 & 0.0317 & 0.0522 \\
C16 & 0.1773 & 0.0835 & 0.0651 & 0.0545 & 0.0264 & 0.0318 \\
C17 & 0.0884 & 0.0623 & 0.0210 & 0.0231 & 0.0125 & 0.0136 \\
\hline \hline
M. & 0.1854 & 0.1196 & 0.0655 & 0.0577 & 0.0228 & 0.0273 \\
\hline
T. & 3.1524 & 2.0333 & 1.1129 & 0.9801 & 0.3877 & 0.4642 \\
\hline
\end{tabular}
\end{center}
\end{table}

\begin{table}[!h]
\caption{\textbf{\emph{Our approach}}. Averaged  SMAPEs, based on
 $10$ running of random $10$--fold cross--validation, for testing  trigonometric regression combined with Classical (C.) and Bayesian (B.) residual analysis } \label{T7}
\begin{center}
\begin{tabular}{|c|c|c|}
\hline
 {\bf SC} &  {\bf C. k10} & {\bf B. k10} \\
  \hline
& &  \\
C1 &$0.0024$ & $0.7106(10)^{-3}$ \\
C2 & $0.0019$ & $0.4003(10)^{-3}$   \\
C3 & $0.0016$ & $ 0.6797(10)^{-3}$ \\
C4 & $0.0017$ & $0.4367(10)^{-3}$ \\
C5 &$ 0.0023$ &$ 0.6530(10)^{-3}$ \\
C6 &$ 0.0018$ &$ 0.5854(10)^{-3}$ \\
C7 & $0.0017$ & $0.6341(10)^{-3}$ \\
C8 & $0.0016$ & $0.6593(10)^{-3}$ \\
C9 & $0.0013$ & $0.5979(10)^{-3}$  \\
C10 & $0.0019$ &$ 0.6954(10)^{-3}$ \\
C11 & $0.0017$ & $0.5444(10)^{-3}$ \\
C12 & $0.0016$ & $0.5016(10)^{-3}$ \\
C13 & $0.0020$ &$ 0.4832(10)^{-3}$ \\
C14 & $0.0026$ & $0.6544(10)^{-3}$ \\
C15 & $0.0023$ & $0.6616(10)^{-3}$ \\
C16 & $0.0015$ & $0.7134(10)^{-3}$ \\
C17 & $0.0022$ & $0.6781(10)^{-3}$ \\
\hline \hline
M. & $ 0.0019$  &  $ 0.60524(10)^{-3}$ \\
\hline
T. & $ 0.0321$  & $ 0.0103$   \\
\hline
\end{tabular}
\end{center}
\end{table}
\subsection{Bootstrap based classification results}
\label{BCR}

For the ML regression models
tested, in the hard-- and soft--data categories,
bootstrap confidence intervals ($1-\alpha =0.95$ confidence level) for the spatially averaged SMAPEs,
based on $1000$ bootstrap samples,  are constructed.  Our approach requires the soft--data information to be incorporated. As before, the computed bootstrap confidence intervals $\mathcal{I}_{i},$
$i=1,\dots,5,$ are respectively  based on the bias corrected and accelerated
percentile method ($\mathcal{I}_{1}$); Normal approximated interval
with bootstrapped bias and standard error ($\mathcal{I}_{2}$); basic
percentile method ($\mathcal{I}_{3}$); bias corrected percentile
method ($\mathcal{I}_{4}$), and Student--based confidence interval
($\mathcal{I}_{5}$) (see Tables \ref{tbbihh} and \ref{tbbihhsd}). The
  bootstrap histogram, and probability density of the
spatially averaged SMAPEs are displayed in Figures \ref{figrmbe} and
\ref{figrmbe2}, for the hard--data category, and in Figures  \ref{figrmbesd}, \ref{figrmbe2sddd}, and \ref{figrmbe3sd}, for the soft--data category. The data--driven performance--based model classification results obtained  are discussed in Section \ref{CR}.
{\small
\begin{table}\caption{\textbf{\emph{Hard--data category}}. Bootstrap confidence intervals ($1-\alpha =0.95$) for the spatially
averaged  SMAPEs from $1000$ bootstrap samples   ($T=265,$ $P=17$)} \label{tbbihh}
\begin{center}
\begin{tabular}{lll}
 \hline
CI/ML & {\bf GRNN} & {\bf MLP}   \\
$\mathcal{I}_{1}$ & $[2.1(10)^{-3},
    4.1(10)^{-3}]$ & $[0.5(10)^{-3},
    1(10)^{-3}]$  \\
$\mathcal{I}_{2}$ & $[2(10)^{-3},
    3.9(10)^{-3}]$ & $[0.4776(10)^{-3},
    0.9483(10)^{-3}]$
\\
$\mathcal{I}_{3}$
 & $[2(10)^{-3},
    4(10)^{-3}]$ & $[0.4746(10)^{-3},
    0.9713(10)^{-3}]$
    \\
 $\mathcal{I}_{4}$ & $[2(10)^{-3},
    4(10)^{-3}]$ & $[0.5118(10)^{-3},
    0.9878(10)^{-3}]$\\
 $\mathcal{I}_{5}$ & $[1.7(10)^{-3},
    3.9(10)^{-3}]$  & $[0.2780(10)^{-3},
    0.9244(10)^{-3}]$  \\
  \hline
  CI/ML &  {\bf SVR} & {\bf BNN}\\
$\mathcal{I}_{1}$ & $[0.3682(10)^{-3},
    0.7219(10)^{-3}]$ &
$[0.3720(10)^{-3},
    0.6659(10)^{-3}]$\\
  $\mathcal{I}_{2}$ &   $[0.3516(10)^{-3},
    0.6763(10)^{-3}]$ & $[0.3587(10)^{-3},
    0.6379(10)^{-3}]$\\ $\mathcal{I}_{3}$
& $[0.3493(10)^{-3},
    0.6770(10)^{-3}]$
& $[0.3668(10)^{-3},
    0.6509(10)^{-3}]$\\
$\mathcal{I}_{4}$ &  $[0.3508(10)^{-3},
    0.6865(10)^{-3}]$ & $[0.3654(10)^{-3},
    0.6379(10)^{-3}]$ \\
    $\mathcal{I}_{5}$ & $[0.3050(10)^{-3},
    0.6661(10)^{-3}]$ & $[0.3099(10)^{-3},
    0.6335(10)^{-3}]$ \\
\hline
CI/ML &   {\bf RBF} &  {\bf GP} \\
$\mathcal{I}_{1}$ &  $[0.3260(10)^{-3},
    0.5310(10)^{-3}]$ & $[0.3243(10)^{-3},
    0.5350(10)^{-3}]$\\
   $\mathcal{I}_{2}$ &     $[0.3155(10)^{-3},
    0.5159(10)^{-3}]$ &  $[0.3065(10)^{-3},
    0.5126(10)^{-3}]$\\
   $\mathcal{I}_{3}$
 & $[0.3140(10)^{-3},
    0.5270(10)^{-3}]$ & $[0.3095(10)^{-3},
    0.5188(10)^{-3}]$\\
$\mathcal{I}_{4}$ &   $[0.3247(10)^{-3},
    0.5338(10)^{-3}]$ & $[0.3152(10)^{-3},
    0.5222(10)^{-3}]$\\
$\mathcal{I}_{5}$ &   $[0.2677(10)^{-3},
    0.5141(10)^{-3}]$ & $[0.2505(10)^{-3},
    0.5046(10)^{-3}]$\\
    \hline
    \end{tabular}
\end{center}
\end{table}
\begin{table}\caption{\textbf{\emph{Soft--data category}}. Bootstrap confidence intervals ($1-\alpha =0.95$) for the spatially
averaged  SMAPEs from $1000$ bootstrap samples   ($T=265,$ $P=17$)} \label{tbbihhsd}
\begin{center}
\begin{tabular}{lll}
 \hline
CI/ML & {\bf GRNN} & {\bf MLP}   \\
$\mathcal{I}_{1}$ & $[1.3(10)^{-3},
    2.6(10)^{-3}]$ & $[0.6(10)^{-3},
    1.3(10)^{-3}]$  \\
$\mathcal{I}_{2}$ & $[1.3(10)^{-3},
    2.6(10)^{-3}]$
     & $[0.6(10)^{-3},
    1.3(10)^{-3}]$
\\
$\mathcal{I}_{3}$
 & $[1.3(10)^{-3},
    2.7(10)^{-3}]$ & $[0.6(10)^{-3},
    1.3(10)^{-3}]$
    \\
 $\mathcal{I}_{4}$ & $[1.3(10)^{-3},
    2.6(10)^{-3}]$ & $[0.7(10)^{-3},
    1.3(10)^{-3}]$\\
 $\mathcal{I}_{5}$ & $[1(10)^{-3},
    2.7(10)^{-3}]$  & $[0.5(10)^{-3},
    1.3(10)^{-3}]$  \\
  \hline
  CI/ML &  {\bf SVR} & {\bf BNN}\\
$\mathcal{I}_{1}$ & $[0.4096(10)^{-3},
    0.8221(10)^{-3}]$ &
$[0.3588(10)^{-3},
    0.6177(10)^{-3}]$\\
  $\mathcal{I}_{2}$ &   $[0.3764(10)^{-3},
    0.7763(10)^{-3}]$ & $[0.3433(10)^{-3},
    0.6053(10)^{-3}]$\\ $\mathcal{I}_{3}$
& $[0.3900(10)^{-3},
    0.7889(10)^{-3}]$
& $[0.3454(10)^{-3},
    0.6037(10)^{-3}]$\\
$\mathcal{I}_{4}$ &  $[0.4108(10)^{-3},
    0.7805(10)^{-3}]$ & $[0.3559(10)^{-3},
    0.5988(10)^{-3}]$ \\
    $\mathcal{I}_{5}$ & $[0.3105(10)^{-3},
    0.7818(10)^{-3}]$ & $[0.3003(10)^{-3},
    0.6129(10)^{-3}]$ \\
\hline
CI/ML &   {\bf RBF} &  {\bf GP} \\
$\mathcal{I}_{1}$ &  $[0.1794(10)^{-3},
    0.2478(10)^{-3}]$ & $[0.2095(10)^{-3},
    0.3248(10)^{-3}]$\\
   $\mathcal{I}_{2}$ &     $[0.1754(10)^{-3},
    0.2474(10)^{-3}]$ &  $[0.2065(10)^{-3},
    0.3215(10)^{-3}]$\\
   $\mathcal{I}_{3}$
 & $[0.1785(10)^{-3},
    0.2485(10)^{-3}]$ & $[0.2079(10)^{-3},
    0.3262(10)^{-3}]$\\
$\mathcal{I}_{4}$ &   $[0.1743(10)^{-3},
    0.2494(10)^{-3}]$ & $[0.2091(10)^{-3},
    0.3258(10)^{-3}]$\\
$\mathcal{I}_{5}$ &   $[0.1616(10)^{-3},
    0.2542(10)^{-3}]$ & $[0.1941(10)^{-3},
    0.3232(10)^{-3}]$\\
    \hline
    CI/OA &{\bf Classical}  &   {\bf Bayesian}  \\
$\mathcal{I}_{1}$ & $[2.2(10)^{-3},
    3.7(10)^{-3}]$ &  $[0.2943(10)^{-3},
    0.5177(10)^{-3}]$ \\
$\mathcal{I}_{2}$ & $[2.1(10)^{-3},
    3.3(10)^{-3}]$
 &
 $[0.2802(10)^{-3},
    0.4854(10)^{-3}]$
\\
$\mathcal{I}_{3}$
 & $[2.1(10)^{-3},
    3.3(10)^{-3}]$   & $[0.2833(10)^{-3},
    0.4884(10)^{-3}]$
    \\
 $\mathcal{I}_{4}$ &   $[2.2(10)^{-3},
    3.4(10)^{-3}]$ &  $[0.2900(10)^{-3},
    0.5124(10)^{-3}]$ \\
 $\mathcal{I}_{5}$
  &  $[1.8(10)^{-3},
    3.2(10)^{-3}]$  &  $[0.2418(10)^{-3},
    0.4664(10)^{-3}]$\\
  \hline
\end{tabular}
\end{center}
\end{table}}

\begin{center}
 \begin{figure}[!h]
 \centering
 \includegraphics[width=5cm,height=5.5cm]{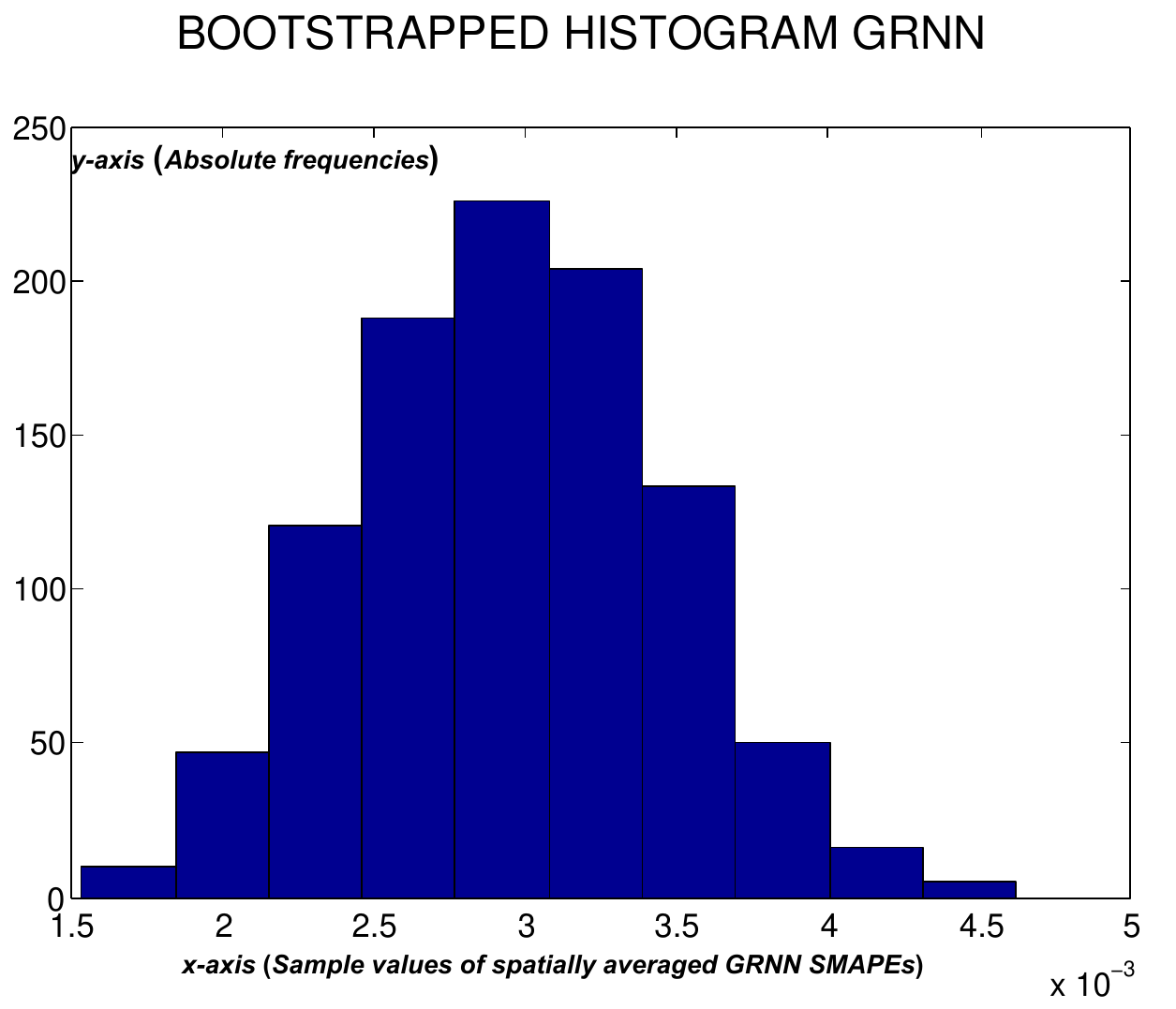}
 \includegraphics[width=5cm,height=5.5cm]{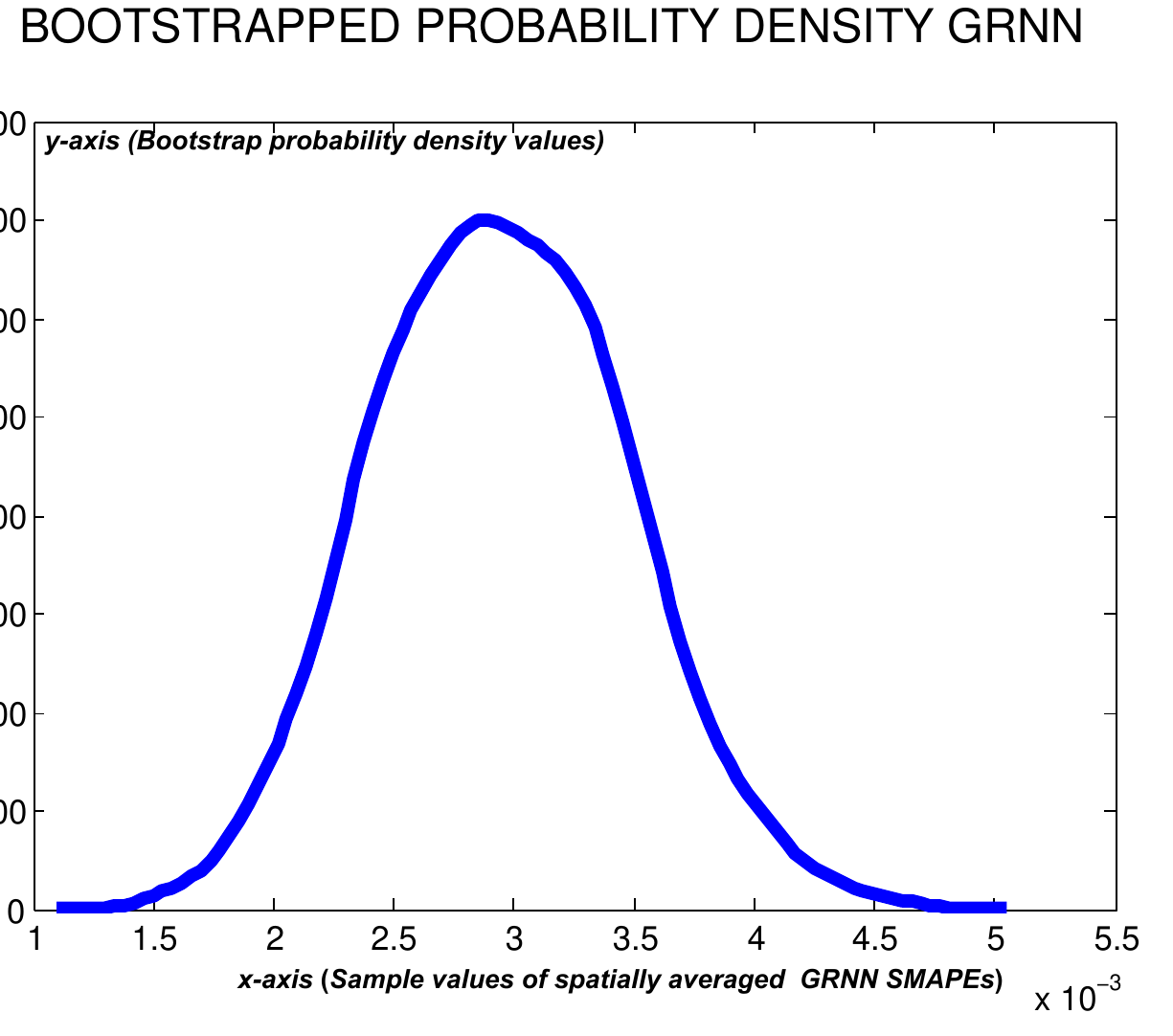}
 \vspace*{0.3cm}
 \includegraphics[width=5cm,height=5.5cm]{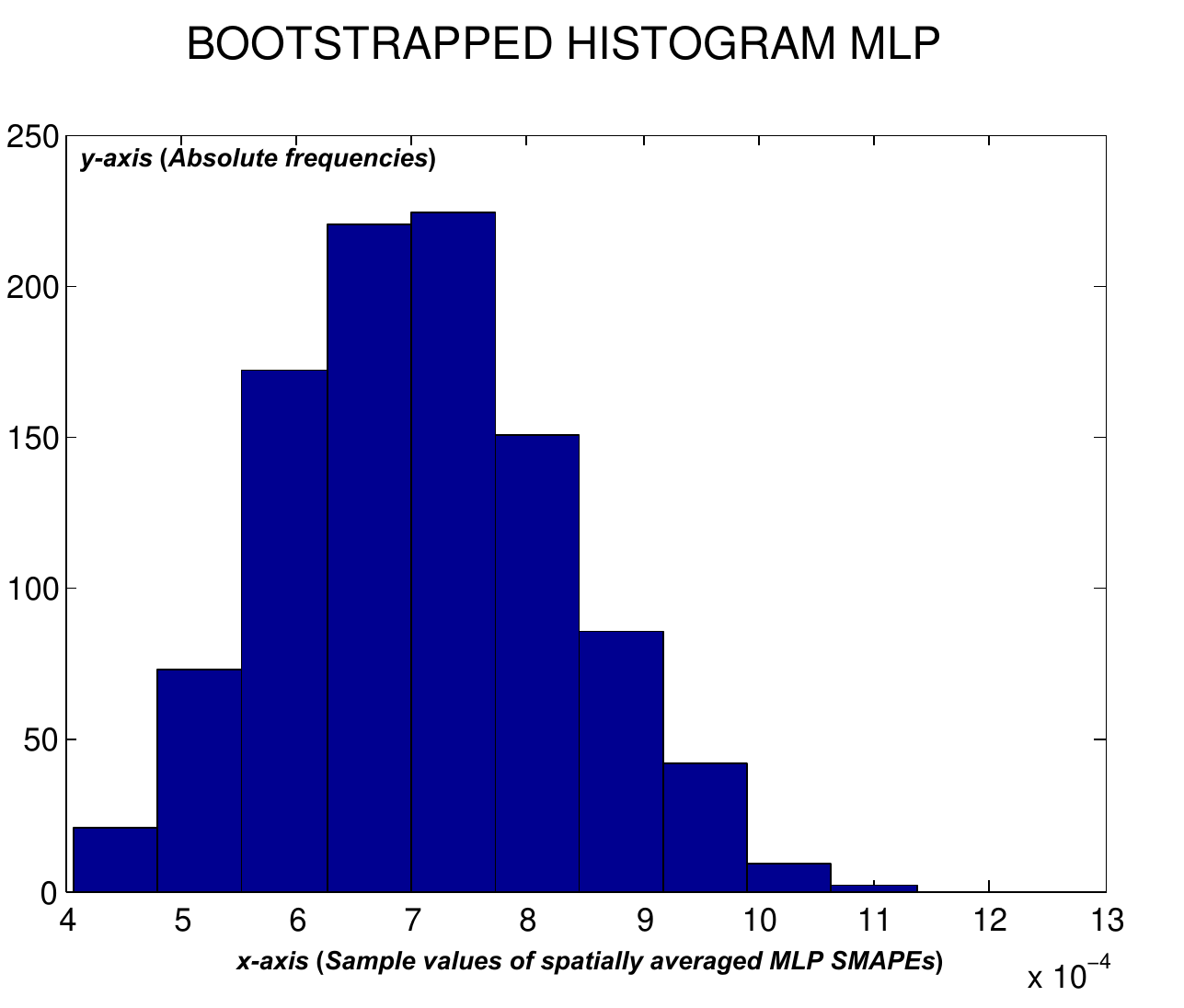}
 \includegraphics[width=5cm,height=5.5cm]{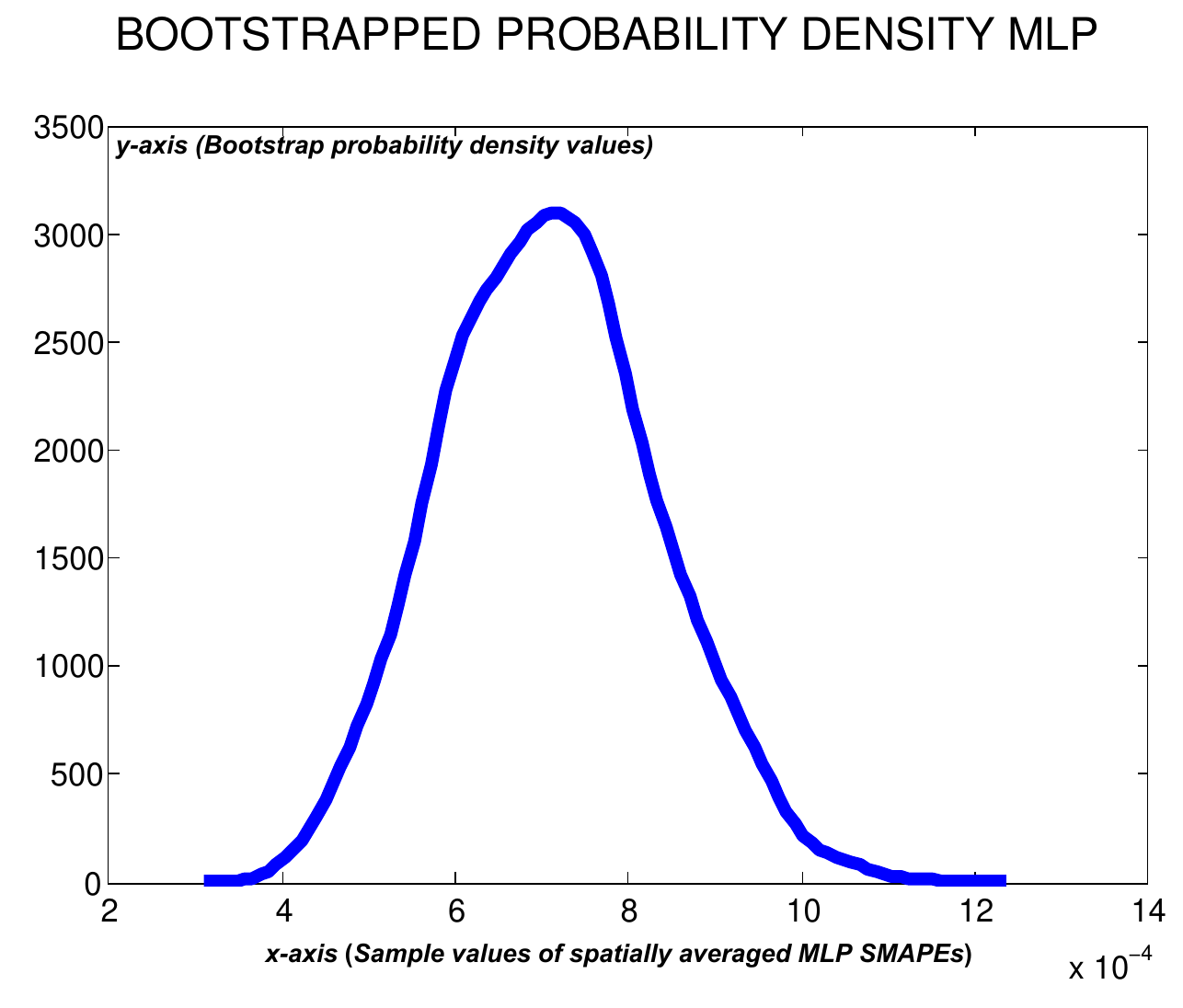}
 \vspace*{0.3cm}
 \includegraphics[width=5cm,height=5.5cm]{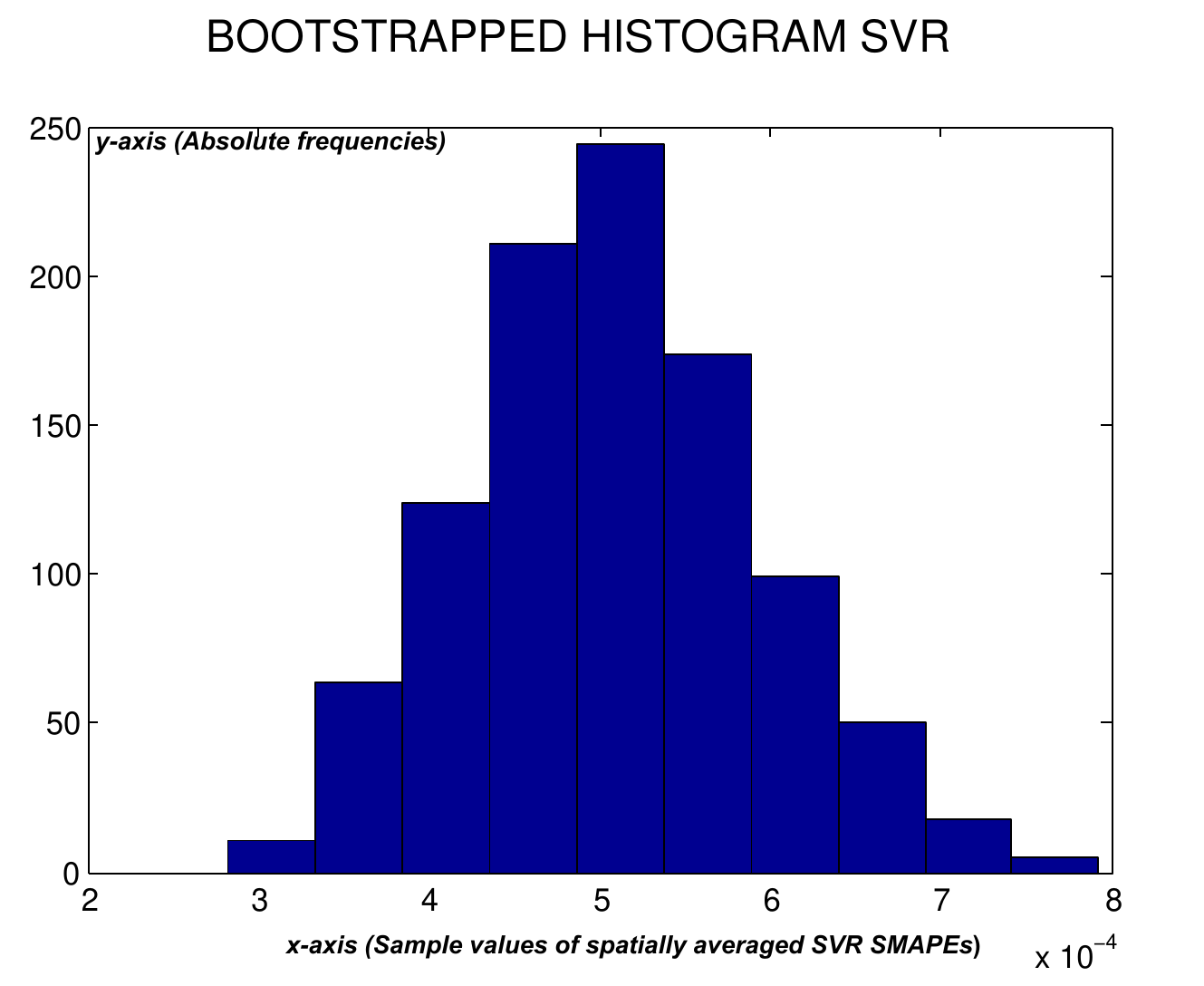}
 \includegraphics[width=5cm,height=5.5cm]{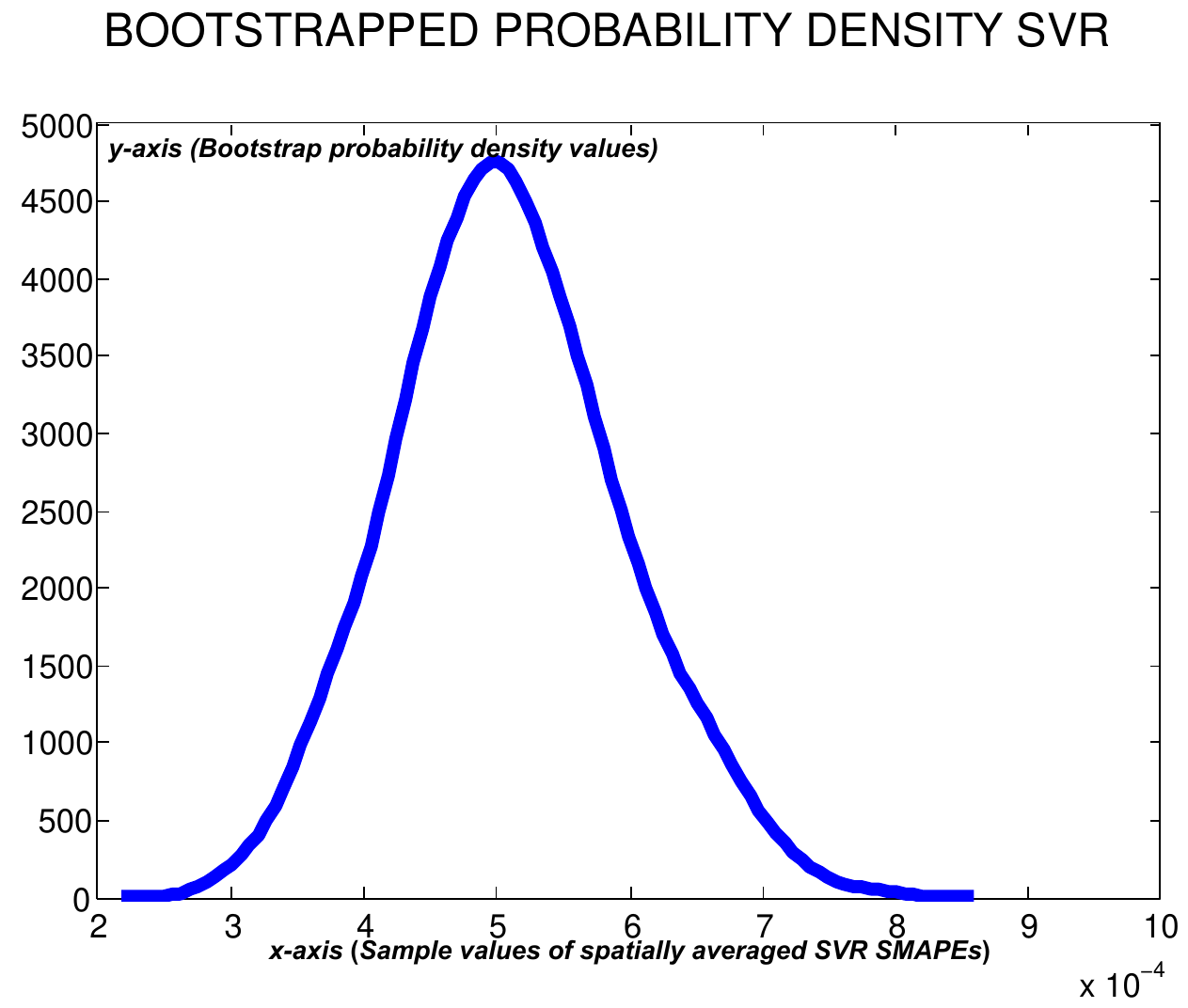}
 \caption{\textbf{\emph{Hard--data category}}. From $1000$ bootstrap  samples, spatially averaged SMAPEs histograms and probability
 densities are plotted, for GRNN (top), MLP (center), and linear SVR (bottom)}
 \label{figrmbe}
 \end{figure}
\end{center}

\begin{center}
 \begin{figure}[!h]
 \centering
 \includegraphics[width=5cm,height=5.5cm]{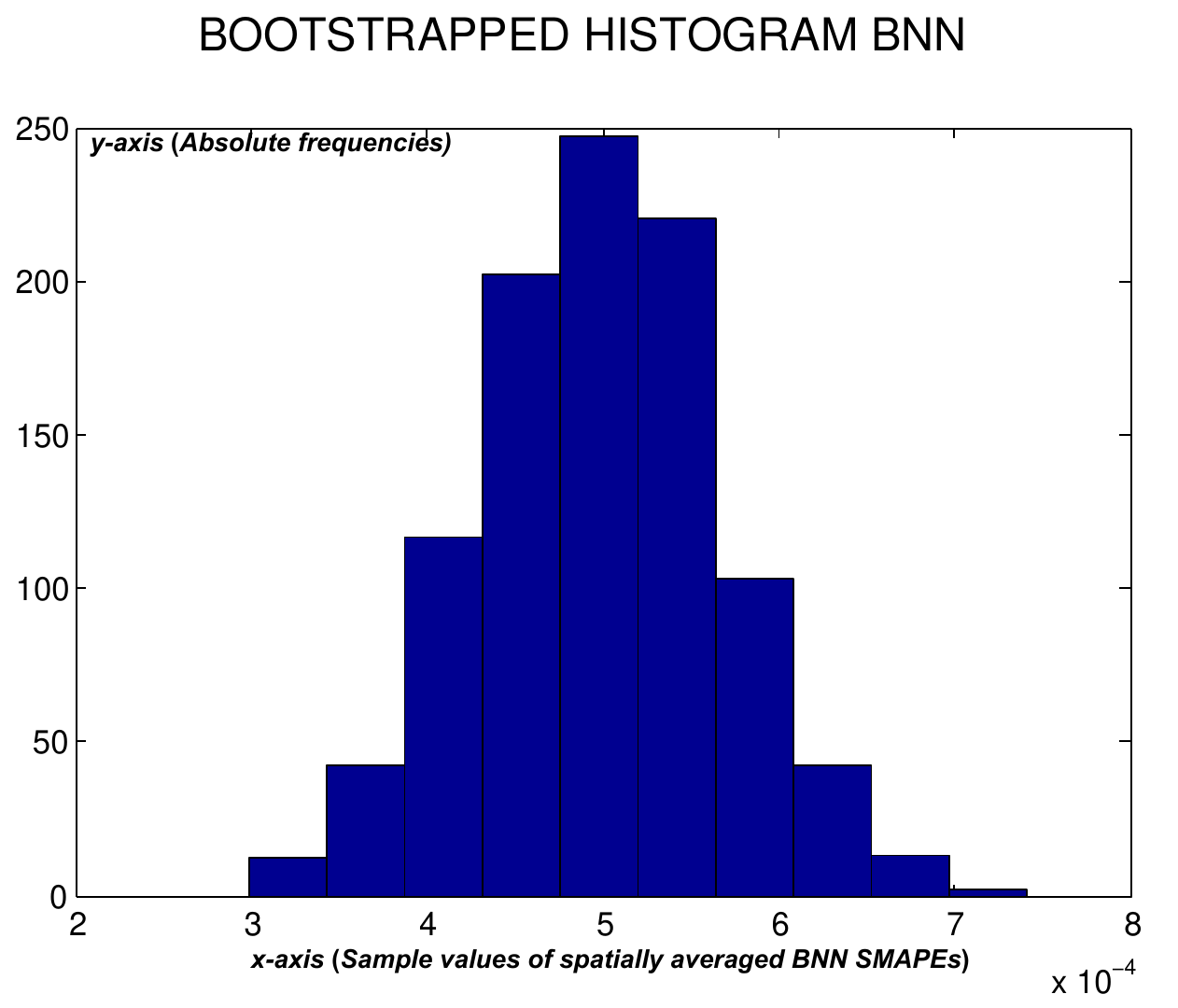}
 \includegraphics[width=5cm,height=5.5cm]{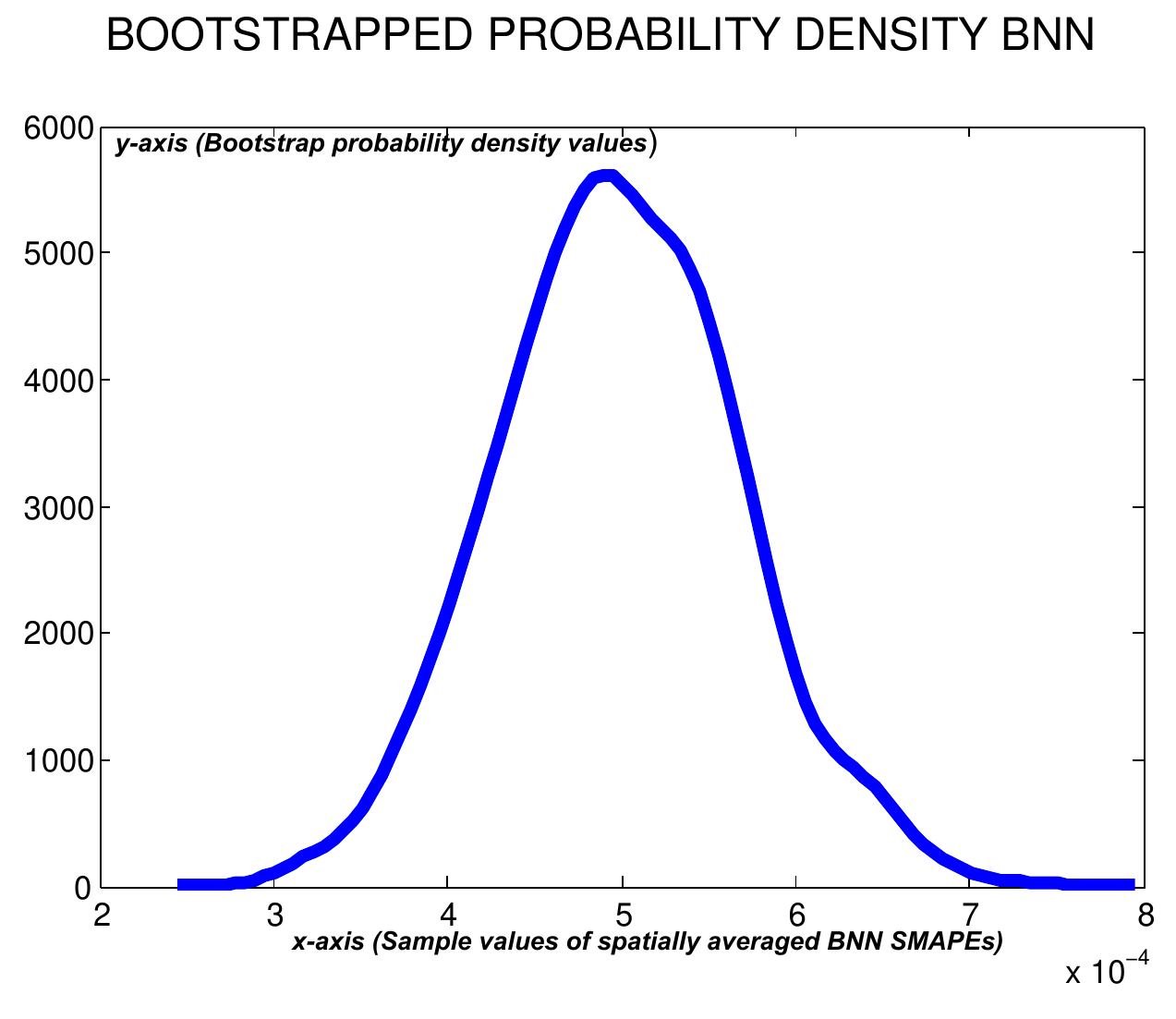}
 \vspace*{0.3cm}
 \includegraphics[width=5cm,height=5.5cm]{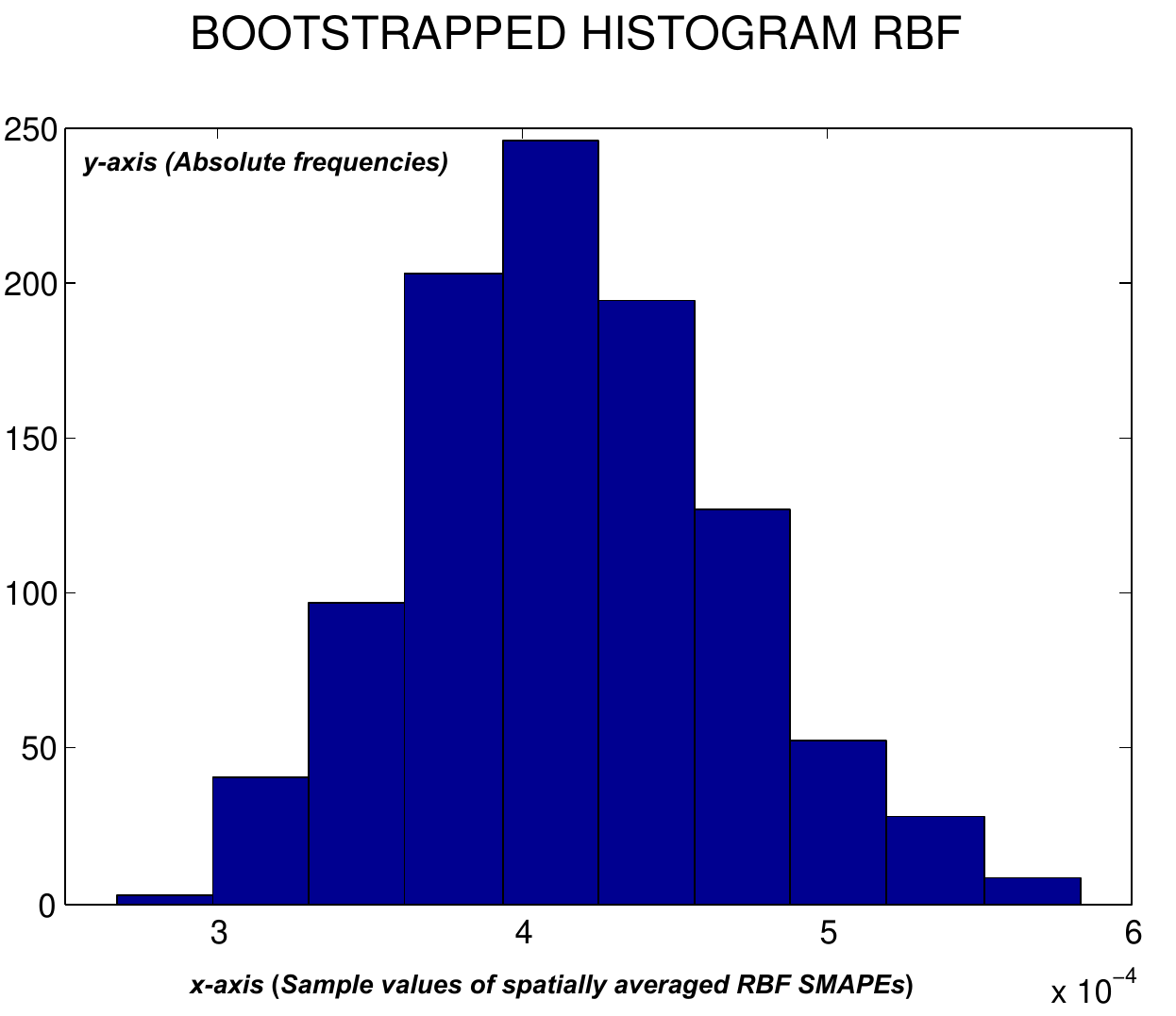}
 \includegraphics[width=5cm,height=5.5cm]{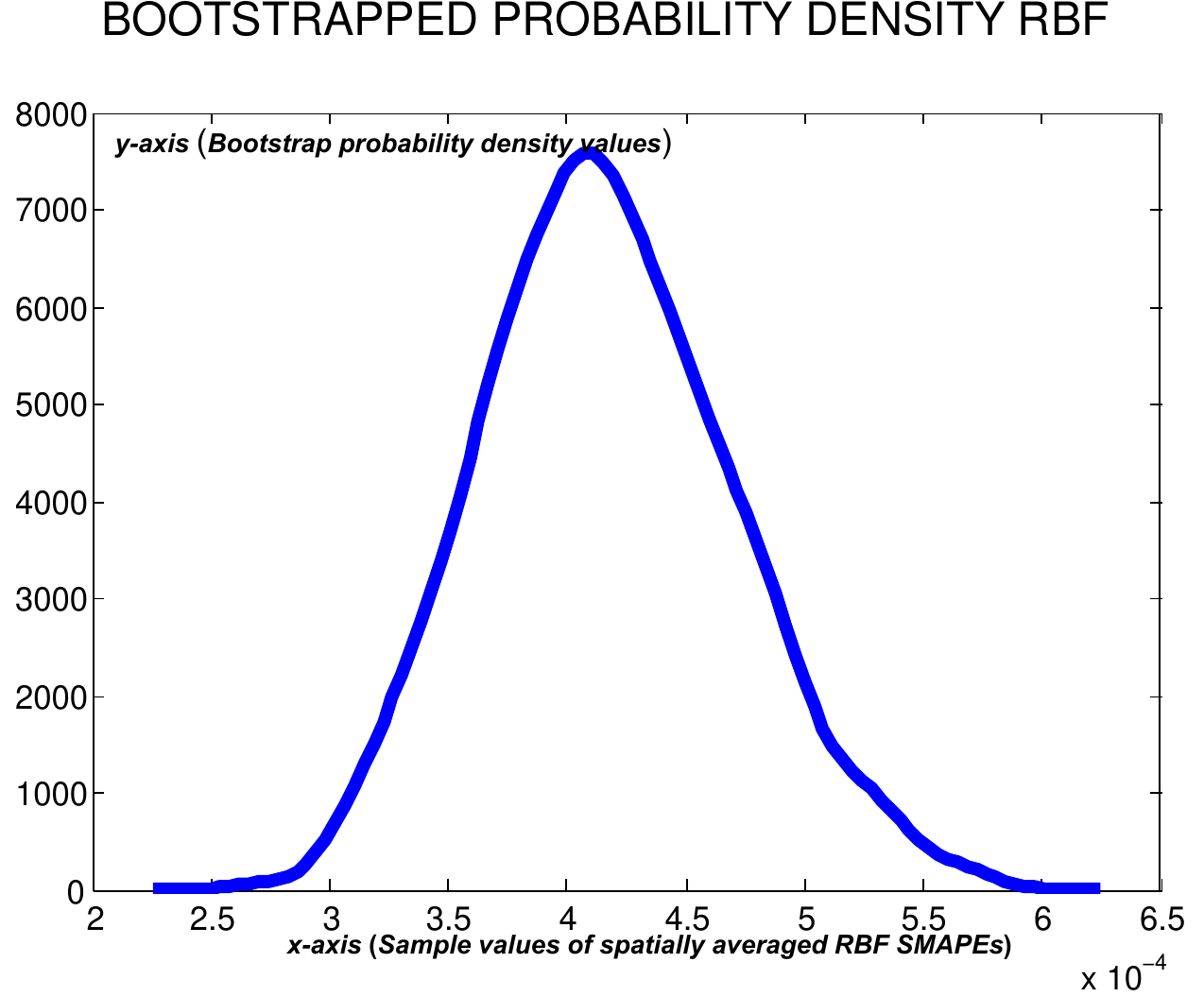}
 \vspace*{0.3cm}
 \includegraphics[width=5cm,height=5.5cm]{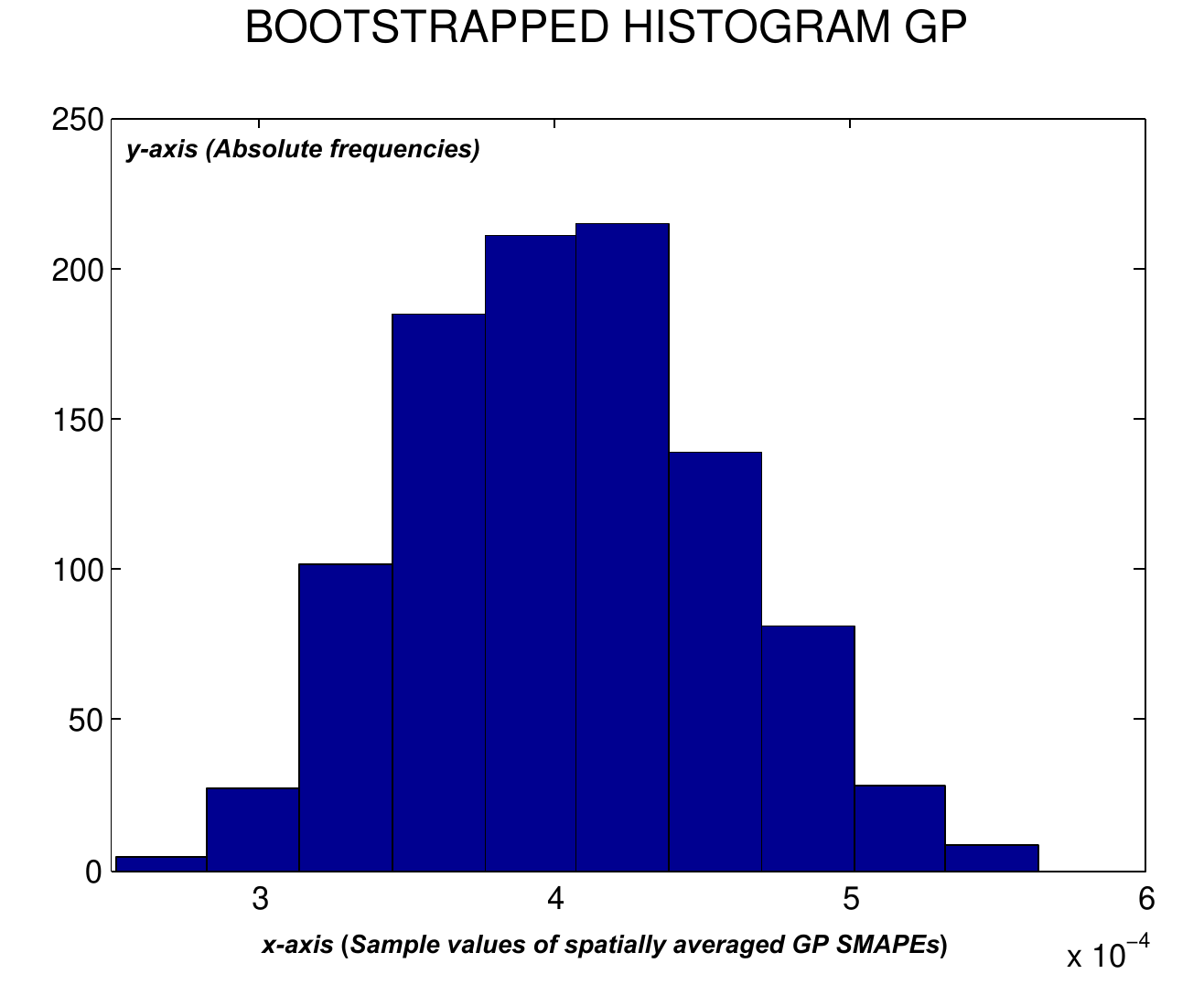}
 \includegraphics[width=5cm,height=5.5cm]{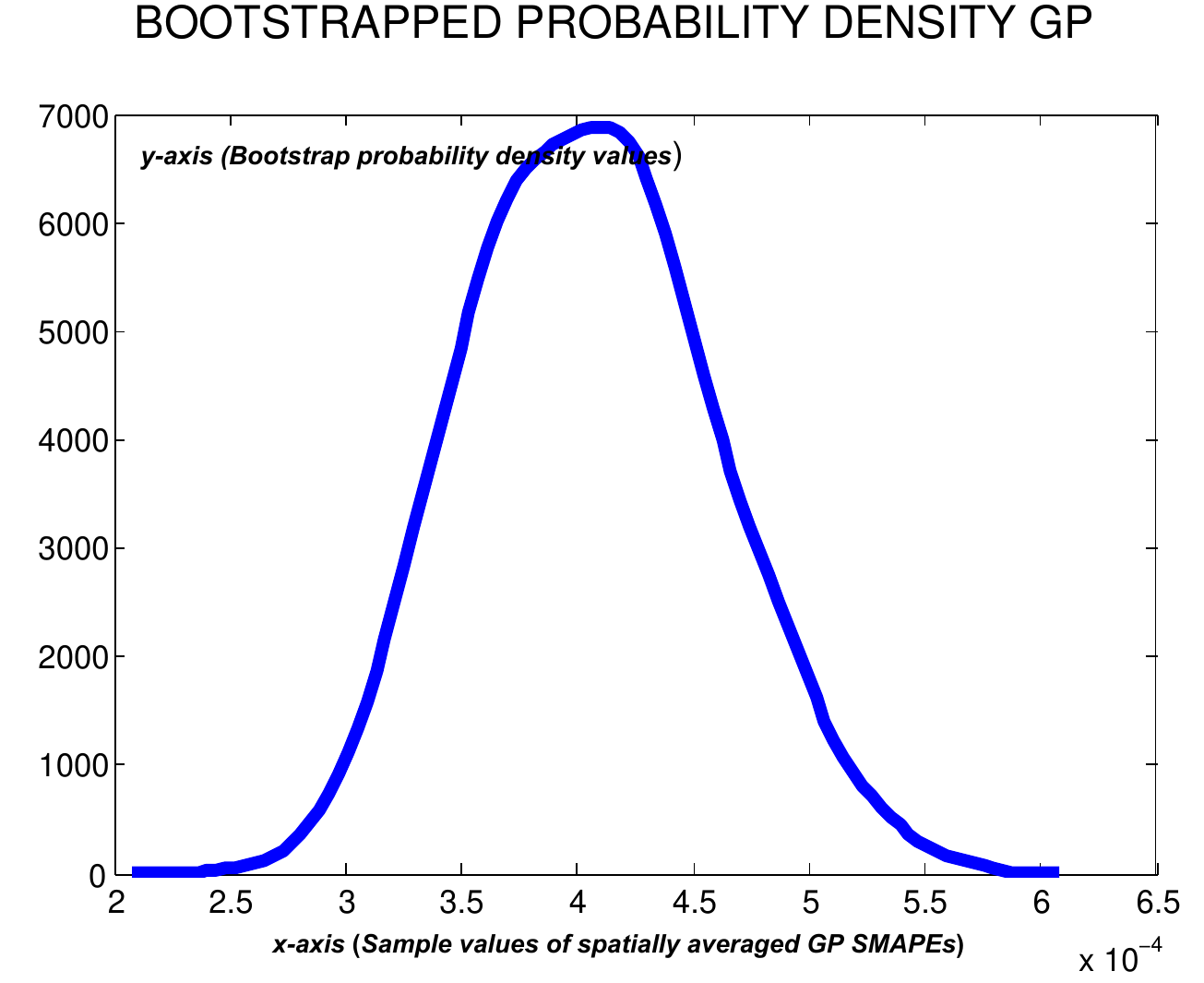}
 \caption{\textbf{\emph{Hard--data category}}. From $1000$ bootstrap  samples, spatially averaged SMAPEs histograms and probability
 densities are plotted, for BNN (top), RBF (center), and GP (bottom)}
 \label{figrmbe2}
 \end{figure}
\end{center}

\begin{center}
 \begin{figure}[!h]
 \centering
 \includegraphics[width=5cm,height=5.5cm]{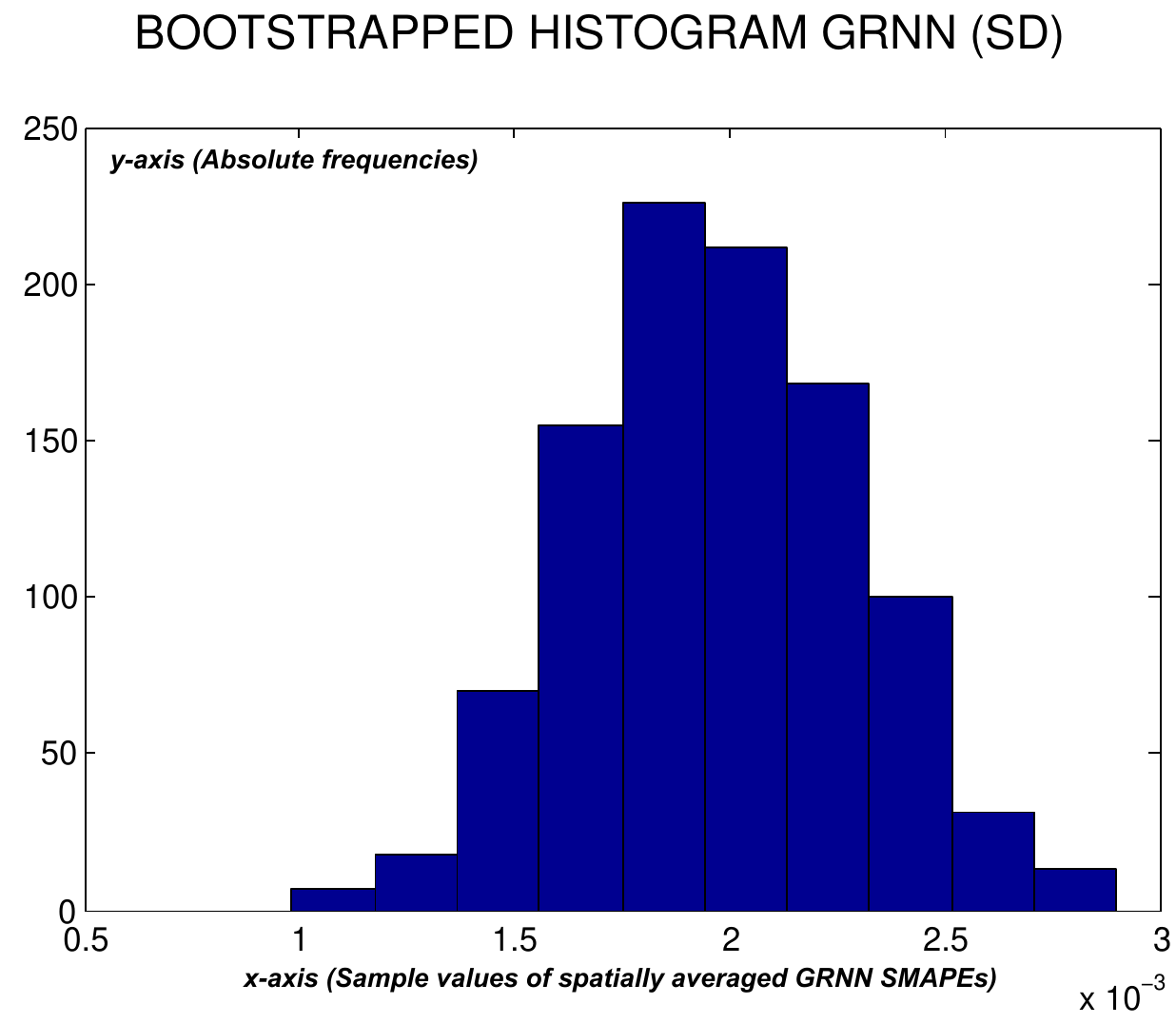}
 \includegraphics[width=5cm,height=5.5cm]{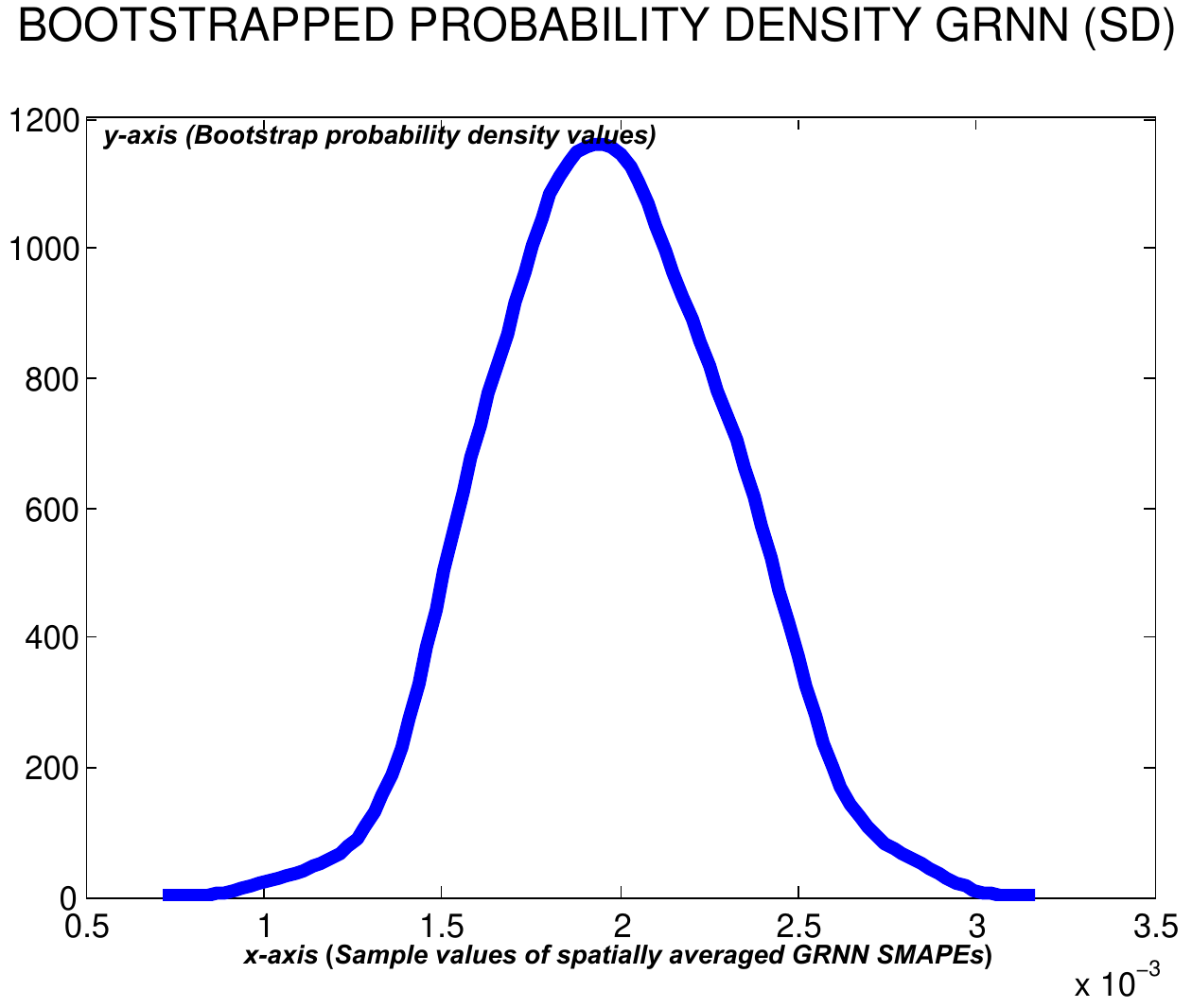}
 \vspace*{0.3cm}
 \includegraphics[width=5cm,height=5.5cm]{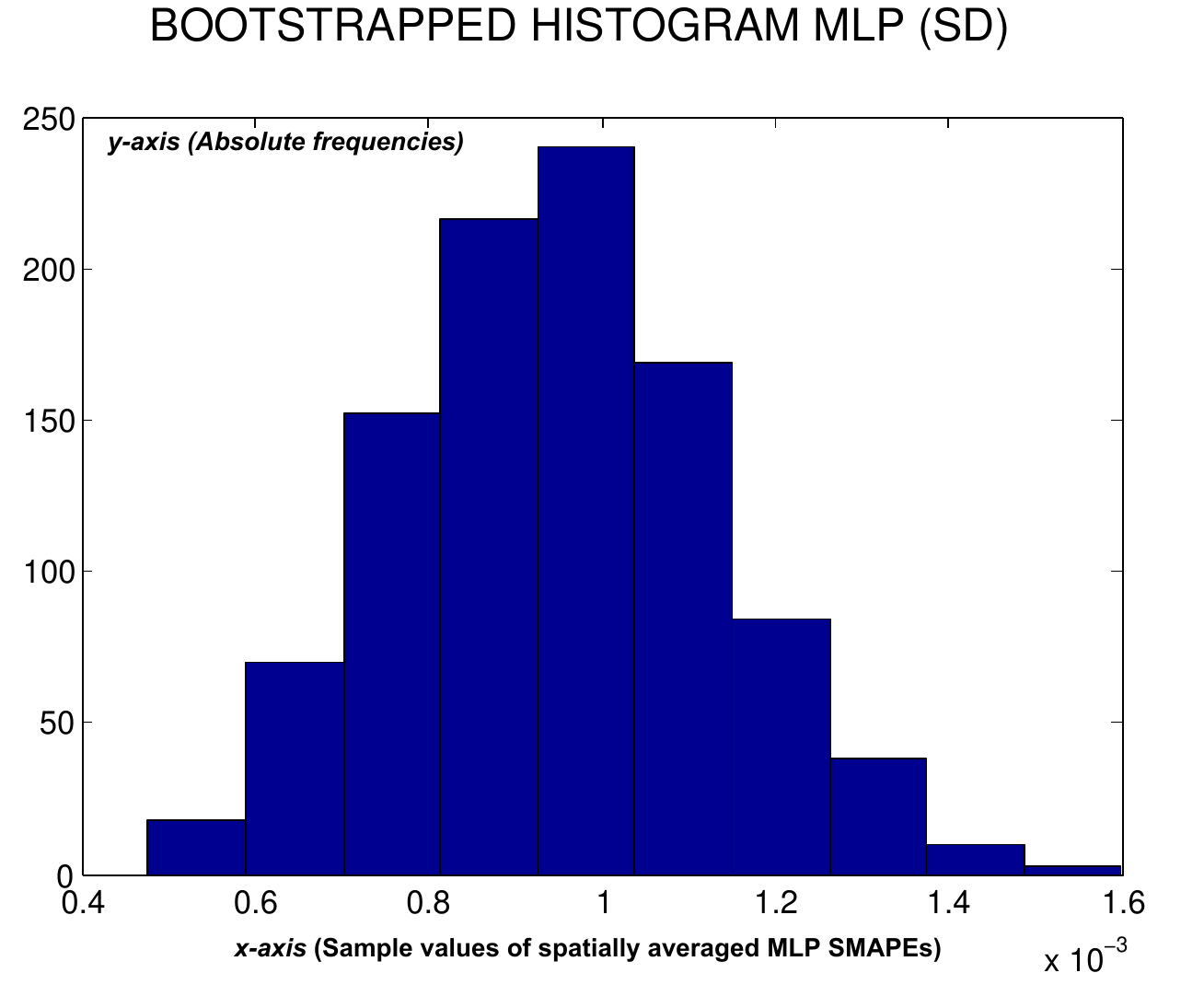}
 \includegraphics[width=5cm,height=5.5cm]{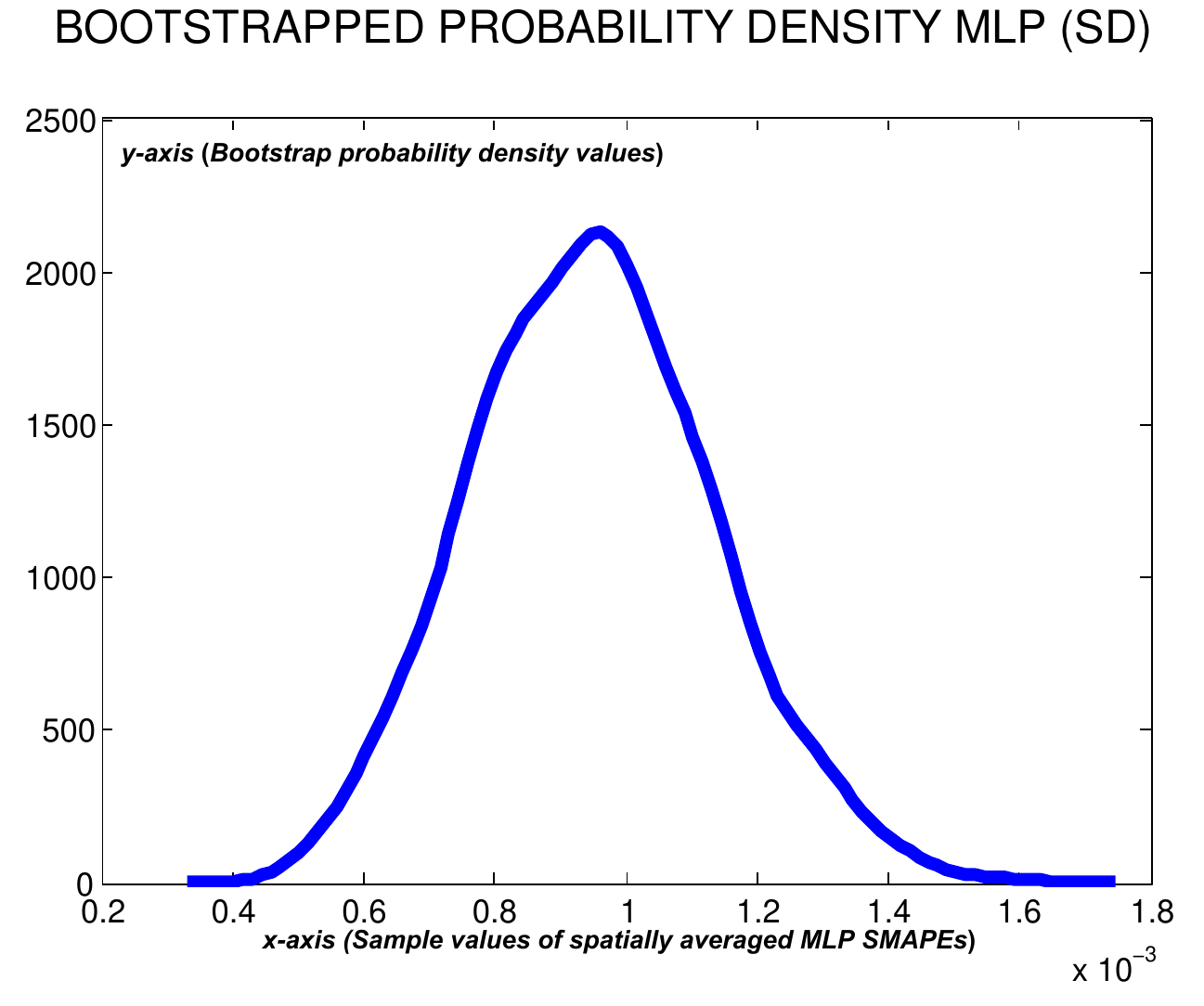}
 \vspace*{0.3cm}
 \includegraphics[width=5cm,height=5.5cm]{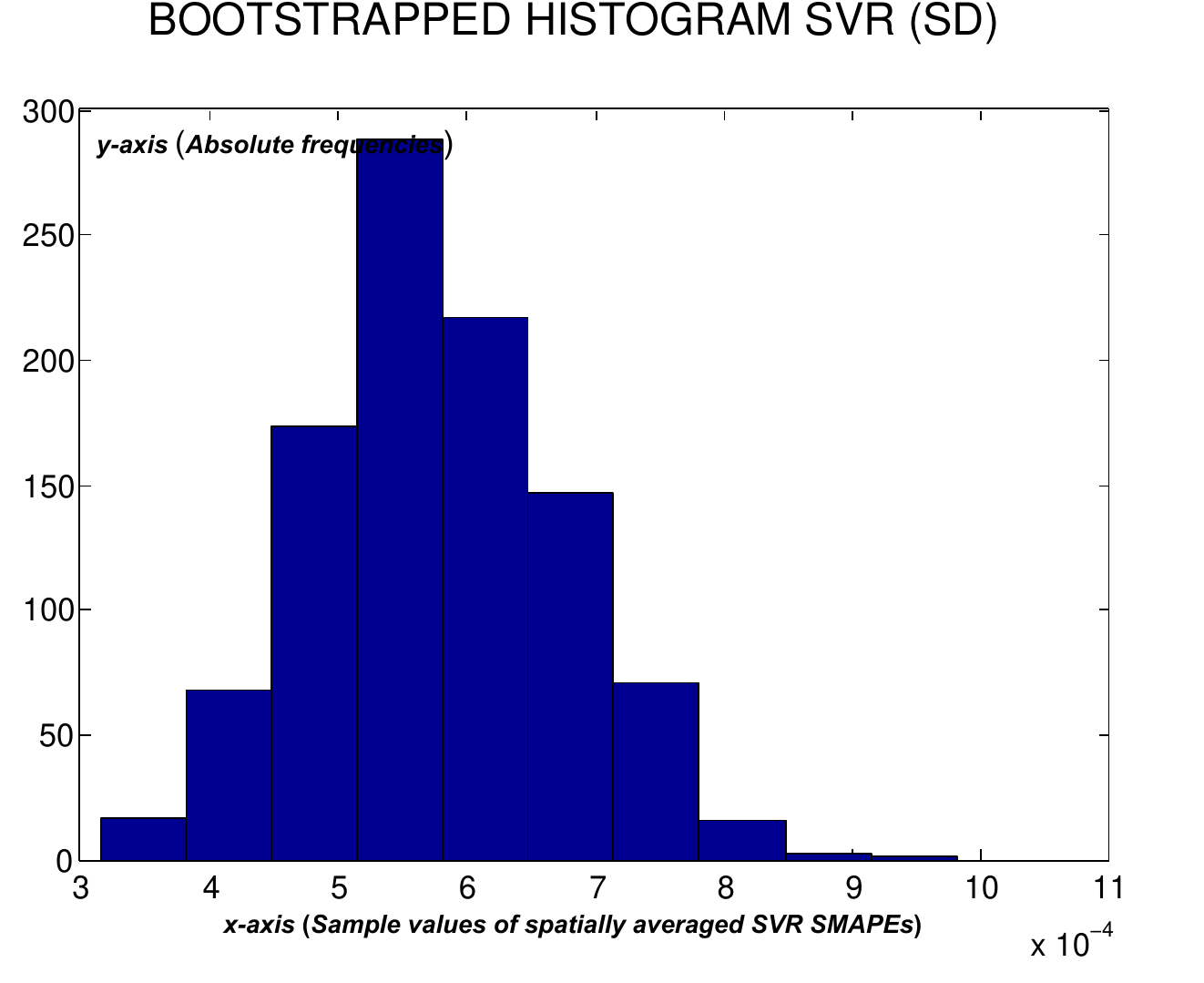}
 \includegraphics[width=5cm,height=5.5cm]{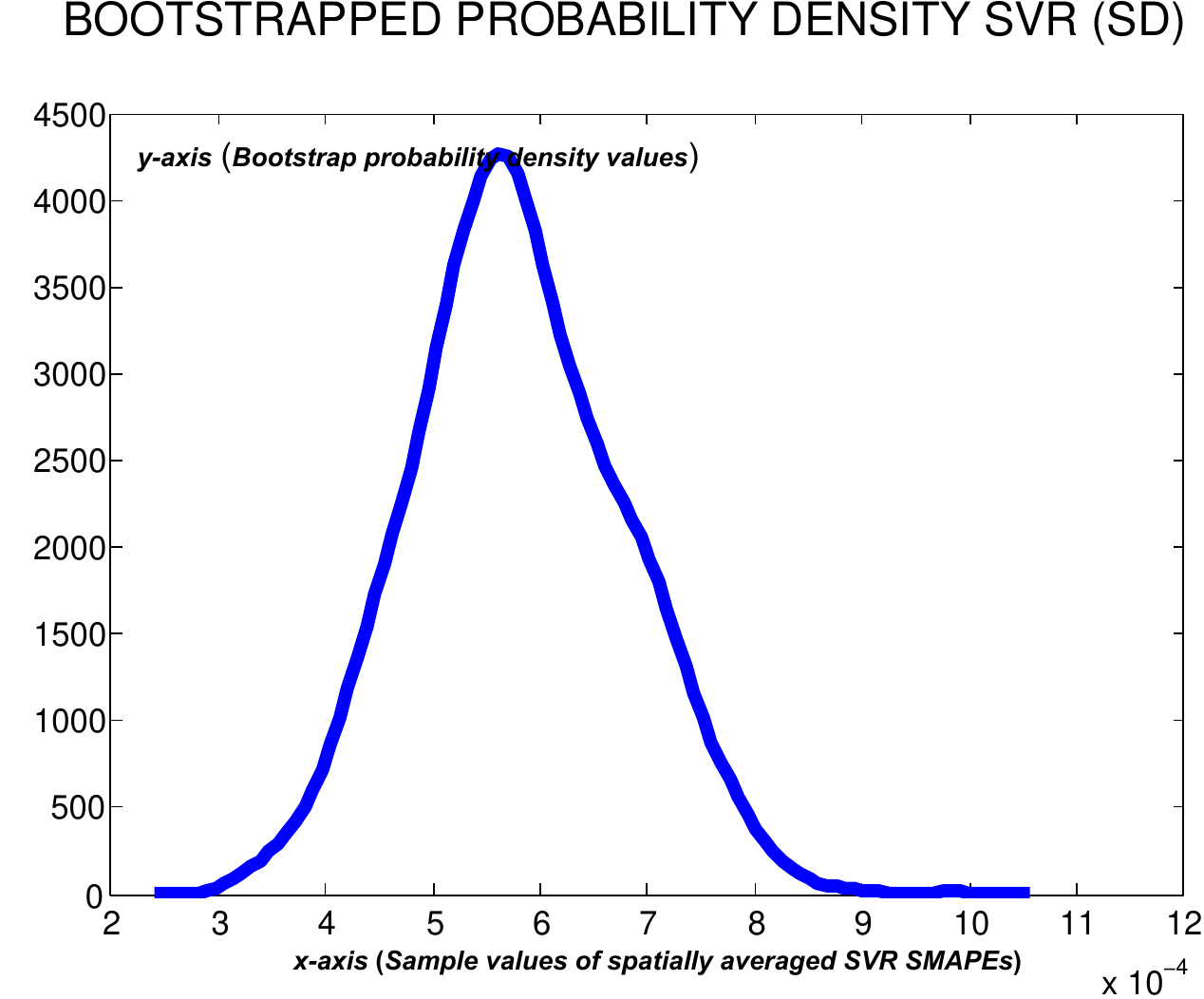}
 \caption{\textbf{\emph{Soft--data category}}. From $1000$ bootstrap  samples, spatially averaged SMAPEs  histograms and probability
 densities are plotted, for GRNN (top), MLP (center) and non--linear  SVR (bottom)}
 \label{figrmbesd}
 \end{figure}
\end{center}

\begin{center}
 \begin{figure}[!h]
 \centering
 \includegraphics[width=5cm,height=5.5cm]{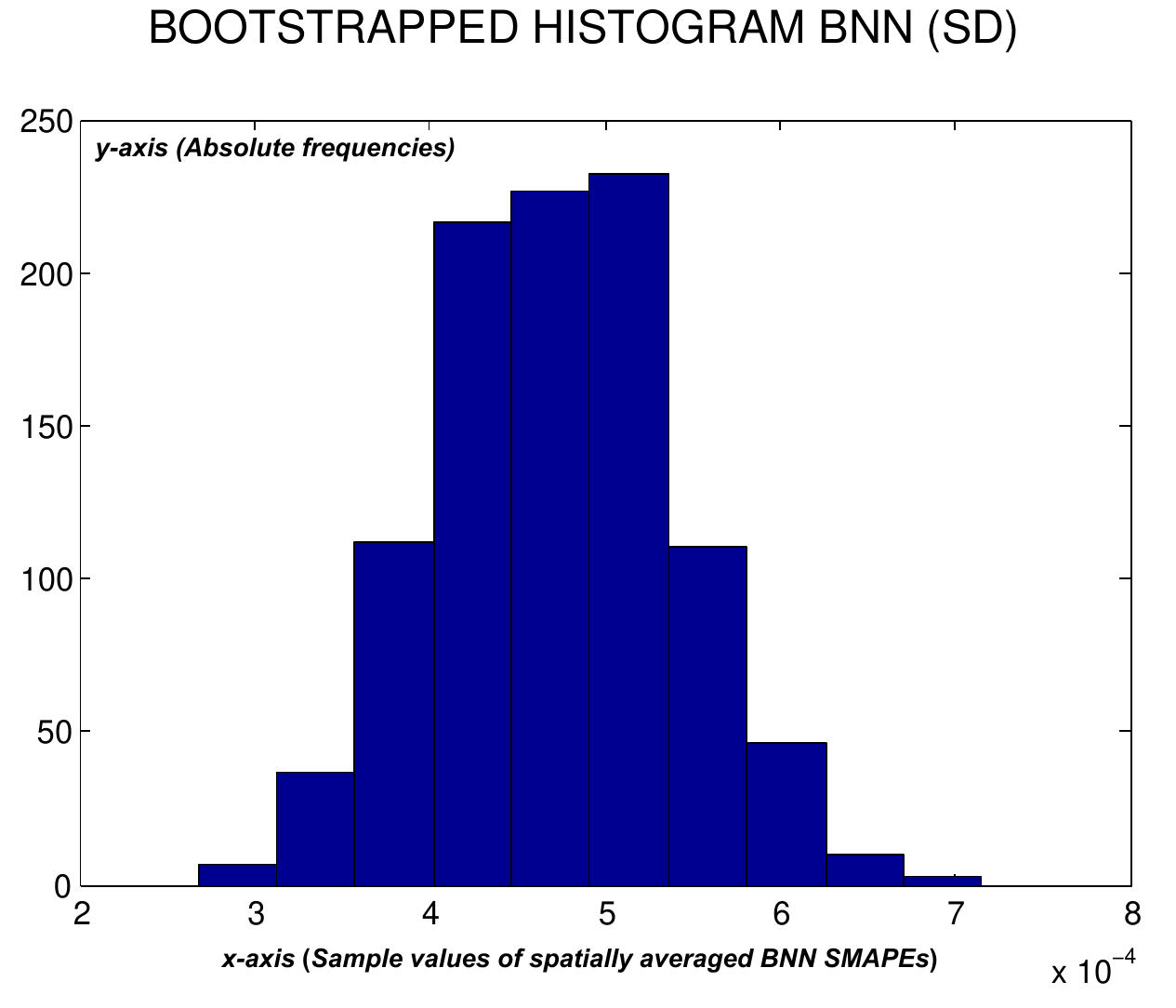}
 \includegraphics[width=5cm,height=5.5cm]{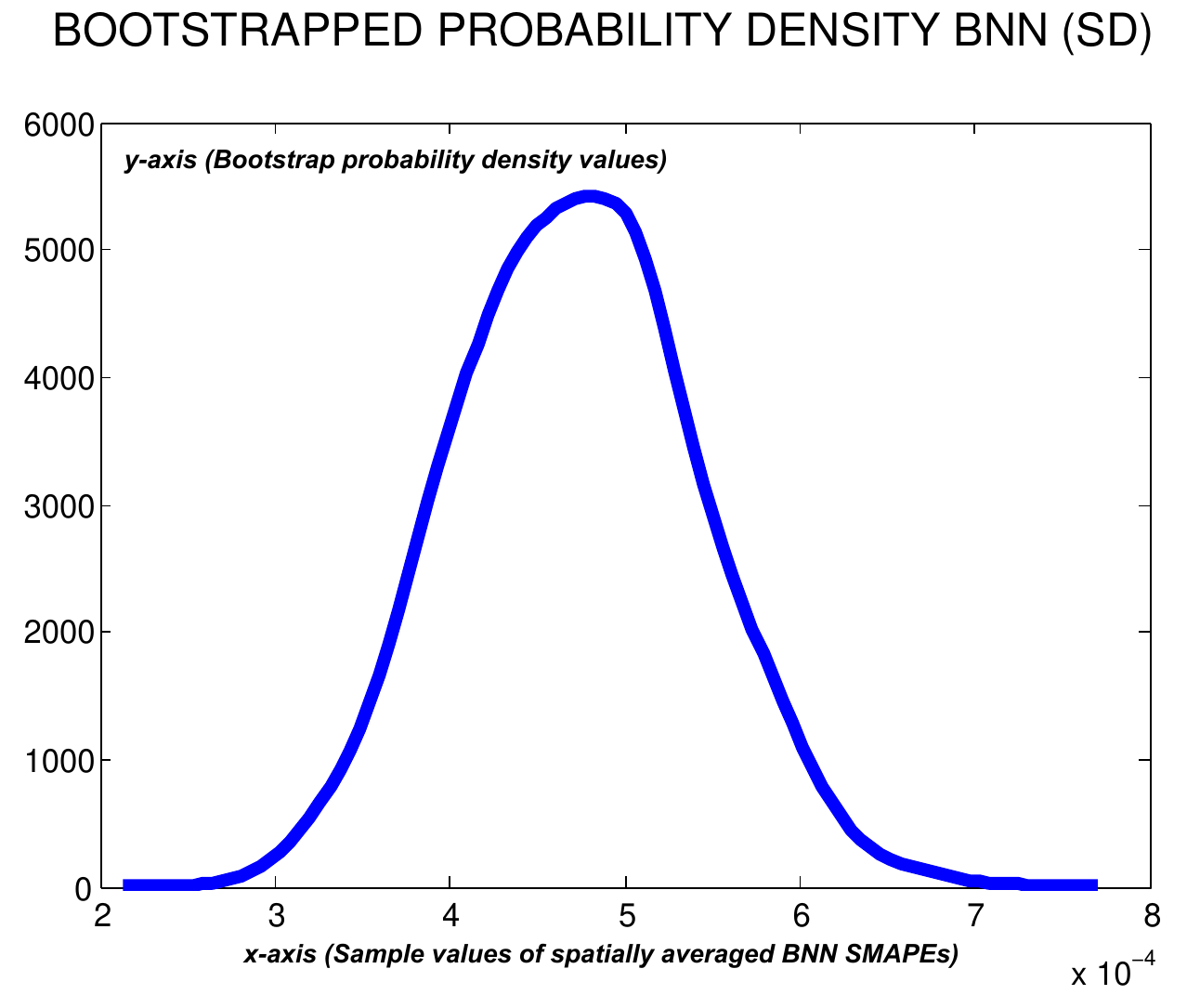}
 \vspace*{0.3cm}
 \includegraphics[width=5cm,height=5.5cm]{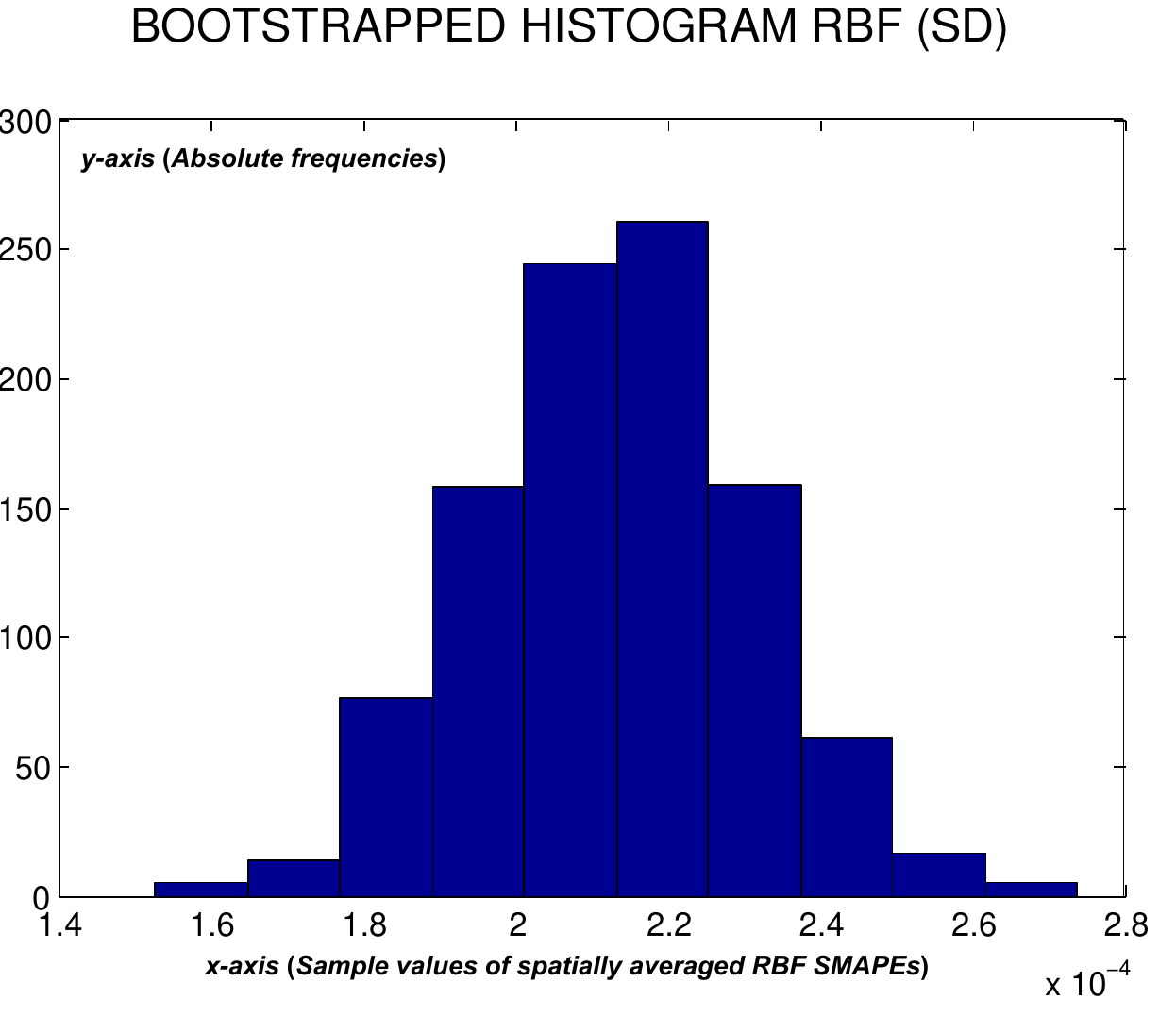}
 \includegraphics[width=5cm,height=5.5cm]{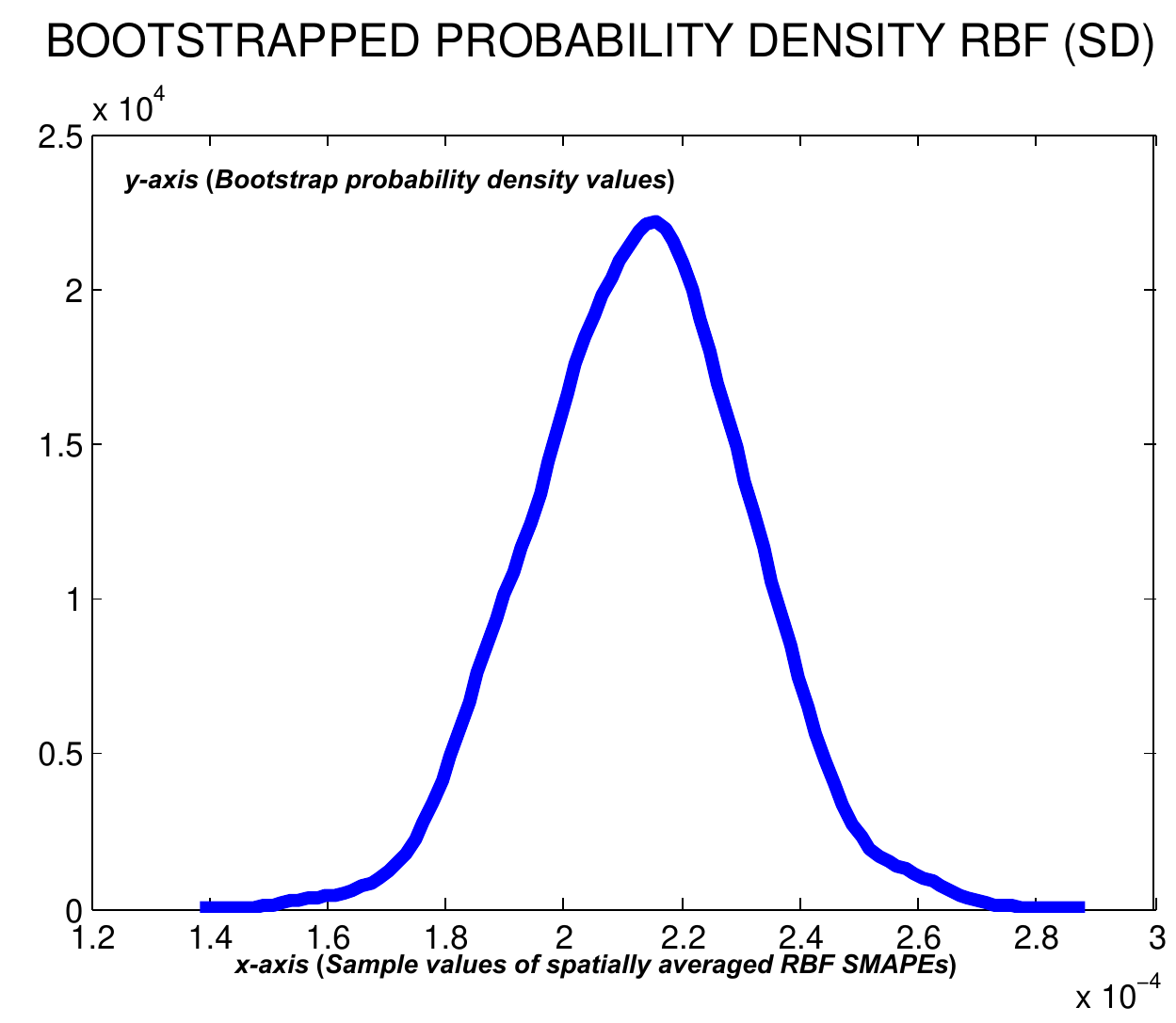}
 \vspace*{0.3cm}
 \includegraphics[width=5cm,height=5.5cm]{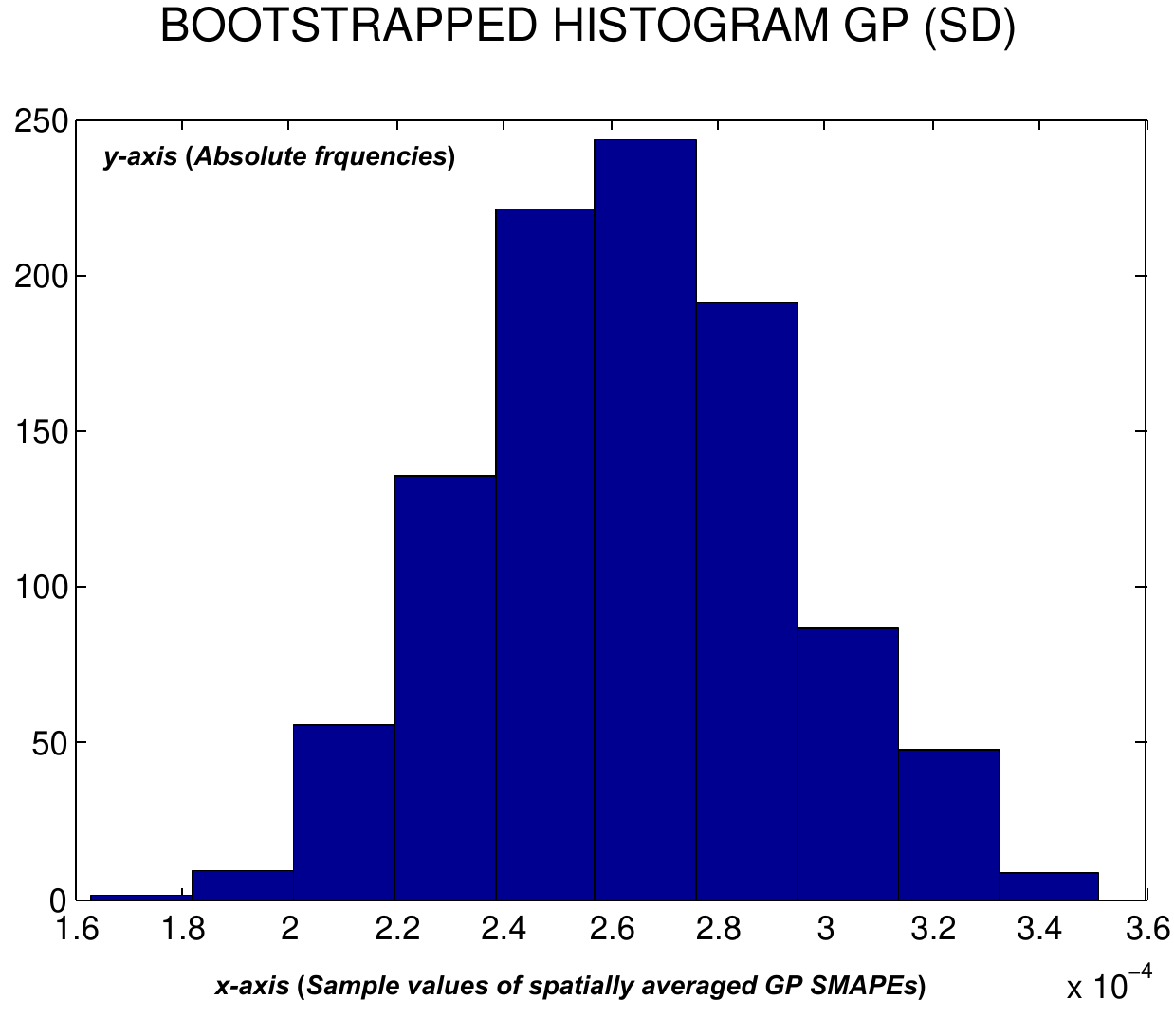}
 \includegraphics[width=5cm,height=5.5cm]{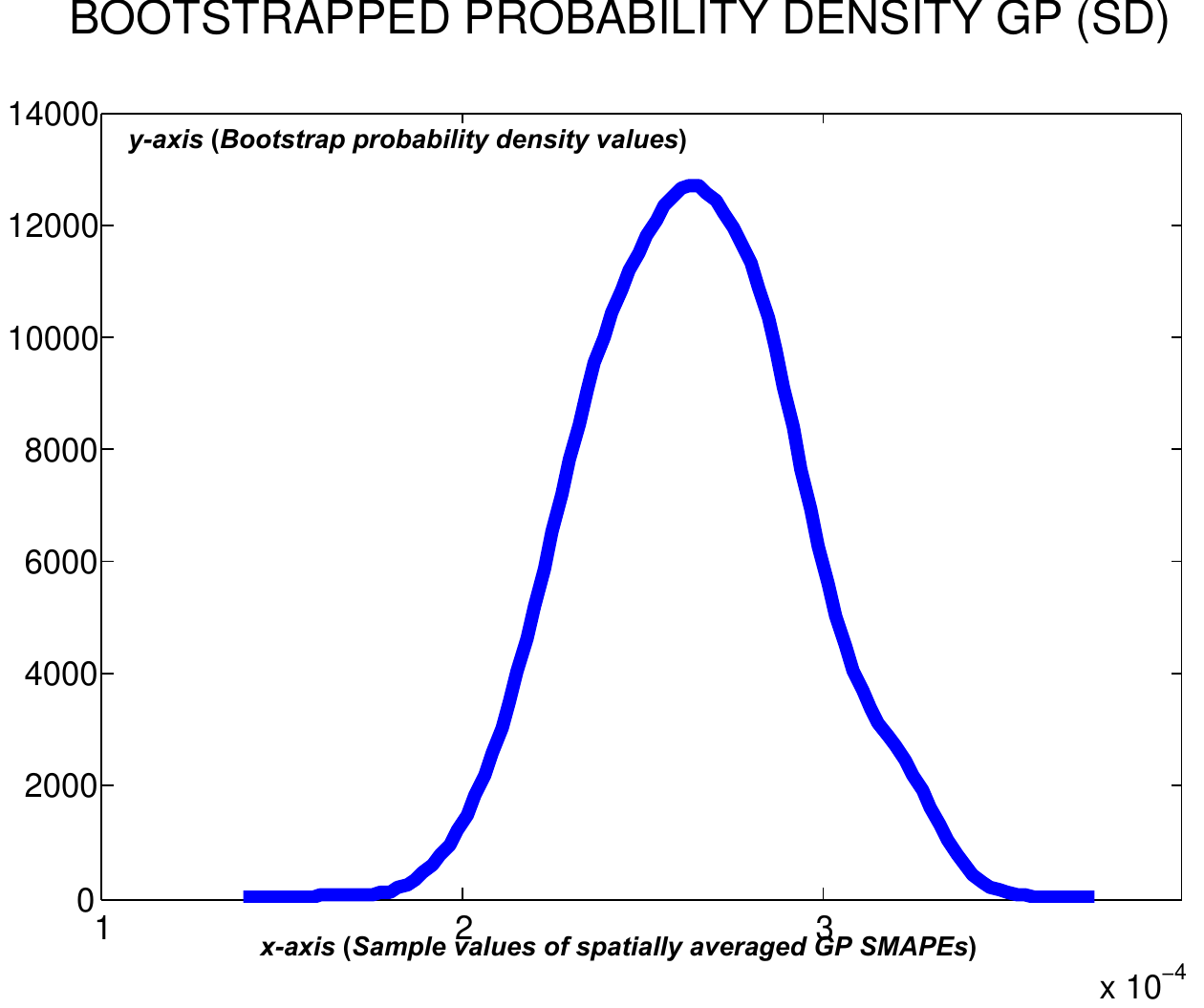}
 \caption{\textbf{\emph{Soft--data category}}. From $1000$ bootstrap  samples, spatially averaged SMAPEs  histograms and probability
 densities are plotted, for BNN (top), RBF (center) and  GP (bottom)}
 \label{figrmbe2sddd}
 \end{figure}
\end{center}

\begin{center}
 \begin{figure}[!h]
 \centering
\includegraphics[width=5cm,height=5.5cm]{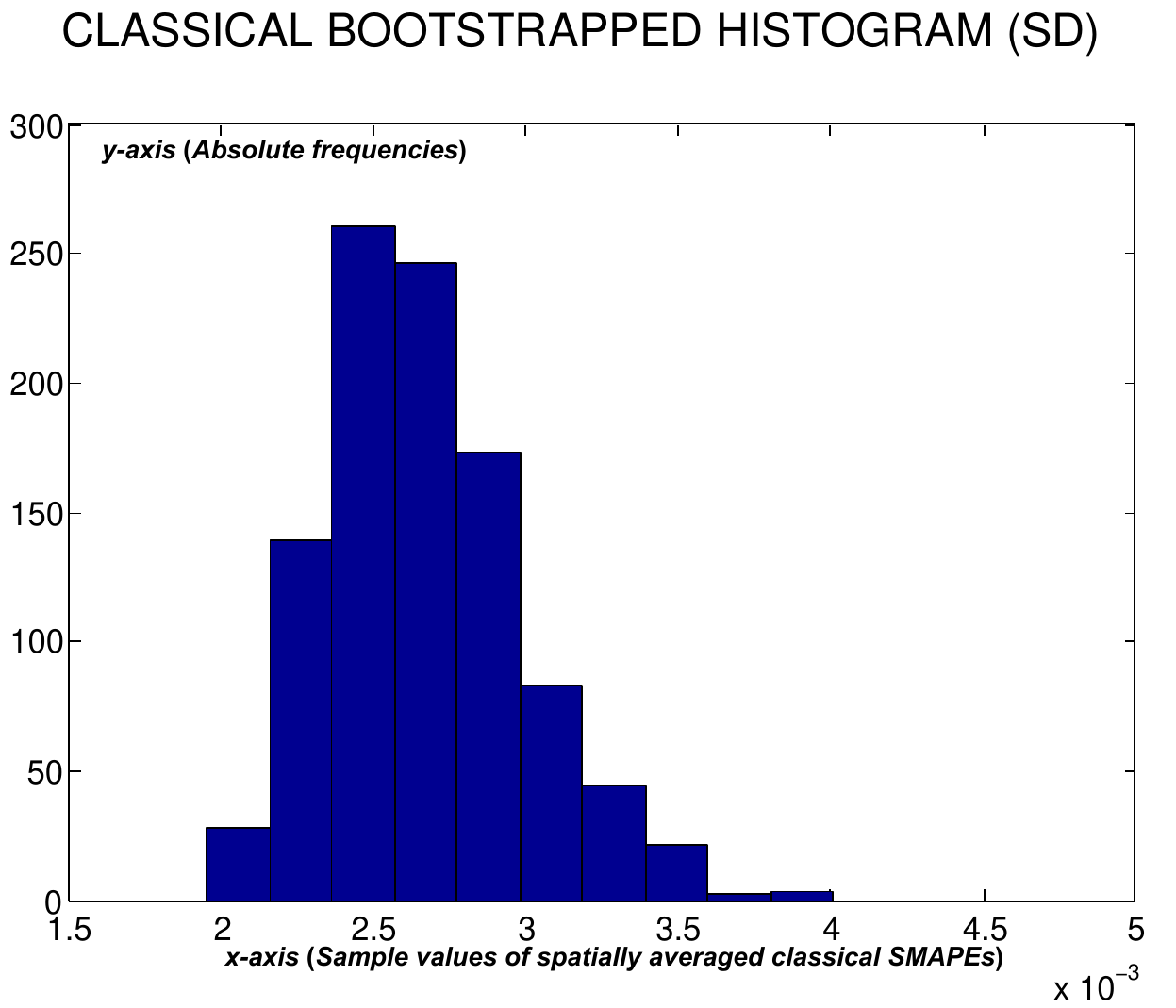}
\includegraphics[width=5cm,height=5.5cm]{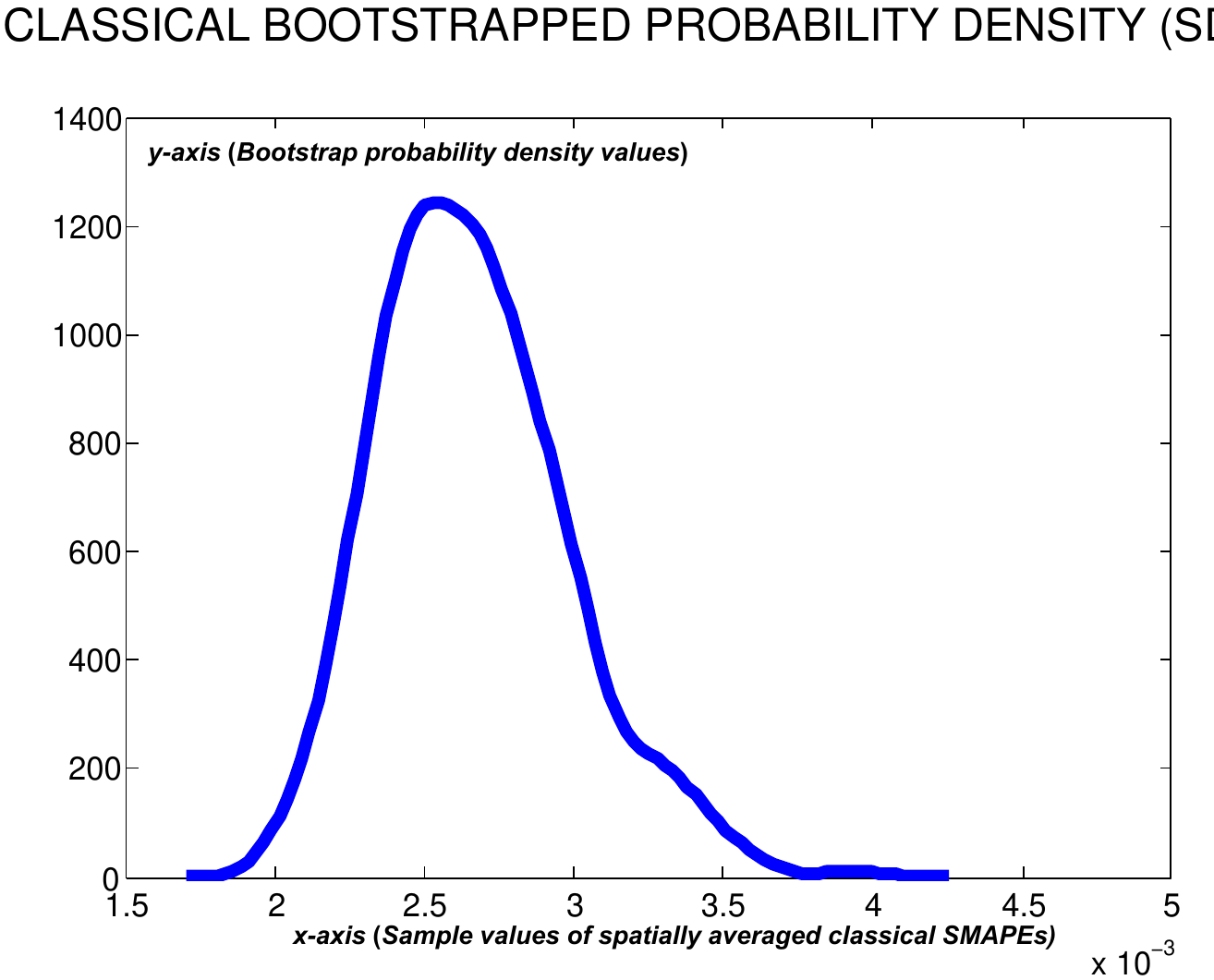}

\medskip

\includegraphics[width=5cm,height=5.5cm]{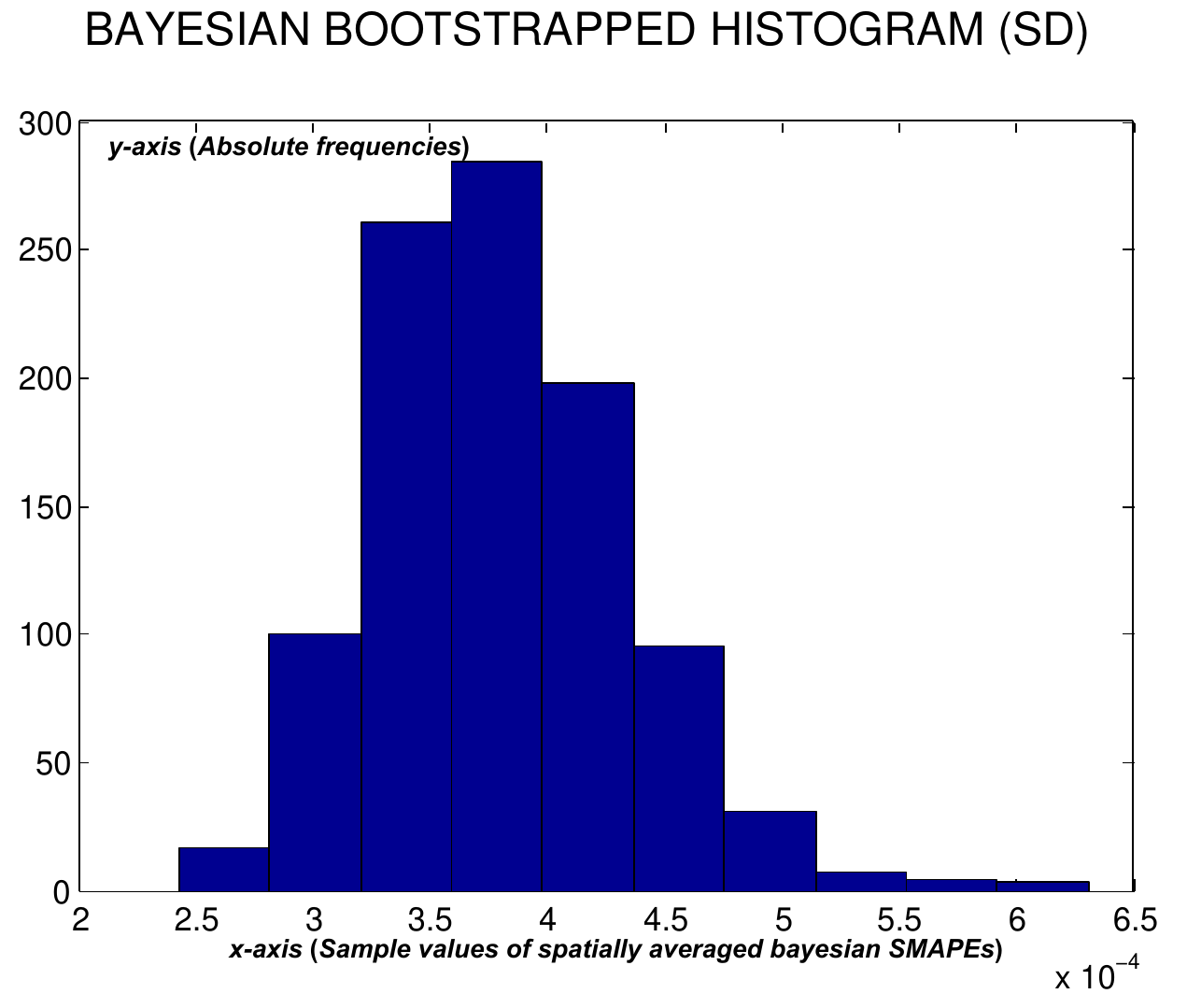}
\includegraphics[width=5cm,height=5.5cm]{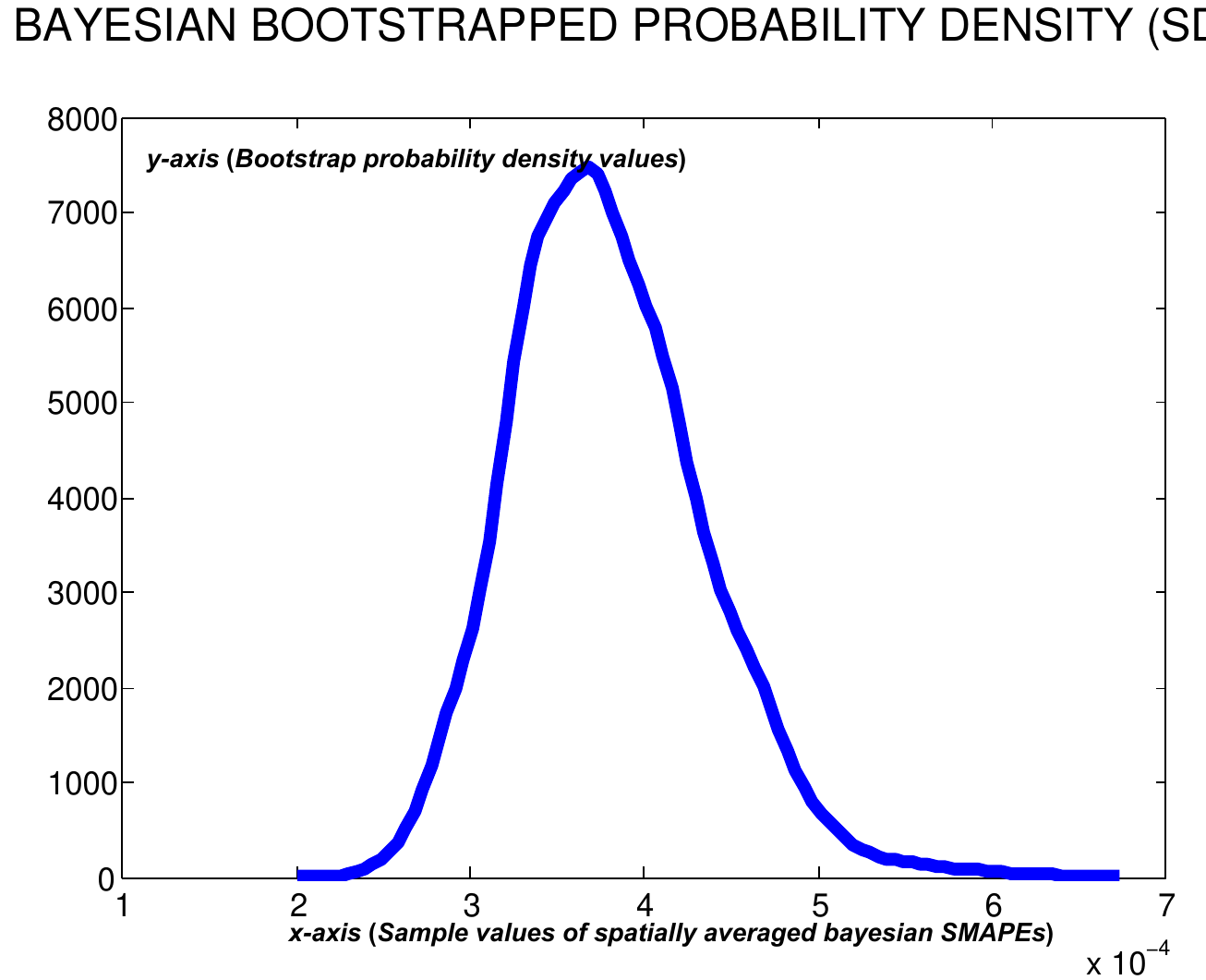}
\caption{\textbf{\emph{Soft--data category}}. From $1000$ bootstrap  samples, spatially averaged SMAPEs  histograms and probability
 densities are plotted, for trigonometric regression, combined with empirical--moment based classical  (top),
 and Bayesian  (bottom)
 residual prediction }
\label{figrmbe3sd}
 \end{figure}
\end{center}
\clearpage
\section{Final comments}
\label{CR}
    One can observe the agreement between the respective performance--based model classification  results, obtained from   random $k$--fold cross--validation, and  bootstrap estimation in Sections \ref{ecs} and \ref{BCR}.   In the hard--data category,
     the best performance is displayed by RBF and GP. Similar bootstrapping characteristics are observed for BNN and SVR,  with slightly larger values of  spatially averaged SMAPEs,  reflected  in the     location of the mode, in the histograms and probability  densities displayed in Figures \ref{figrmbe} and
\ref{figrmbe2}. These four regression methodologies show a similar degree of variability, regarding the spatially averaged SMAPEs sample values. A higher variability than RBF, GP, BNN and SVR is displayed by  the bootstrap sample values of  spatially averaged SMAPEs in MLP validation. MLP bootstrapped  mode is also slightly shifted to the right. The worst performance corresponds to GRNN (see also  Table \ref{T2}). In the soft--data category, where our approach is incorporated to the empirical comparative study, almost the same empirical ML model ranking holds. Some
differences are found  in the bootstrap confidence intervals, and histogram and probability densities computed. For instance, GRNN seems to be favored by soft--data category, while MLP displays worse performance in this category. Hence, smaller differences between GRNN and MPL are displayed in the soft--data category. A slightly improvement in the soft--data  category of BNN relative to SVR is observed, preserving almost the same performance.  RBF and GP  display  better performance in the  soft--data category, being RFB  a bit superior to GP in this category (see Table \ref{tbbihhsd}, and Figure \ref{figrmbe2sddd}). The  trigonometric regression, and multivariate time series residual prediction approach based on the  empirical moments  displays  similar results to GRNN, with slightly  better performance of GRNN, observed in the bootstrap intervals and histogram/probability density (see Figures \ref{figrmbesd} and \ref{figrmbe3sd}). However, as given in  Figures \ref{figrmbe2sddd} and \ref{figrmbe3sd},  the trigonometric regression and Bayesian residual prediction presents almost the same `performance as  BNN, with some slightly better probability distribution features of our approach respect to BNN (see also bootstrap intervals). Our approach is less affected by the random splitting of the sample, in the implementation of the random $k$--fold cross validation procedure, since  a dynamical spatial residual  model is fitted in a second (objective) step. Thus, the proposed multivariate time series classical and Bayesian regression residual modeling  fits the short--term spatial linear correlations displayed by the soft--data category. However, the price we pay  for increasing model complexity is reflected in the resulting SMAPEs based random $k$--fold and bootstrap  model classification results obtained.

The spatial
component  effect  is reflected in Tables     \ref{T2} (hard--data), where spatial heterogeneities displayed by random $10$-fold cross--validation \linebreak SMAPEs errors are observed (see also Table 3 in the Supplementary Material). While  Table  \ref{T5} (see also  Table 4 in the Supplementary Material)  reveals the benefits obtained in some of the ML regression models tested  from soft--data information. Particularly, in this category, possible spatial linear correlations  are incorporated to the analysis, in terms of  soft--data.

 \section*{Supplementary material}
 We briefly describe the implementation of  ML models in the hard-- and soft--data categories.

\subsection*{Multilayer Perceptron (MLP)} MLP shares the philosophy  of nonlinear regression, in terms of a  link function $g,$  the hidden node output, defining the following approximation of the response:
\begin{equation}
\widehat{y}=\eta_{0}+\sum_{k=1}^{NH}\eta_{k}g(\beta_{k}^{T}\mathbf{x}),\label{MP}
\end{equation}
\noindent  from the \emph{input} vector
$\mathbf{x}^{T}=(\mathbf{1},\mathbf{x})$ augmented with $1,$ and the
weight vector $\beta_{k}$ associated with the $k$th  hidden node,
$k=1,\dots,NH,$  defining  $\boldsymbol{\beta
}=(\beta_{1}^{T},\dots,\beta_{NH}^{T})^{T}.$
Usually, the logistic function $g(u)= \frac{1}{1+\exp(-u)}$  is considered.
The hidden node outputs are
also weighted by the components  of the vector $(\eta_{0},\dots, \eta _{NH}).$
In this context, $\widehat{\mathbf{y}}$ is usually referred as the
network output. MLP allows the approximation of any given continuous
function on a compact set, from a given  network with
a finite number of hidden nodes. Optimization algorithms are applied
to obtain the weights from the least--squares loss function. From (\ref{MP}),
 our implementation of MLP from soft--data is given by
\begin{equation}
\widehat{y}_{m}=\widehat{y}(\mathbf{x}_{m})=\widehat{\ln\left(\lambda_{t+1}\right)
(\phi_{m})}=\eta_{0}+\sum_{k=1}^{NH}\eta_{k}g(\beta_{k}^{T}\mathbf{x}_{m}),\label{MPS}
\end{equation}
\noindent where, for $m=1,\dots,M(T),$  \begin{equation}\mathbf{x}_{m} =\left(\ln\left(\lambda_{t}\right)(\phi_{m} ),\dots,
\ln\left(\lambda_{t+1-j_{0}}\right)(\phi_{m})\right)^{T}.\label{iv}\end{equation}\noindent   Parameter $j_{0}$ refers to
    the number of temporal lags incorporated in the prediction. The truncation parameter $M(T)$ plays a similar role  to parameter $k(T)$ in the paper.
Here, $T$ denotes the number of temporal training nodes.

\subsection*{Radial Basis Function Neural Network (RBF)} RBF  works with   node   functions,  depending on a center and scale parameters, fitting the local smoothness of the response.  Specifically,  from an   initial  blank
network, the nodes are sequentially added, around the training
pattern, until an acceptable  error  is reached. All the output
layer weights are then recomputed using the least squares formula.
Gaussian functions have been widely selected as node functions. Particularly, in this case, our implementation from soft--data   has been achieved
from the formula:
  For $m=1,\dots,M(T),$
\begin{equation}
\widehat{y}_{t+1}(\mathbf{x}_{m})=\widehat{\ln\left(\lambda_{t+1}\right)
(\phi_{m})}=\sum_{j=1}^{NH}\eta_{j}\exp\left(\frac{\|\mathbf{x}_{m}-c_{j}\|^{2}}{\beta^{2}}\right)\label{rbf}
\end{equation}
\noindent where, as usual,  the weight parameters  $\eta_{j},$  $j=1,\dots,NH,$
define the linear combination of radial basis
functions. Here,  the scalar spread parameter $\beta,$ and the  vector parameters $c_{j},$ $j=1,\dots,NH,$  respectively
provide the width, and the  centers  of the node functions. The input vectors are  defined as in  (\ref{iv}).

\subsection*{Support Vector Regression (SVR)} SVR implementation involves a loss function leading to a
balance between model complexity and precision (accurate
prediction). A bias parameter $b$ is also considered in its
formulation as reflected in the following equation:
\begin{equation}
y=f(\mathbf{x})=\boldsymbol{\beta }^{T}\mathbf{x}+b\label{svm}
\end{equation}
\noindent where the loss function
\begin{equation}
\mathcal{L}=\frac{1}{2}\|\boldsymbol{\beta } \|^{2}+C\sum_{m=1}^{M}
\left|y_{m}-f(x_{m})\right|_{\epsilon},\label{lssfff}
\end{equation}
\noindent is considered. Here,   $x_{m}$ and $y_{m}$ respectively
denote the $m$th training input vector and the target output, for
$m=1,\dots, M,$  and
\begin{eqnarray}&&\hspace*{-1cm}\left|y_{m}-f(x_{m})\right|_{\epsilon}=\max\left\{0,\left|y_{m}-f(x_{m})\right|-\epsilon\right\}.\nonumber\end{eqnarray}
\noindent Thus, the errors below $\epsilon $ are not penalized. The
solution to the optimization problem associated with the loss
function (\ref{lssfff}) is obtained from the corresponding gradient of the Lagrangian function, involving  Lagrange multipliers, that determine the optimal weights from the training data points.

From equation (\ref{svm}), in the soft--data category,  we solve the constrained optimization problem:
\begin{eqnarray}
&&y_{t+1}(\mathbf{x}_{m})=f(\mathbf{x}_{m})=\boldsymbol{\beta }^{T}\mathbf{x}_{m}+b \nonumber\\
&&\mathcal{L}=\frac{1}{2}\boldsymbol{\beta
}^{T}\boldsymbol{\beta
}+C\sum_{m=1}^{M(T)}\left|\ln\left(\lambda_{t+1}\right)
(\phi_{m})-\sum_{j=1}^{j_{0}}\ln(\lambda_{t+1-j})(\phi_{m})\beta_{j}-b\right|_{\epsilon}\nonumber\\
&&=\frac{1}{2}\boldsymbol{\beta
}^{T}\boldsymbol{\beta
}+C\sum_{m=1}^{M(T)}\left|y_{m}-f(\mathbf{x}_{m})\right|_{\epsilon},\label{lssf}
\end{eqnarray}
\noindent where,  for $m=1,\dots,M(T),$ the imput vector  $\mathbf{x}_{m}$ is defined as in (\ref{iv}).
\subsection*{Bayesian Neural Network (BNN)}
The design of BNN involves  Bayesian estimation, and the concept of
regularization. The  network parameters or weights  are considered
random variables,  following a prior probability distribution.
Smooth fits are usually favored in the selection of the  prior
probability of the weights to reduce model complexity.
 The posterior probability distribution of the
weights is obtained after data are observed.  The network prediction
is then  computed. Specifically, the optimal prediction is obtained
by minimizing the following expression:
\begin{equation}
J=\nu E_{\mathbf{O}}+(1-\nu )E_{\mathbf{W}}, \label{BNNof}
\end{equation}
\noindent where $E_{\mathbf{O}}$ is the sum of the square errors in
the network output, based on the posterior distribution of the
parameters,  $E_{\mathbf{W}}$ represents the sum of  the squares of
the  weights or the network parameters, and $\nu \in (0,1)$ denotes
the regularization parameter. Let $L$ be the number of parameters.
An $L$--dimensional  Gaussian prior probability distribution is
usually assumed for the network parameters with zero--mean and
variance--covariance matrix $\frac{1}{2(1-\nu )}I_{L\times L},$ with
$I_{L\times L}$ denoting the $L\times L$ identity matrix. Thus,
\begin{equation}
p(\mathbf{w})=\left[\frac{1-\nu}{\pi}\right]^{L/2}\exp\left(-(1-\nu
)E_{W}(\mathbf{w})\right). \label{pdf}
\end{equation}

 \noindent This prior puts more weight onto small network parameter values close to zero. The posterior probability density, given the observed data $\mathbf{O}=\mathbf{o},$ and the value $\nu $ of the regularization parameter, is defined as
\begin{equation}
p(\mathbf{w}/\mathbf{o},\nu
)=\frac{p(\mathbf{o}/\mathbf{w},\nu)p(\mathbf{w}/\nu )
}{p(\mathbf{o}/\nu )}. \label{ppd}
\end{equation}

Considering that the errors are also Gaussian distributed, the
conditional probability of the data $\mathbf{O}$ given the
parameters $\nu $ and $\mathbf{w},$ is obtained as
\begin{equation}
p(\mathbf{o}/\mathbf{w},\nu
)=\left(\frac{\nu}{\pi}\right)^{M/2}\exp\left(-\nu
E_{\mathbf{O}}(\mathbf{o})\right), \label{pde}
\end{equation}
\noindent where $M$ denotes the number of training data points. From
equations  (\ref{pdf})--(\ref{pde}),
\begin{equation}
p(\mathbf{w}/\mathbf{o},\nu )=c\exp\left(-
J(\mathbf{w},\mathbf{o},\nu)\right), \label{pde2}
\end{equation}
\noindent where $c$ denotes the normalizing constant. The
conditional probability of the parameter $\nu  $ given de observed
data $\mathbf{O}=\mathbf{o}$ is also computed under a Bayesian
framework as
\begin{equation}
p(\nu /\mathbf{o})=\frac{P(\mathbf{o}/\nu )p(\nu
)}{p(\mathbf{o})}.\label{almix}
\end{equation}

Equations  (\ref{pde2}) and (\ref{almix}) are maximized to obtain
the optimal weights and the regularization parameter $\nu,$
respectively.

In our  soft--data implementation from  (\ref{iv}), the
corresponding optimization problem is formulated by conditioning to
the empirical spatial projections. Note that, in the selected Gaussian prior
probability framework, the error projections are also Gaussian, and
our choice of the function  basis, diagonalizing the empirical
autocovariance operator  of the errors,  leads to a projected error
vector with independent Gaussian components, suitable  normalized by
the empirical eigenvalues. Hence, optimization from equations
 (\ref{pde2}) and (\ref{almix}) can be implemented in a  similar way to hard--data BNN.

\subsection*{Generalized Regression Neural Network (GRNN)} GRNN is based on kernel regression.
The kernel estimator is computed from the weighted sum of the
observed responses, or target outputs associated with the training
data points in a neighborhood of the objective data point $x,$ where
prediction must be computed. Thus, the training data points are selected  in the vicinity of the given  objective point $x.$ Specifically, the following formula is applied in
  the approximation of the response value at the point $x:$
\begin{equation}
\widehat{y}(x)=\sum_{j=1}^{T}\frac{\mathcal{K}\left(\frac{\|x-x_{j}\|}{h}\right)}{\sum_{l=1}^{T}\mathcal{K}\left(\frac{\|x-x_{l}\|}{h}\right)}y_{j},\label{ke}
\end{equation}
\noindent where  $y_{j}$  is the target output for training data
point $x_{j},$ for $j=1,\dots,T,$  and $\mathcal{K}$ is the kernel
function. Usually an isotropic rapidly decreasing kernel function is
considered, e.g., the Gaussian kernel
$\mathcal{K}(u)=\exp\left(-u^{2}/2\right)/\sqrt{2\pi}$  constitutes
a common choice. The  bandwidth parameter $h$ defines  the
smoothness of the fit. Thus, $h$ controls the size of the smoothing
region. Hence, large values of $h$ correspond to a stronger
smoothing than the smallest values allowing a larger degree of local
variation. In our  soft--data implementation of (\ref{ke}), denote by $\Phi_{M(T)} (H),$ the subspace of $H$ obtained by  projection of functions in $H$  onto $\{\phi_{1},\dots,\phi_{M(T)}\},$
and $\mathbf{y}_{t}=\left(\ln(\lambda_{t})(\phi_{1}),\dots, \ln(\lambda_{t})(\phi_{M(T)})\right)^{T},$ $t\geq 1,$ then,
\begin{equation}
\widehat{\mathbf{y}_{t+1}}=\sum_{j=1}^{j_{0}-1}\frac{\mathcal{K}\left(\frac{\|\mathbf{y}_{t}-\mathbf{y}_{t-j}\|_{\Phi_{M(T)}(H)}}{h}\right)}{\sum_{l=1}^{j_{0}-1}
\mathcal{K}\left(\frac{\|\mathbf{y}_{t}-\mathbf{y}_{t-l}\|_{\Phi_{M(T)}(H)}}{h}\right)}\mathbf{y}_{t-j+1},\label{kebb}
\end{equation}
\noindent where, as before,  parameter $j_{0}$ refers to
    the number of temporal lags incorporated in the prediction.
\subsection*{Gaussian Processes (GP)}

A good performance  is usually observed in the implementation of GP
regression, based on  the multivariate normal probability
distribution assumption, characterizing the observed responses at
the different training data points. Specifically, we consider the
observation model
\begin{equation}
\mathbf{Y}=\mathbf{Z}+\boldsymbol{\epsilon}, \label{observationmod}
\end{equation}
\noindent where the additive noise vector  $\boldsymbol{\epsilon}$ is independent of $\mathbf{Z},$ and
has independent and identically distributed zero--mean  Gaussian components with variance $\sigma^{2}_{\boldsymbol{\epsilon}}.$ A multivariate normal distribution of
the  random vector
$\mathbf{Z}\sim\mathcal{N}(\mathbf{0},\boldsymbol{\Sigma}),$ with
covariance matrix  $\boldsymbol{\Sigma}_{\mathbf{Z}}(X,X)$ is assumed.
This matrix provides the variances and covariances between the function values  $Z(x_{i}),$ and $Z(x_{j}),$ $i,j=1,\dots, N,$ at the training data points $X=(x_{1},\dots, x_{N}).$ The conditional Gaussian distribution of $\mathbf{Z},$ given $\mathbf{Y},$ leads  to the solution to the inverse estimation problem (\ref{observationmod}).
Thus,
 the estimation of $\mathbf{Z},$ for a   given input vector $\mathbf{x}_{\star},$ is obtained as  \begin{equation}
\widehat{\mathbf{Z}}_{\mathbf{x}_{\star}}=E\left[\mathbf{Z}/
\mathbf{y},\mathbf{x}_{\star},X\right]=
\boldsymbol{\Sigma}_{\mathbf{Z}}(\mathbf{x}_{\star},X)\left[\boldsymbol{\Sigma}_{\mathbf{Z}}(X,X)+\sigma^{2}_{\boldsymbol{\epsilon}}\mathbf{I}\right]^{-1}\mathbf{y}.
\label{brule}
\end{equation}

 In the soft--data category, an alternative multivariate implementation is achieved in terms of projection $\Phi_{M(T)}$ onto $\{\phi_{1},\dots, \phi_{M(T)}\}.$
 Hence,    $\boldsymbol{\Sigma}_{\mathbf{Z}}$ is replaced by the matrix  covariance operator   \begin{eqnarray}
 &&\hspace*{-0.5cm}R_{\mathbf{Z}}^{X,X}=\left(\begin{array}{ccc}
 \Phi_{M(T)}^{\star }\mathcal{R}_{1,1} \Phi_{M(T)} &  \dots &   \Phi_{M(T)}^{\star }\mathcal{R}_{1,T} \Phi_{M(T)}\\
 \Phi_{M(T)}^{\star }\mathcal{R}_{2,1} \Phi_{M(T)} &  \dots &   \Phi_{M(T)}^{\star }\mathcal{R}_{2,T}\Phi_{M(T)}\\
 \vdots & \dots &  \vdots \\
   \Phi_{M(T)}^{\star }\mathcal{R}_{T,1} \Phi_{M(T)} &    \dots &   \Phi_{M(T)}^{\star }\mathcal{R}_{T,T} \Phi_{M(T)}\\
  \end{array} \right),\nonumber\end{eqnarray}
 \noindent associated with the $T$ temporal training nodes considered, and the projection operator $\Phi_{M(T)}$ onto $\{\phi_{1},\dots,\phi_{M(T)}\},$ involved in the definition of our training soft--data points $X.$  Here, $\mathcal{R}_{i,j}= E\left[\left[\ln (\Lambda_{i})-E[\ln(\Lambda_{i})]\right]\otimes \left[\ln (\Lambda_{j})-E[\ln(\Lambda_{j})]\right]\right]\mathfrak{},$ $i,j=1,\dots,T,$ define the autocovariance and cross-covariance operators of the functional values  at the training data points. In this case,  $\sigma^{2}_{\boldsymbol{\epsilon}}\mathbf{I}$ becomes the diagonal matrix autocovariance operator $diag\left(R_{0,\boldsymbol{\epsilon}}\right)$ of the $H^{T}$--valued innovation process $\boldsymbol{\epsilon}=(\epsilon_{1},\dots,\epsilon_{T}).$ That is,
\begin{eqnarray}
&&\widehat{\mathbf{Z}}_{\mathbf{x}_{\star} }=E\left[\mathbf{Z}/\mathbf{y},
\mathbf{x}_{\star}, X, \Phi_{M(T)}\right]\nonumber\\ &&= R_{\mathbf{Z}}^{\mathbf{x}_{\star},X}\left[R_{\mathbf{Z}}^{X,X}+
\Phi_{M(T)}^{\star }diag\left(R_{0,\boldsymbol{\epsilon}}\right)\Phi_{M(T)}\right]^{-1}\Phi_{M(T)}(\mathbf{y}).\nonumber
\end{eqnarray}

\subsection*{Observed hard-- and soft--data  ML SMAPE sample values}

The  Symmetric Mean Absolute Percentage Errors (SMAPEs) associated with ML   estimation results, based on the overall functional sample, from
hard-- and soft--data, are respectively displayed in
Tables \ref{T3} and \ref{T6}. See also Figures  \ref{figestorlrcod0}--\ref{fig1proyd2} below, where the observed and estimated COVID--19 mortality log--risk and cumulative cases curves are respectively displayed.
\begin{table}[!h]
\caption{\textbf{\emph{Hard--data}}. As indicated, displayed values must be multiplied by $10^{-2}$} \label{T3}
\begin{center}
\begin{tabular}{|c|c|c|c|c|c|c|}
\hline
 {\bf SC} ($\mathbf{x10^{-2}}$)&
 {\bf GRNN} & {\bf MLP} & {\bf SVR} & {\bf BNN} & {\bf RBF} &  {\bf GP} \\
  \hline
C1 & 0.1964 & 0.0611 & 0.0665 & 0.0535 & 0.0483 & 0.0490 \\
C2 & 0.6117 & 0.1172 & 0.0588 & 0.0678 & 0.0609 & 0.0567 \\
C3 & 0.1565 & 0.0375 & 0.0328 & 0.0284 & 0.0300 & 0.0273 \\
C4 & 0.0969 & 0.0314 & 0.0129 & 0.0191 & 0.0167 & 0.0185 \\
C5 & 0.2044 & 0.0427 & 0.0290 & 0.0356 & 0.0334 & 0.0328 \\
C6 & 0.1571 & 0.0319 & 0.0161 & 0.0230 & 0.0217 & 0.0218 \\
C7 & 0.4889 & 0.0545 & 0.0619 & 0.0583 & 0.0569 & 0.0503 \\
C8 & 0.0808 & 0.0273 & 0.0161 & 0.0180 & 0.0189 & 0.0153 \\
C9 & 0.7231 & 0.1976 & 0.0850 & 0.1102 & 0.0318 & 0.0352 \\
C10 & 0.2185 & 0.0526 & 0.0532 & 0.0446 & 0.0428 & 0.0415 \\
C11 & 0.1255 & 0.0487 & 0.0298 & 0.0350 & 0.0338 & 0.0316 \\
C12 & 0.5216 & 0.1666 & 0.1210 & 0.1022 & 0.0870 & 0.0887 \\
C13 & 0.3584 & 0.0592 & 0.0548 & 0.0613 & 0.0490 & 0.0412 \\
C14 & 0.1342 & 0.0286 & 0.0201 & 0.0202 & 0.0182 & 0.0180 \\
C15 & 0.6064 & 0.1470 & 0.1307 & 0.0931 & 0.0864 & 0.0927 \\
C16 & 0.2456 & 0.0681 & 0.0674 & 0.0573 & 0.0514 & 0.0532 \\
C17 & 0.0655 & 0.0356 & 0.0147 & 0.0207 & 0.0181 & 0.0200 \\
\hline \hline
M. & 0.2936 & 0.0710 & 0.0512 & 0.0499 & 0.0415 & 0.0408 \\
\hline
T. & 4.9916 & 1.2078 & 0.8707 & 0.8480 & 0.7053 & 0.6936 \\
\hline
\end{tabular}
\end{center}
\end{table}

\begin{table}[!h]
\caption{\textbf{\emph{Soft--data}}.
As indicated, displayed values must be multiplied by $10^{-2}$ } \label{T6}
\begin{center}
\begin{tabular}{|c|c|c|c|c|c|c|}
\hline
 {\bf SC} ($\mathbf{x10^{-2}}$) & {\bf GRNN} & {\bf MLP} & {\bf SVR} & {\bf BNN} & {\bf RBF} &  {\bf GP} \\
  \hline
C1 & 0.1526 & 0.0857 & 0.0592 & 0.0501 & 0.0223 & 0.0305 \\
C2 & 0.1831 & 0.1393 & 0.0619 & 0.0606 & 0.0248 & 0.0285 \\
C3 & 0.1024 & 0.0804 & 0.0457 & 0.0374 & 0.0268 & 0.0278 \\
C4 & 0.0431 & 0.0212 & 0.0157 & 0.0154 & 0.0117 & 0.0120 \\
C5 & 0.0610 & 0.0362 & 0.0242 & 0.0240 & 0.0133 & 0.0138 \\
C6 & 0.0260 & 0.0167 & 0.0125 & 0.0135 & 0.0118 & 0.0121 \\
C7 & 0.3734 & 0.1301 & 0.0914 & 0.0737 & 0.0298 & 0.0398 \\
C8 & 0.0762 & 0.0435 & 0.0290 & 0.0238 & 0.0244 & 0.0180 \\
C9 & 0.4850 & 0.2467 & 0.1470 & 0.0818 & 0.0199 & 0.0360 \\
C10 & 0.1663 & 0.0607 & 0.0460 & 0.0421 & 0.0236 & 0.0276 \\
C11 & 0.1533 & 0.0540 & 0.0394 & 0.0383 & 0.0180 & 0.0209 \\
C12 & 0.3641 & 0.2325 & 0.1320 & 0.0887 & 0.0369 & 0.0480 \\
C13 & 0.2827 & 0.1350 & 0.0707 & 0.0699 & 0.0215 & 0.0304 \\
C14 & 0.0361 & 0.0197 & 0.0117 & 0.0121 & 0.0102 & 0.0104 \\
C15 & 0.3590 & 0.1843 & 0.1226 & 0.1049 & 0.0290 & 0.0506 \\
C16 & 0.1759 & 0.0754 & 0.0566 & 0.0489 & 0.0243 & 0.0302 \\
C17 & 0.0878 & 0.0352 & 0.0190 & 0.0219 & 0.0122 & 0.0134 \\
\hline \hline
N. & 0.1840 & 0.0939 & 0.0579 & 0.0475 & 0.0212 & 0.0265 \\
\hline
T. & 3.1282 & 1.5966 & 0.9845 & 0.8070 & 0.3605 & 0.4497 \\
\hline
\end{tabular}
\end{center}
\end{table}
 \begin{figure}[hptb]
\begin{center}
 \includegraphics[width=1\textwidth]{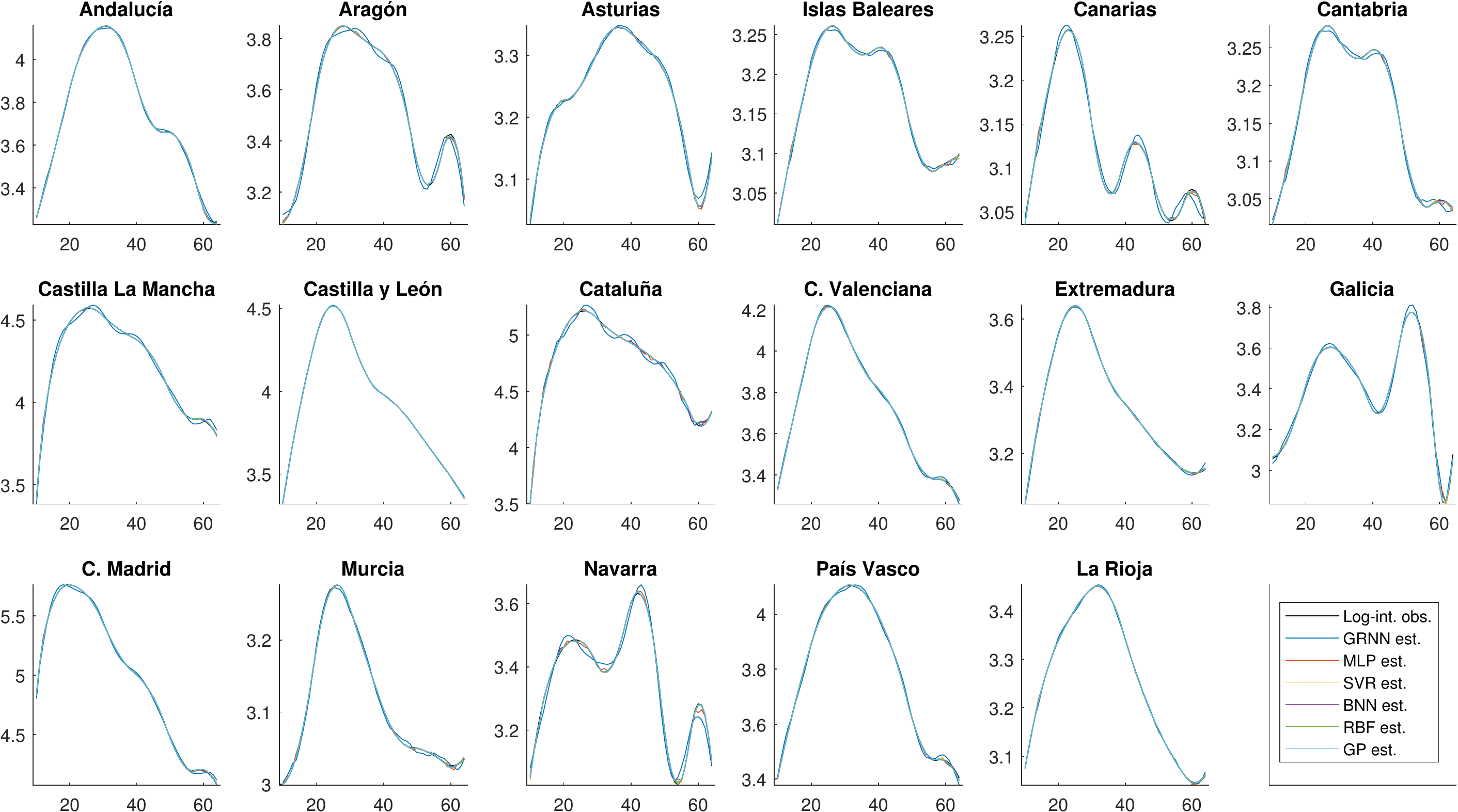}
\end{center}
 \caption{\textbf{\emph{Hard--data category}}. Observed and estimated COVID--19 mortality log--risk curves, from the implementation of  Generalized Regression Neural Network (GRNN), Multilayer Perceptron (MLP), Support Vector Regression (SVR), Bayesian Neural Network (BNN), Radial Basis Function Neural Network (RBF),
and Gaussian Processes (GP)}
 \label{figestorlrcod0}
 \end{figure}
 \begin{figure}[hptb]
 \begin{center}
 \includegraphics[width=1\textwidth]{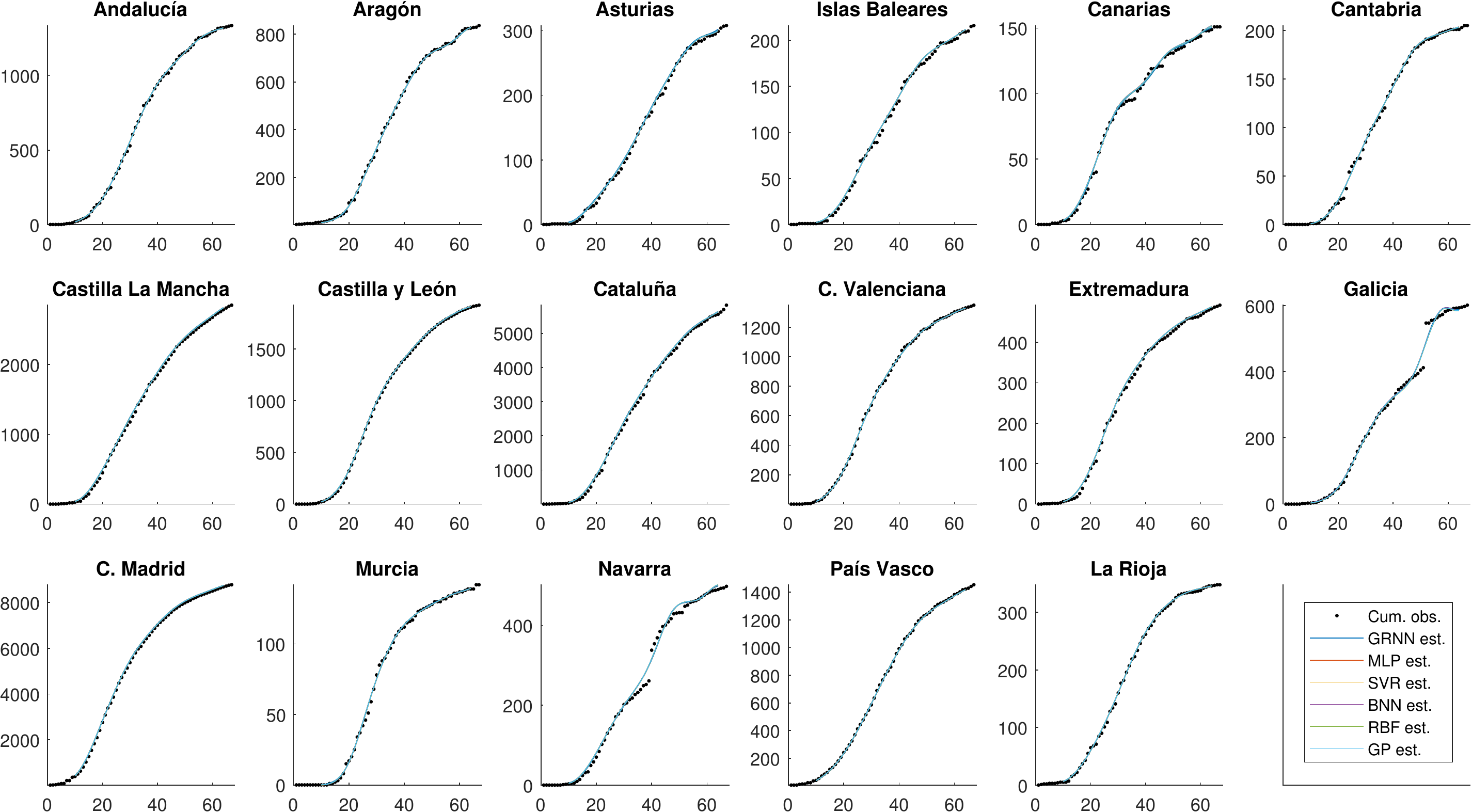}
 \end{center}
 \caption{\textbf{\emph{Hard--data category}}. Observed and estimated COVID--19 mortality  cumulative cases curves from the implementation of Generalized Regression Neural Network (GRNN), Multilayer Perceptron (MLP), Support Vector Regression (SVR), Bayesian Neural Network (BNN), Radial Basis Function Neural Network (RBF),
and Gaussian Processes (GP)}
 \label{figestorlrcod}
 \end{figure}
\begin{figure}[!h]
 \begin{center}
 \includegraphics[width=1\textwidth]{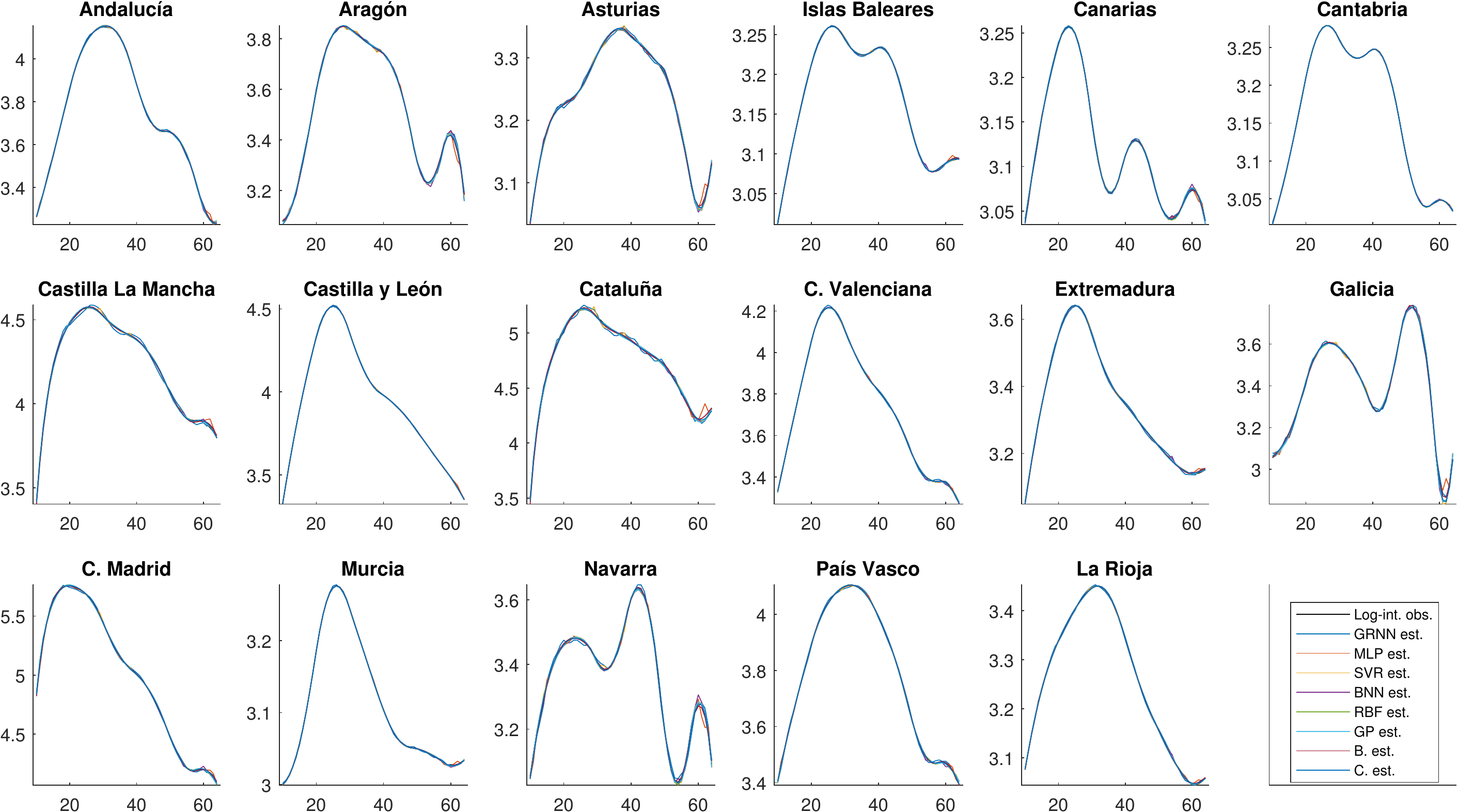}
 \end{center}
  \caption{\textbf{\emph{Soft--data category}}. Observed and estimated COVID--19 mortality log--risk curves, from the implementation of  Generalized Regression Neural Network (GRNN), Multilayer Perceptron (MLP), Support Vector Regression (SVR), Bayesian Neural Network (BNN), Radial Basis Function Neural Network (RBF),
Gaussian Processes (GP), and trigonometric regression combined with classical and Bayesian residual prediction}
 \label{fig1proyd}
 \end{figure}

 \begin{figure}[!h]
  \begin{center}
 \includegraphics[width=1\textwidth]{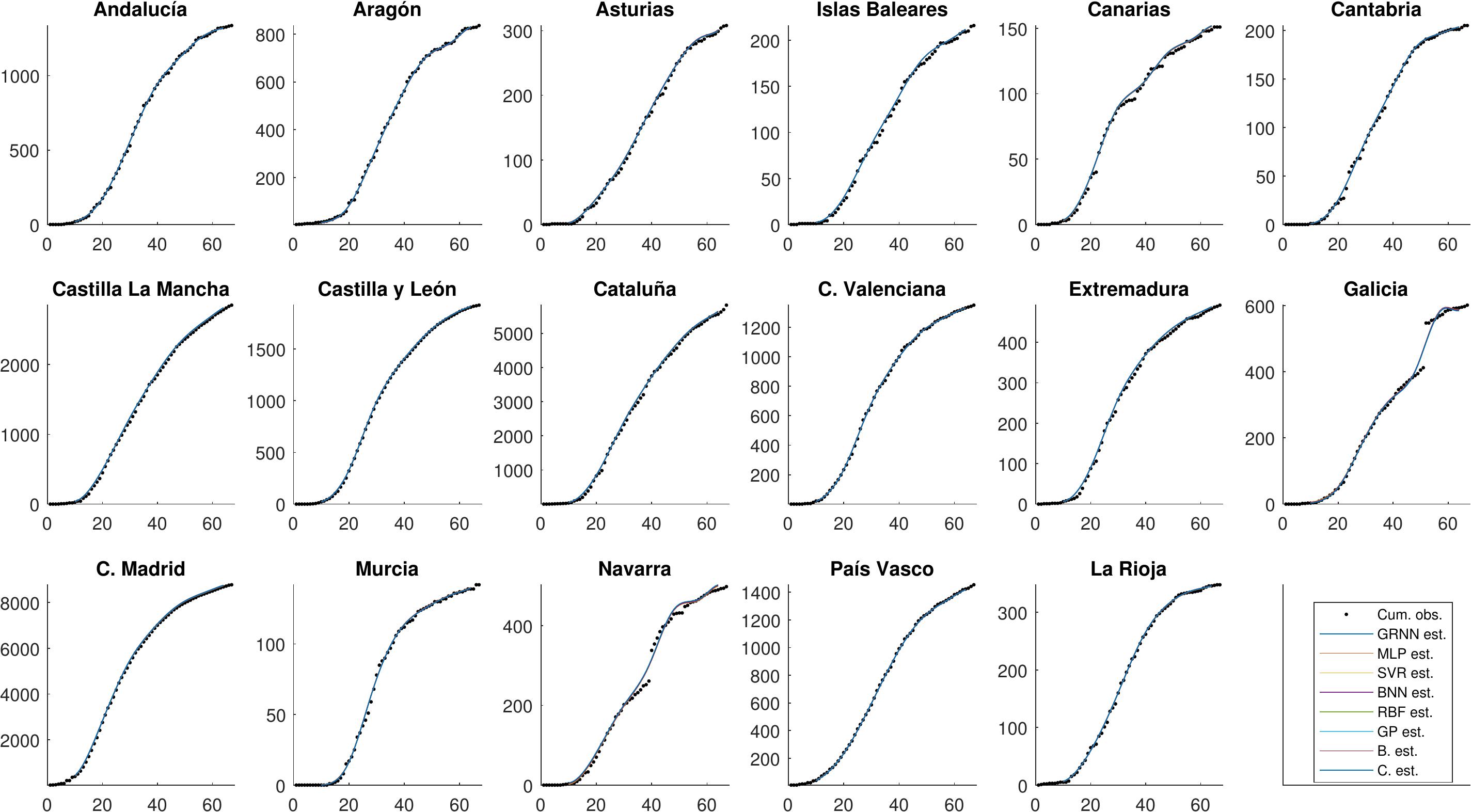}
 \end{center}
 \caption{\textbf{\emph{Soft--data category}}. Observed and estimated COVID--19 mortality cumulative cases curves, from the implementation of Generalized Regression Neural Network (GRNN), Multilayer Perceptron (MLP), Support Vector Regression (SVR), Bayesian Neural Network (BNN), Radial Basis Function Neural Network (RBF),
Gaussian Processes (GP), and trigonometric regression combined with classical and Bayesian residual prediction}
\label{fig1proyd2}
 \end{figure}

\clearpage
\subsection*{Randon $5$--fold cross--validation} The  random $5$--fold cross--validation  SMAPEs obtained from implementation of the six  ML regression models tested are displayed in Table \ref{T1}, for  hard--data category, and in Table \ref{T4},  for  soft--data category.
\begin{table}[!h]
\caption{\textbf{\emph{Hard--data category}}. Averaged SMAPEs for
$10$ running of random $5$--fold cross--validation. (As indicated, displayed values must be multiplied by $10^{-2}$)} \label{T1}
\begin{center}
\begin{tabular}{|c|c|c|c|c|c|c|}
\hline
 {\bf SC}($\mathbf{x10^{-2}}$) & {\bf GRNN} & {\bf MLP} &
 {\bf SVR} & {\bf BNN} & {\bf RBF} &  {\bf GP} \\
  \hline
C1 & 0.1962 & 0.0845 & 0.0890 & 0.0709 & 0.0635 & 0.0592 \\
C2 & 0.6150 & 0.1531 & 0.0711 & 0.0805 & 0.0782 & 0.0710 \\
C3 & 0.1541 & 0.0479 & 0.0438 & 0.0338 & 0.0371 & 0.0325 \\
C4 & 0.0984 & 0.0388 & 0.0190 & 0.0226 & 0.0214 & 0.0226 \\
C5 & 0.2065 & 0.0555 & 0.0378 & 0.0414 & 0.0429 & 0.0397 \\
C6 & 0.1585 & 0.0423 & 0.0227 & 0.0257 & 0.0266 & 0.0263 \\
C7 & 0.4957 & 0.0786 & 0.0771 & 0.0707 & 0.0712 & 0.0587 \\
C8 & 0.0804 & 0.0352 & 0.0226 & 0.0215 & 0.0247 & 0.0189 \\
C9 & 0.7280 & 0.2170 & 0.1061 & 0.0866 & 0.0421 & 0.0475 \\
C10 & 0.2208 & 0.0724 & 0.0751 & 0.0577 & 0.0541 & 0.0494 \\
C11 & 0.1273 & 0.0618 & 0.0373 & 0.0460 & 0.0451 & 0.0385 \\
C12 & 0.5237 & 0.1706 & 0.1441 & 0.1449 & 0.1126 & 0.1065 \\
C13 & 0.3637 & 0.0717 & 0.0701 & 0.0671 & 0.0605 & 0.0480 \\
C14 & 0.1359 & 0.0388 & 0.0307 & 0.0235 & 0.0220 & 0.0213 \\
C15 & 0.6105 & 0.1705 & 0.1592 & 0.1228 & 0.1101 & 0.1121 \\
C16 & 0.2479 & 0.0931 & 0.0926 & 0.0753 & 0.0665 & 0.0632 \\
C17 & 0.0667 & 0.0414 & 0.0195 & 0.0255 & 0.0241 & 0.0246 \\
\hline \hline
M. & 0.2958 & 0.0867 & 0.0657 & 0.0598 & 0.0531 & 0.0494 \\
\hline
T. & 5.0293 & 1.4731 & 1.1177 & 1.0166 & 0.9026 & 0.8400 \\
\hline
\end{tabular}
\end{center}
\end{table}

\begin{table}[!h]
\caption{\textbf{\emph{Soft--data category}}. Averaged SMAPEs, based
on $10$ running of random $5$--fold cross--validation. (As indicated, displayed values must be multiplied by $10^{-2}$)} \label{T4}
\begin{center}
\begin{tabular}{|c|c|c|c|c|c|c|}
\hline
 {\bf SC}($\mathbf{x10^{-2}}$) & {\bf GRNN} & {\bf MLP} & {\bf SVR} & {\bf BNN} & {\bf RBF} &  {\bf GP} \\
  \hline
C1 & 0.1569 & 0.0958 & 0.0695 & 0.0604 & 0.0240 & 0.0337 \\
C2 & 0.1836 & 0.1835 & 0.0697 & 0.0880 & 0.0286 & 0.0329 \\
C3 & 0.1029 & 0.1309 & 0.0474 & 0.0491 & 0.0273 & 0.0284 \\
C4 & 0.0433 & 0.0299 & 0.0171 & 0.0181 & 0.0133 & 0.0130 \\
C5 & 0.0609 & 0.0524 & 0.0282 & 0.0264 & 0.0153 & 0.0159 \\
C6 & 0.0259 & 0.0239 & 0.0133 & 0.0148 & 0.0130 & 0.0129 \\
C7 & 0.3783 & 0.2136 & 0.1018 & 0.0963 & 0.0309 & 0.0439 \\
C8 & 0.0774 & 0.0467 & 0.0315 & 0.0320 & 0.0292 & 0.0207 \\
C9 & 0.4968 & 0.2926 & 0.1458 & 0.1329 & 0.0254 & 0.0417 \\
C10 & 0.1710 & 0.0935 & 0.0541 & 0.0489 & 0.0277 & 0.0316 \\
C11 & 0.1556 & 0.0914 & 0.0456 & 0.0441 & 0.0221 & 0.0247 \\
C12 & 0.3759 & 0.2235 & 0.1509 & 0.1396 & 0.0402 & 0.0560 \\
C13 & 0.2894 & 0.1599 & 0.0903 & 0.0775 & 0.0259 & 0.0368 \\
C14 & 0.0375 & 0.0250 & 0.0124 & 0.0153 & 0.0114 & 0.0109 \\
C15 & 0.3646 & 0.2410 & 0.1378 & 0.1297 & 0.0334 & 0.0573 \\
C16 & 0.1792 & 0.0828 & 0.0677 & 0.0578 & 0.0292 & 0.0344 \\
C17 & 0.0900 & 0.0724 & 0.0216 & 0.0256 & 0.0129 & 0.0144 \\
\hline \hline
M. & 0.1876 & 0.1211 & 0.0650 & 0.0622 & 0.0241 & 0.0299 \\
\hline
T. & 3.1893 & 2.0589 & 1.1047 & 1.0567 & 0.4100 & 0.5090 \\
\hline
\end{tabular}
\end{center}
\end{table}

\section*{Acknowledgements}

This work has been supported in part by projects PGC2018-099549-B-I00
of the Ministerio de Ciencia, Innovaci\'on y Universidades, Spain
(co-funded with FEDER funds), and  by grant A-FQM-345-UGR18 cofinanced by ERDF Operational Programme 2014-2020, and the Economy and Knowledge Council of the Regional Government of Andalusia, Spain.

The subject of this paper was originally developed, in a first stage, under the seminars hold in the \emph{Unidad de Transferencia del IMUS}
about \emph{Matem\'aticas y la COVID}. We also thank  the
organizer,  Professor Emilio Carrizosa.


\begin{thebibliography}{9}
\bibitem{Aalen08}
O. O. Aalen, O. Borgan and H. K. Gjessing (2008). \emph{Survival and
event history analysis: a process point of view}. Springer Science
\& Business Media, New--York.
\bibitem{Abboud19}
 C. Abboud, O. Bonnefon, E. Parent and S. Soubeyrand (2019). Dating and localizing an invasion
from post-introduction data and a coupled
reaction--diffusion--absorption model. \emph{Journal of Mathematical
Biology} \textbf{79}, 765--789.

\bibitem{Agostinelli01}
C. Agostinelli (2001). Robust model selection in regression via
weighted likelihood methodology. \emph{Statistics \& Probability
Letters} \textbf{56}, 289–-300.

\bibitem{Alpaydin04}
E. Alpaydin  (2004). \emph{Introduction to Machine Learning}.  MIT
Press, Cambridge, MA.

\bibitem{Anderson}
H. Anderson and T. Britton (2000). \emph{Stochastic epidemic models
and their statistical analysis}. Springer--Verlag, New--York.

\bibitem{Angulo13}
J. Angulo, H.--L. Yu, A. Langousis, A. Kolovos, J. Wang, A. E. Madrid and  G Christakos (2013).
Spatiotemporal infectious disease modeling: A BME-SIR Approach.
\emph{PLoS One}  8(9): e72168.

\bibitem{Barstugan20} M. Barstugan, U. Ozkaya and S. Ozturk  (2020). Coronavirus (COVID--19)
classification using ct images by machine learning methods.
arXiv preprint arXiv:2003.09424

\bibitem{Beretta} E.
 E. Beretta, T. Hara, W. Ma and Y. Takeuchi (2001). Global asymptotically stability of an SIR epidemic model
with distributed time delay. \emph{Nonlinear Anal Theory Methods
Appl} \textbf{47}, 4107--4115.

\bibitem{Blanqueroetal20}
R. Blanquero, E. Carrizosa, M. A. Jim\'enez--Cordero and B.
Mart\'{\i}n--Barrag\'an (2020). Selection of time instants and
intervals with support vector regression for multivariate functional
data. \emph{Computers \& Operations Research} \textbf{123} 10.1016/j.cor.2020.105050.

\bibitem{Bolker96}B. M.
Bolker and B. Grenfell (1996). Impact of vaccination on the spatial
correlation and persistence of measles dynamics. \emph{Proceedings
of the National Academy of Sciences} \textbf{93}, 12648--12653.

\bibitem{Bosq2000}
D. Bosq  (2000). \emph{Linear processes in function spaces}. Lecture
notes in statistics \textbf{149}.
 Springer, New--York.

\bibitem{Bosq14}
D. Bosq and M. D.  Ruiz--Medina (2014).
 Bayesian estimation in a high dimensional parameter framework.
\emph{Electron J Statist} \textbf{8}, 1604--1640.




\bibitem{Chao12} D. L. Chao, J. D. Bloom,  B. F. Kochin,   R. Antia and  I. M. Longini
(2012). The global spread of drug-resistant influenza. \emph{Journal
of the Royal Society Interface} \textbf{9}, 648--656.

\bibitem{Chapelle}
O. Chapelle, V. Vapnik and  Y. Bengio (2002). Model selection for
small sample regression. \emph{Machine Learning} \textbf{48}, 9–-23.

\bibitem{Chien20}
L.--Ch. Chien  and  L.--W. Chen (2020).
Meteorological impacts on the incidence of COVID-19 in the U.S.
\emph{Stoch Environ Res Risk Assess} \textbf{34}, 1675–-1680.

\bibitem{Christakos00} G. Christakos (2000). Modern spatiotemporal geostatistics. Oxford University Press, New--York, NY.
\bibitem{Christakos02} G. Christakos (2002).   On assimilation of uncertain physical knowledge bases: Bayesian and non--Bayesian techniques.
\emph{Adv. Water Resources} \textbf{25}: 1257--1274.
\bibitem{Christakos} G.
Christakos, P. Bogaert and M. L. Serre (2002). Advanced functions
of temporal GIS, Springer-Verlag, New York, N.Y.
\bibitem{ChristakosHris} G. Christakos and D.T. Hristopulos (1998).  Spatiotemporal environmental health modelling: a Tractatus Stochaticus.
Kluwer, Boston.
\bibitem{Christakos08} G. Christakos (2008).  Bayesian maximum entropy. In \emph{Advanced mapping of environmental data: geostatistics, machine learning, and Bayesian maximum entropy}. Wiley, New York, NY, pp. 247--306.

\bibitem{Daubechies88}
Daubechies, I. (1988). Orthonormal basis of compactly supported wavelets. Comm. Pure Appl. Math. \textbf{41}, 909--9996.
\bibitem{Du20}   Z. Du, X. Xu, Y. Wu, L. Wang,  L. A. Cowling and  B. J. Meyers (2020). Serial interval of COVID-19 among publicly
reported confirmed cases. \emph{Emerg. Infect. Dis.} \textbf{26}(6).
\bibitem{Dushoff04} J. Dushoff,    J. Plotkin, S. Levin and  D. Earn  (2004). Dynamical resonance
can account for seasonality of influenza epidemics.
\emph{Proceedings of the National Academy of Sciences of the United
States of America} \textbf{101}, 16915--16916.
\bibitem{Elhia14}
M. Elhia, A. Laaroussi, M. Rachik, Z. Rachik and E. Labriji  (2014).
Global stability of a susceptible--infected--recovered (SIR)
epidemic model with two infectious stages and treatment. \emph{Int J
Sci Res} \textbf{3}, 114--121.
\bibitem{Fleming91}
T. R. Fleming and D. P. Harrington (1991). \emph{Counting processes
and survival analysis}. Wiley Series in Probability and Mathematical
Statistics: Applied Probability and Statistics. John Wiley \& Sons,
Inc., New--York.
\bibitem{Guin14}
L. N. Guin and  P. K. Mandal  (2014). Spatiotemporal dynamics of
reaction--diffusion models of interacting populations. \emph{Appl
Math Model} \textbf{38}, 4417--4427.

\bibitem{Hastie01} T. Hastie, R. Tibshirani and J. Friedman (2001). \emph{The elements of statistical learning}. Springer Series in
Statistics. Springer--Verlag, New--York.

\bibitem{He20}
 J. He ,   G. Chen, Y. Jiang , R. Jin ,  A. Shortridge,  S. Agusti,  M. Hea,  J.
 Wua,  C.M.  Duarte,  G. Christakos   (2020).
 \newblock Comparative infection modeling and control of COVID-19 transmission
patterns in China, South Korea, Italy and Iran.
\newblock{Science of the Total Environment}~{747}


\bibitem{Huppert}
A. Huppert and  G. Katriel  (2013). Mathematical modelling and
prediction in infectious disease epidemiology. \emph{Clin Microbiol
Infect} \textbf{19}, 999--1005.

\bibitem{Ivanov15}
 A.V. Ivanov, N.N. Leonenko,  M.D. Ruiz Medina, and B.M. Zhurakovsky (2015).
 Estimation of harmonic component in regression with cyclically dependent errors. \emph{Stastics: A Journal of Theoretical and Applied Statistics}   \textbf{49},   156--186.


\bibitem{Ivorra20a}
B. Ivorra, M.R. Ferr\'andez, M. Vela-P\'erez and  A.M. Ramos
(2020). Mathematical modeling of the spread of the coronavirus
disease 2019 (COVID-19) taking into account the undetected
infections. The case of China. \emph{Commun Nonlinear Sci Numer
Simulat} \textbf{88}, 105--303.

\bibitem{Ivorra20b}
B. Ivorra, A.M. Ramos and  D. Ngom  (2015). Be-CoDiS: A
mathematical model to predict the risk of human diseases spread
between countries. Validation and application to the 2014 ebola
virus disease epidemic. \emph{Bull Math Biol} \textbf{77},
1668--1704.
\bibitem{Ji12} C. Ji, D. Jiang and N. Shi  (2012). The behavior of an SIR epidemic model
with stochastic perturbation. \emph{Stoch Anal Appl.} \textbf{30},
755--773.
\bibitem{Keeling2}
M.J. Keeling, D.A.  Rand and A.J. Morris (1997). Correlation
models for childhood epidemics. \emph{Proceedings of the Royal
Society of London} \textbf{264}, 1149--1156.
\bibitem{Keeling}
M.J. Keeling and  P. Rohani  (2008). \emph{Modeling infectious
diseases in humans and animals}. Princeton University Press,
Princeton.

\bibitem{Kermack} W.
Kermack  and A. McKendrick (1927). Contributions to the mathematical
theory of epidemics - I. \emph{Proceedings of the Royal Society of
Edinburgh A} \textbf{115}, 700--721.
\bibitem{Khan20}
M.A. Khan and A. Atangana (2020). Modeling the dynamics of novel
coronavirus (2019-nCov) with fractional derivative. \emph{Alex. Eng.
J.}. doi.org/10.1016/j.aej.2020.02.033.
\bibitem{Kucharski20}
A.J. Kucharski, T.W. Russell, C. Diamond, Y. Liu, J. Edmunds and
S. Funk et al. (2020). Early dynamics of transmission and control of
COVID-19: a mathematical modelling study. \emph{Lancet Infect Dis.}
doi.org/10.1016/S1473-3099(20)30144-4.
\bibitem{Kuznetsov20}
Y.A. Kuznetsov and C. Piccardi  (1994). Bifurcation analysis of
periodic SEIR and SIR epidemic models. \emph{J Math Biol}
\textbf{32}, 109--121.



\bibitem{Laaroussi18}
A.E. Laaroussi,  M. Rachik  and M. Elhia (2018). An optimal control
problem for a spatiotemporal SIR model \emph{Int. J. Dynam. Control}
\textbf{6}, 384--397.

\bibitem{Langousis20}
A. Langousis and  A.A. Carsteanu (2020).
Undersampling in action and at scale: application to the COVID-19 pandemic.  \emph{Stoch Environ Res Risk Assess} \textbf{34}, 1281–-1283.

\bibitem{Malesios16}
C. Malesios, N. Demiris, P. Kostoulas, K. Dadousis, T.
Koutroumanidis and Z. Abas (2016). Spatio-temporal modelling of
foot-and-mouth disease outbreaks. \emph{Epidemiol. Infect.}
\textbf{144}, 2485--2493.

\bibitem{McCluskey10}
C.C. McCluskey  (2010). Complete global stability for an SIR
epidemic model with delay distributed or discrete. \emph{Nonlinear
Anal Real World Appl} \textbf{11}, 55--59.

\bibitem{Milner08}F. A.
Milner and R. Zhao  (2008). SIR model with directed spatial
diffusion. \emph{Math Popul Stud} \textbf{15}, 160--181.

\bibitem{Mohammady21}
M. Mohammady, H. Reza Pourghasemi, M. Amiri and  J.P. Tiefenbacher (2021).
Spatial modeling of susceptibility to subsidence using machine learning techniques.
https://doi.org/10.1007/s00477-020-01967-x

\bibitem{Nishiura20}
H. Nishiura, N.M. Linton and A. R. Akhmetzhanov (2020). Serial
interval of novel coronavirus (COVID-19) infections. \emph{Int. J.
Infect. Dis.} \textbf{93}, 284--286.

\bibitem{Pak2020}
D. Pak, K. Langohr, J. Ning, J. Cort\'es Mart\'{\i}nez, G.G\'omez--Melis and Y. Shen (2020).
Modeling the coronavirus disease 2019 incubation period: impact on quarantine policy
doi.org/10.1101/2020.06.27.20141002.


\bibitem{Pathak10}
S. Pathak, A. Maiti and  G. Samanta  (2010). Rich dynamics of an SIR
epidemic model. \emph{Nonlinear Anal Model Control} \textbf{15},
71--81.


\bibitem{Ramos2020}
A.M. Ramosa, M.R. Ferr\'andez, M. Vela-P\'erez and  B. Ivorra (2020).
A simple but complex enough $\theta $--SIR type model to be used with
COVID--19 real data. Application to the case of Italy. doi.org/10.13140/RG.2.2.32466.17601.


\bibitem{Remuzzi20}
A. Remuzzi and G. Remuzzi (2020). COVID-19 and Italy: what next? The
Lancet. doi.org/10.1016/S0140-6736(20)30690-5
\bibitem{Roosa2020}
K. Roosa, Y. Lee, R. Luo, A. Kirpich, R. Rothenberg, J. Hyman  et
al. (2020). Real--time forecasts of the COVID-19 epidemic in China
from February 5th to February 24th. \emph{Infect Dis Modell}
\textbf{5}, 256--263.
\bibitem{Roques16}
 L. Roques and O. Bonnefon (2016). Modelling population dynamics in realistic landscapes with linear
elements: A mechanistic-statistical reaction-diffusion approach.
\emph{PloS One} \textbf{11}(3):e0151217.
\bibitem{Roques11}
L. Roques, S. Soubeyrand and J. Rousselet (2011). A
statistical-reaction-diffusion approach for analyzing expansion
processes. \emph{J Theor Biol.} \textbf{274}, 43--51.
\bibitem{Sekiguchi10}
 M. Sekiguchi and  E. Ishiwata (2010). Global dynamics of a discretized
SIRS epidemic model with time delay. \emph{J Math Anal Appl}
\textbf{371}, 195--202.

\bibitem{Sivakumar20}
B. Sivakumar  (2020).
 COVID-19 and water. \emph{Stoch Environ Res Risk Assess}. https://doi.org/10.1007/s00477-020-01837-6
\bibitem{Sujath20a}
    R. Sujath, J.M. Chatterjee and  A.E. Hassanien (2020).
A machine learning forecasting model for COVID-19 pandemic in India. \emph{Stoch Environ Res Risk Assess}.
\textbf{34}, 959–-972.


\bibitem{Sujath20b}
R. Sujath, J.M. Chatterjee and  A.E. Hassanien (2020).
Correction to: A machine learning forecasting model for COVID-19 pandemic in India. \emph{Stoch Environ Res Risk Assess}.
https://doi.org/10.1007/s00477-020-01843-8

\bibitem{Takano20}
Y. Takano and R. Miyashiro  (2020). Best subset selection via
cross--validation criterion. \emph{TOP} \textbf{28}, 475–-488.
\bibitem{Tornatore}
E. Tornatore,  S.M.  Buccellato and  P.  Vetro  (2005). Stability
of a stochastic SIR system. \emph{Phys A Stat Mech Its Appl}
\textbf{354}, 111--126.


\bibitem{Volz08} E. Volz (2008). SIR dynamics in random networks with heterogeneous connectivity.
\emph{Journal of Mathematical Biology} \textbf{56}, 293--310.

\bibitem{Wang20}
C. Wang, P.W. Horby, F. Hayden and G.F. Gao (2020). A novel
coronavirus outbreak of global health concern. Lancet \textbf{395},
470--473.

\bibitem{Wasiur19}
R.K.  Wasiur, B. Choiy, E. Kenahz and G.A. Rempa (2019). Survival
dynamical systems for the population-level analysis of epidemics.
arXiv.1901.00405.

\bibitem{Webb81}
 G. Webb (1981). A reaction-diffusion model for a deterministic diffusive
epidemic. \emph{J Math Anal Appl} \textbf{84}, 150--161.


\bibitem{Xu07} Y. Xu, L. Allena and   A.  Perelson (2007). Stochastic model of an influenza
epidemic with drug resistance. \emph{Journal of Theoretical Biology}
\textbf{248}, 179--193.

\bibitem{Yu09}J. Yu, D. Jiang  and  N. Shi (2009). Global stability of two-group SIR model
with random perturbation. \emph{J Math Anal Appl.} \textbf{360},
235--244.

\bibitem{Zhang20}
Wb. Zhang, Y. Ge, M. Liu  et al.  (2020)  Risk assessment of the step-by-step return-to-work policy in Beijing following the COVID-19 epidemic peak. \emph{Stoch Environ Res Risk Assess}. https://doi.org/10.1007/s00477-020-01929-3



\bibitem{Zhang08}
F. Zhang, Z. Li and F. Zhang (2008). Global stability of an SIR
epidemic model with constant infectious period. \emph{Appl Math
Comput.} \textbf{199}, 285--291.



\bibitem{Zhou06} T. Zhou, Z. Fu and B. Wang (2006). Epidemic dynamics on complex networks.
\emph{Progress in Natural Science} \textbf{16}, 452--457.

\bibitem{Zhou20}
F. Zhou, T. Yu, R. Du, G. Fan, Y. Liu, Z. Liu, J. Xiang, Y. Wang, B.
Song, X. Gu, et al (2020). Clinical course and risk factors for
mortality of adult inpatients with COVID--19 in Wuhan, China: a
retrospective cohort study. \emph{The Lancet}.
doi.org/10.1016/S0140-6736(20)30566-3.

\end{thebibliography}
\end{document}